\documentclass{article}

\usepackage[final,nonatbib]{neurips_2020}

\usepackage[utf8]{inputenc} 
\usepackage[T1]{fontenc}    
\usepackage{hyperref}       
\usepackage{url}            
\usepackage{booktabs}       
\usepackage{amsfonts}       
\usepackage{nicefrac}       
\usepackage{microtype}      
\usepackage[sort&compress,numbers]{natbib}
\usepackage{wrapfig}
\usepackage{listings}
\usepackage{algorithm}
\usepackage{algorithmic}
\usepackage{caption}
\usepackage{amsthm}
\usepackage{mathtools}
\usepackage{xspace}
\usepackage{siunitx}
\usepackage[pdftex]{graphicx}
\usepackage{subcaption}

\usepackage{amsmath,amsfonts,bm}

\def\eqref#1{equation~\ref{#1}}

\def\1{\bm{1}}

\DeclareMathAlphabet{\mathsfit}{\encodingdefault}{\sfdefault}{m}{sl}
\SetMathAlphabet{\mathsfit}{bold}{\encodingdefault}{\sfdefault}{bx}{n}

\usepackage{authblk}

\newtheorem{proposition}{Proposition}
\newtheorem{definition}{Definition}
\usepackage{xcolor}
\definecolor{pinegreen}{rgb}{0.0, 0.47, 0.44}
\definecolor{deepmagenta}{rgb}{0.8, 0.0, 0.8}
\definecolor{amber}{rgb}{1.0, 0.49, 0.0}

\newcommand{\softmax}[1]{\texttt{Softmax}\,(#1)}

\newcommand{\gru}[2]{\texttt{GRU}\,(#1, \ #2)}
\newcommand{\slots}[0]{\texttt{slots}}
\newcommand{\attn}[0]{\texttt{attn}}
\newcommand{\inp}[0]{\texttt{inputs}}
\newcommand{\mlp}[0]{\texttt{MLP}}
\newcommand{\sam}[0]{Slot Attention\xspace}

\newcommand{\updates}[0]{\texttt{updates}}
\newcommand{\layernorm}[0]{\texttt{LayerNorm}}

\DeclareCaptionLabelFormat{andtable}{\tablename~\thetable~\& #1~#2}

\definecolor{lightgrey}{rgb}{0.43,0.43,0.43}
\definecolor{crimson}{rgb}{0.86,0.08,0.24}
\hypersetup{
  colorlinks,
  citecolor=lightgrey,
  linkcolor=crimson,
  urlcolor=lightgrey}

\newcommand\blfootnote[1]{%
  \begingroup
  \renewcommand\thefootnote{}\footnote{#1}%
  \addtocounter{footnote}{-1}%
  \endgroup
}

\title{Object-Centric Learning with Slot Attention}

\author[2,3,$\dagger$,*]{Francesco Locatello}
\author[1]{Dirk Weissenborn}
\author[1]{Thomas Unterthiner}
\author[1]{Aravindh Mahendran}
\author[1]{Georg Heigold}
\author[1]{Jakob Uszkoreit}
\author[1,$\ddagger$]{Alexey Dosovitskiy}
\author[1,$\ddagger$,*]{Thomas Kipf}

\affil[1]{Google Research, Brain Team}
\affil[2]{Dept.~of Computer Science, ETH Zurich}
\affil[3]{Max-Planck Institute for Intelligent Systems}

\begin{document}

\maketitle

\begin{abstract}
\looseness=-1Learning object-centric representations of complex scenes is a promising step towards enabling efficient abstract reasoning from low-level perceptual features. Yet, most deep learning approaches learn distributed representations that do not capture the compositional properties of natural scenes. In this paper, we present the Slot Attention module, an architectural component that interfaces with perceptual representations such as the output of a convolutional neural network and produces a set of task-dependent abstract representations which we call slots. These slots are exchangeable and can bind to any object in the input by specializing through a competitive procedure over multiple rounds of attention. We empirically demonstrate that Slot Attention can extract object-centric representations that enable generalization to unseen compositions when trained on unsupervised object discovery and supervised property prediction tasks.
\end{abstract}

\section{Introduction}
\blfootnote{\kern-1.7em$^\dagger$Work done while interning at Google, $^*$equal contribution, $^\ddagger$equal advising. Contact: \href{mailto:tkipf@google.com}{\texttt{tkipf@google.com}}}%
\looseness=-1Object-centric representations have the potential to improve sample efficiency and generalization of machine learning algorithms across a range of application domains, such as visual reasoning~\citep{yi2019clevrer}, modeling of structured environments~\citep{kulkarni2019unsupervised}, multi-agent modeling~\citep{sun2019stochastic,berner2019dota,vinyals2019grandmaster}, and simulation of interacting physical systems~\citep{battaglia2016interaction,mrowca2018flexible,sanchez2020learning}. Obtaining object-centric representations from raw perceptual input, such as an image or a video, is challenging and often requires either supervision~\citep{watters2017visual,sun2019stochastic,yi2019clevrer,knyazev2020graph} or task-specific architectures~\citep{devin2018deep,kulkarni2019unsupervised}. As a result, the step of learning an object-centric representation is often skipped entirely. Instead, models are typically trained to operate on a structured representation of the environment that is obtained, for example, from the internal representation of a simulator~\citep{battaglia2016interaction,sanchez2020learning} or of a game engine~\citep{berner2019dota,vinyals2019grandmaster}.

\looseness=-1To overcome this challenge, we introduce the Slot Attention module, a differentiable \textit{interface} between perceptual representations (e.g., the output of a CNN) and a \textit{set} of variables called \textit{slots}. Using an iterative attention mechanism, \sam produces a set of output vectors with permutation symmetry. Unlike \textit{capsules} used in Capsule Networks~\citep{sabour2017dynamic,hinton2018matrix}, slots produced by \sam do not specialize to one particular type or class of object, which could harm generalization. Instead, they act akin to \textit{object files}~\citep{kahneman1992reviewing}, i.e., slots use a common representational format: each slot can store (and bind to) any object in the input. This allows \sam to generalize in a systematic way to unseen compositions, more objects, and more slots.

\sam is a simple and easy to implement architectural component that can be placed, for example, on top of a CNN~\citep{lecun1995convolutional} encoder to extract object representations from an image and is trained end-to-end with a downstream task. In this paper, we consider image reconstruction and set prediction as downstream tasks to showcase the versatility of our module both in a challenging unsupervised object discovery setup and in a supervised task involving set-structured object property prediction.

\looseness=-1\textbf{Our main contributions} are as follows:
(i) We introduce the Slot Attention module, a simple architectural component at the interface between perceptual representations (such as the output of a CNN) and representations structured as a set.
(ii) We apply a \sam-based architecture to unsupervised object discovery, where it matches or outperforms relevant state-of-the-art approaches~\citep{greff2019multi,burgess2019monet}, while being more memory efficient and significantly faster to train.
(iii) We demonstrate that the Slot Attention module can be used for supervised object property prediction, where the attention mechanism learns to highlight individual objects without receiving direct supervision on object segmentation.

\section{Methods}
\begin{figure}[t!]
    \centering
    \begin{subfigure}[b]{0.445\textwidth}
        \includegraphics[width=\textwidth]{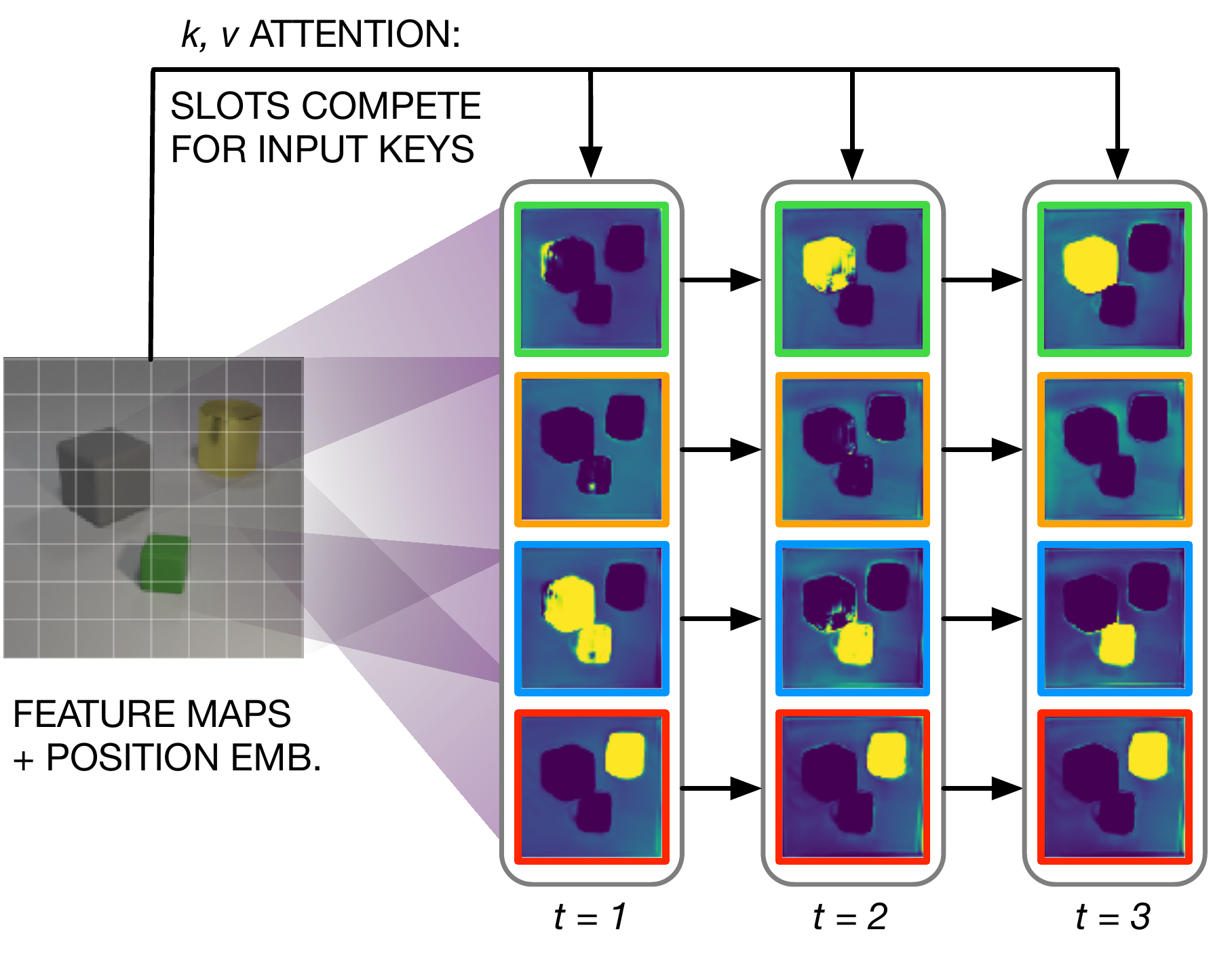}
        \caption{Slot Attention module.}
        \label{fig:slota_a}
    \end{subfigure}
    \quad
    \begin{subfigure}[b]{0.52\textwidth}
        {
            \includegraphics[width=\textwidth]{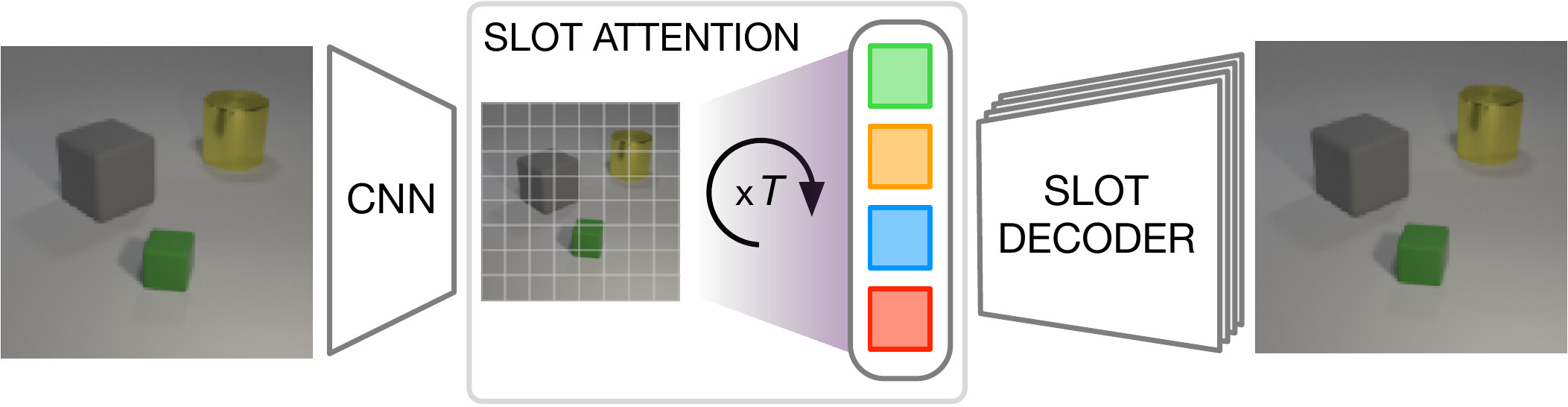}
            \caption{Object discovery architecture.}
            \label{fig:slota_c}
        }
        \vspace{0.8em}
        {
            \includegraphics[width=\textwidth]{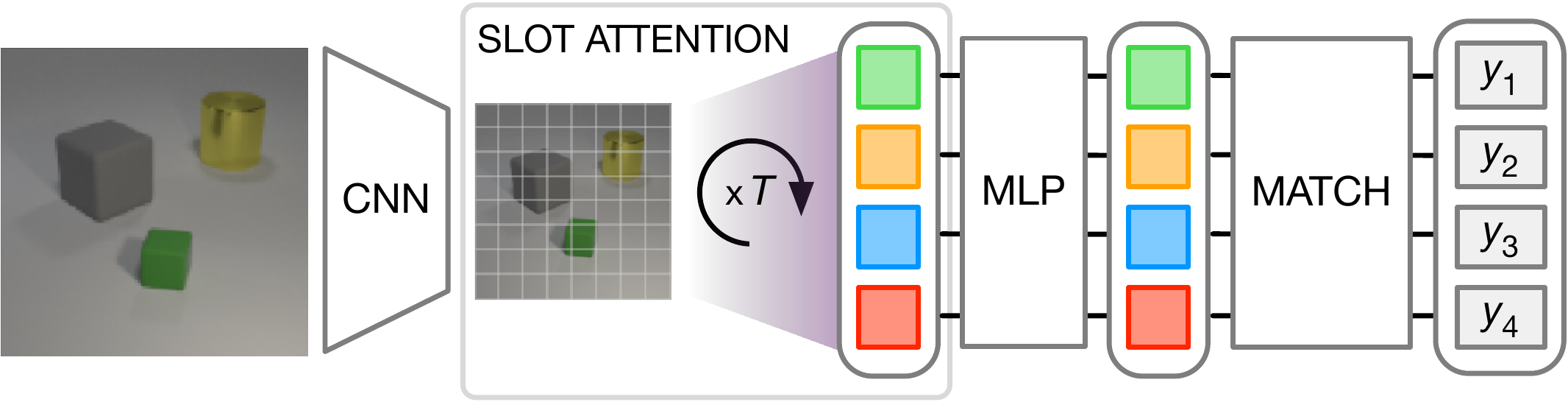}
            \caption{Set prediction architecture.}
            \label{fig:slota_b}
        }
    \end{subfigure}
    \caption{
    \looseness=-1 (\textbf{a}) Slot Attention module and example applications to (\textbf{b}) unsupervised object discovery and (\textbf{c}) supervised set prediction with labeled targets $y_i$. See main text for details.}\label{fig:slota}
\end{figure}

In this section, we introduce the Slot Attention module (Figure~\ref{fig:slota_a}; Section~\ref{sec:slot_attention_module}) and demonstrate how it can be integrated into an architecture for unsupervised object discovery (Figure~\ref{fig:slota_c}; Section~\ref{sec:object_discovery}) and into a set prediction architecture (Figure~\ref{fig:slota_b}; Section~\ref{sec:set_prediction}).

\subsection{Slot Attention Module}
\label{sec:slot_attention_module}
\looseness=-1 The Slot Attention module (Figure~\ref{fig:slota_a}) maps from a set of $N$ input feature vectors to a set of $K$ output vectors that we refer to as \textit{slots}. Each vector in this output set can, for example, describe an object or an entity in the input. The overall module is described in Algorithm~\ref{algo:slot_attention} in pseudo-code\footnote{An implementation of Slot Attention is available at: \url{https://github.com/google-research/google-research/tree/master/slot_attention}.}.

Slot Attention uses an iterative attention mechanism to map from its inputs to the slots. Slots are initialized at random and thereafter refined at each iteration $t=1\ldots T$ to bind to a particular part (or grouping) of the input features. Randomly sampling initial slot representations from a common distribution allows Slot Attention to generalize to a different number of slots at test time.

At each iteration, slots \textit{compete} for explaining parts of the input via a softmax-based attention mechanism \citep{bahdanau2014neural,luong2015effective,vaswani2017attention} and update their representation using a recurrent update function. The final representation in each slot can be used in downstream tasks such as  unsupervised object discovery (Figure~\ref{fig:slota_c}) or supervised set prediction (Figure~\ref{fig:slota_b}).

\begin{algorithm}[t!]
\caption{Slot Attention module. The input is a set of $N$ vectors of dimension $D_{\inp}$ which is mapped to a set of $K$ slots of dimension $D_{\slots}$. We initialize the slots by sampling their initial values as independent samples from a Gaussian distribution with shared, learnable parameters $\mu\in\mathbb{R}^{D_{\slots}}$ and $\sigma\in\mathbb{R}^{D_{\slots}}$. In our experiments we set the number of iterations to $T=3$.}\label{algo:slot_attention}
\begin{algorithmic}[1]

  \STATE \textbf{Input}: $\inp\in\mathbb{R}^{N \times D_{\inp}}$, $\slots \sim \mathcal{N}(\mu, \ \mathrm{diag(\sigma)})\in\mathbb{R}^{K \times D_{\slots}}$\\[0.2em]
  \STATE \textbf{Layer params}: $k, \ q, \ v$: linear projections for attention; $\texttt{GRU}$; $\texttt{MLP}$;  \layernorm \,(x3)\\[0.2em]
  \STATE \quad $\inp = \layernorm\,(\inp)$\label{algo_step:norm_input}\\[0.2em]
  \STATE \quad \textbf{for} $t = 0\ldots T$ \\[0.2em]
  \STATE \qquad $\slots\texttt{\_prev} = \slots$\\[0.2em]
  \STATE \qquad $\slots = \layernorm\,(\slots)$\\[0.2em]
  \STATE \qquad $\attn = \softmax{\frac{1}{\sqrt{D}} k(\inp) \cdot q(\slots)^T,\,\texttt{axis=`slots'}} $  \hfill\COMMENT{{\color{gray}\# norm.~over slots}}\label{algo_step:softmax}
  \STATE \qquad $\updates = \texttt{WeightedMean}\,(\texttt{weights=}\texttt{attn}+\epsilon,\,\texttt{values=}v(\inp))$  \hfill\COMMENT{{\color{gray} \# aggregate}}\label{algo_step:updates}\\[0.2em]
  \STATE \qquad $\slots = \gru{\texttt{state=}\slots\texttt{\_prev}}{\texttt{inputs=}\updates} $  \hfill\COMMENT{{\color{gray}\# GRU update (per slot)}}\\[0.2em]
  \STATE \qquad $\slots \mathrel{+}= \mlp\,(\layernorm\,(\slots))$  \hfill\COMMENT{{\color{gray}\# optional residual MLP (per slot)}}\\[0.2em]
  \STATE \quad \textbf{return} $\slots$\label{algo_step:return}\\[0.2em]
\end{algorithmic}
\end{algorithm}

We now describe a single iteration of Slot Attention on a set of input features, $\inp\in\mathbb{R}^{N\times D_{\inp}}$, with $K$ output slots of dimension $D_{\slots}$ (we omit the batch dimension for clarity). We use learnable linear transformations $k$, $q$, and $v$ to map inputs and slots to a common dimension $D$.

Slot Attention uses dot-product attention~\citep{luong2015effective} with attention coefficients that are normalized over the slots, i.e., the queries of the attention mechanism. This choice of normalization introduces competition between the slots for explaining parts of the input.

We further follow the common practice of setting the softmax temperature to a fixed value of $\sqrt{D}$~\citep{vaswani2017attention}:
\begin{equation}\label{eqn:slot_attention_key_query}
      \attn_{i, j} \coloneqq \frac{e^{M_{i, j}}}{\sum_l e^{M_{i, l}}} \qquad \text{where} \qquad M \coloneqq \frac{1}{\sqrt{D}}k(\inp) \cdot q(\slots)^T \in\mathbb{R}^{N\times K}.
\end{equation}
\looseness=-1 In other words, the normalization ensures that attention coefficients sum to one for each individual input feature vector, which prevents the attention mechanism from ignoring parts of the input. To aggregate the input values to their assigned slots, we use a weighted mean as follows:
\begin{equation}\label{eqn:slot_attention_value}
    \updates \coloneqq W^T \cdot v(\inp) \in \mathbb{R}^{K \times D}  \qquad \text{where} \qquad W_{i, j} \coloneqq \frac{\attn_{i, j}}{\sum_{l=1}^N\attn_{l, j}} \ .
\end{equation}
The weighted mean helps improve stability of the attention mechanism (compared to using a weighted sum) as in our case the attention coefficients are normalized over the slots. In practice we further add a small offset $\epsilon$ to the attention coefficients to avoid numerical instability.

The aggregated \texttt{updates} are finally used to update the slots via a learned recurrent function, for which we use a Gated Recurrent Unit (GRU)~\citep{cho2014learning} with $D_\slots$ hidden units. We found that transforming the GRU output with an (optional) multi-layer perceptron (MLP) with ReLU activation and a residual connection~\citep{he2016deep} can help improve performance. Both the GRU and the residual MLP are applied independently on each slot with shared parameters. We apply layer normalization (LayerNorm)~\citep{ba2016layer} both to the inputs of the module and to the slot features at the beginning of each iteration and before applying the residual MLP. While this is not strictly necessary, we found that it helps speed up training convergence. The overall time-complexity of the module is $\mathcal{O}\left(T\cdot D \cdot N\cdot K \right)$.

We identify two key properties of \sam: (1) permutation invariance with respect to the input (i.e., the output is independent of permutations applied to the input and hence suitable for sets) and (2) permutation equivariance with respect to the order of the slots (i.e., permuting the order of the slots after their initialization is equivalent to permuting the output of the module). More formally:
\begin{proposition}\label{thm:layer_invarianvce_equivariance}
\looseness=-1Let $\textnormal{SlotAttention}(\textnormal{\inp}, \textnormal{\slots})\in\mathbb{R}^{K\times D_{\textnormal{\slots}}}$ be the output of the Slot Attention module (Algorithm~\ref{algo:slot_attention}), where $\textnormal{\inp}\in\mathbb{R}^{N\times D_{\textnormal{\inp}}}$ and $\textnormal{\slots}\in\mathbb{R}^{K\times D_{\textnormal{\slots}}}$\,. Let $\pi_{i}\in\mathbb{R}^{N\times N}$ and $\pi_{s}\in\mathbb{R}^{K\times K}$ be arbitrary permutation matrices. Then, the following holds:
\begin{align*}
    \textnormal{SlotAttention}(\pi_i \cdot \textnormal{\inp}, \pi_s\cdot \textnormal{\slots}) = \pi_s \cdot \textnormal{SlotAttention}(\textnormal{\inp}, \textnormal{\slots})\,.
\end{align*}
\end{proposition}

The proof is in the supplementary material. The permutation equivariance property is important to ensure that slots learn a common representational format and that each slot can bind to any object in the input.

\subsection{Object Discovery}
\label{sec:object_discovery}
\looseness=-1Set-structured hidden representations are an attractive choice for learning about objects in an unsupervised fashion: each set element can capture the properties of an object in a scene, without assuming a particular order in which objects are described. Since \sam transforms input representations into a set of vectors, it can be used as part of the encoder in an autoencoder architecture for unsupervised object discovery. The autoencoder is tasked to encode an image into a set of hidden representations (i.e., slots) that, taken together, can be decoded back into the image space to reconstruct the original input. The slots thereby act as a representational bottleneck and the architecture of the decoder (or decoding process) is typically chosen such that each slot decodes only a region or part of the image~\citep{greff2016tagger,greff2017neural,eslami2016attend,burgess2019monet,greff2019multi,engelcke2019genesis}. These regions/parts are then combined to arrive at the full reconstructed image.

\textbf{Encoder} \ Our encoder consists of two components: (i) a CNN backbone augmented with positional embeddings, followed by (ii) a Slot Attention module. The output of \sam is a set of slots, that represent a grouping of the scene (e.g.~in terms of objects).

\textbf{Decoder} \ Each slot is decoded individually with the help of a spatial broadcast decoder~\citep{watters2019spatial}, as used in IODINE~\citep{greff2019multi}: slot representations are broadcasted onto a 2D grid (per slot) and augmented with position embeddings. Each such grid is decoded using a CNN (with parameters shared across the slots) to produce an output of size $W\times H\times 4$, where $W$ and $H$ are width and height of the image, respectively. The output channels encode RGB color channels and an (unnormalized) alpha mask. We subsequently normalize the alpha masks across slots using a $\texttt{Softmax}$ and use them as mixture weights to combine the individual reconstructions into a single RGB image.

\subsection{Set Prediction}
\label{sec:set_prediction}
Set representations are commonly used in tasks across many data modalities ranging from point cloud prediction~\citep{achlioptas2018learning,fan2017point}, classifying multiple objects in an image~\citep{zhang2019deep}, or generation of molecules with desired properties~\citep{de2018molgan,simonovsky2018graphvae}.
In the example considered in this paper, we are given an input image and a set of prediction targets, each describing an object in the scene. The key challenge in predicting sets is that there are $K!$ possible equivalent representations for a set of $K$ elements, as the order of the targets is arbitrary. This inductive bias needs to be explicitly modeled in the architecture to avoid discontinuities in the learning process, e.g.~when two semantically specialized slots swap their content throughout training~\citep{zhang2019deep,zhang2019fspool}.
The output order of \sam is random and independent of the input order, which addresses this issue. Therefore, \sam can be used to turn a distributed representation of an input scene into a set representation where each object can be separately classified with a standard classifier as shown in Figure~\ref{fig:slota_b}.

\textbf{Encoder} \ We use the same encoder architecture as in the object discovery setting (Section \ref{sec:object_discovery}), namely a CNN backbone augmented with positional embeddings, followed by \sam, to arrive at a set of slot representations.

\textbf{Classifier} \ For each slot, we apply a MLP with parameters shared between slots. As the order of both predictions and labels is arbitrary, we match them using the Hungarian algorithm~\citep{kuhn1955hungarian}. We leave the exploration of other matching algorithms~\citep{cuturi2013sinkhorn,zeng2019dmm} for future work.

\section{Related Work}\label{sec:related_work}
\paragraph{Object discovery}
Our object discovery architecture is closely related to a line of recent work on compositional generative scene models~\citep{greff2016tagger,eslami2016attend,greff2017neural,nash2017multi,van2018relational,kosiorek2018sequential,greff2019multi,burgess2019monet,engelcke2019genesis,stelzner2019faster,crawford2019spatially,jiang2019scalable,lin2020space} that represent a scene in terms of a collection of latent variables with the same representational format. Closest to our approach is the IODINE~\citep{greff2019multi} model, which uses iterative variational inference~\citep{marino2018iterative} to infer a set of latent variables, each describing an object in an image. In each inference iteration, IODINE performs a decoding step followed by a comparison in pixel space and a subsequent encoding step. Related models such as MONet~\citep{burgess2019monet} and GENESIS~\citep{engelcke2019genesis} similarly use multiple encode-decode steps. Our model instead replaces this procedure with a single encoding step using iterated attention, which improves computational efficiency. Further, this allows our architecture to infer object representations and attention masks even in the absence of a decoder, opening up extensions beyond auto-encoding, such as contrastive representation learning for object discovery \citep{kipf2019contrastive} or direct optimization of a downstream task like control or planning. Our attention-based routing procedure could also be employed in conjunction with patch-based decoders, used in architectures such as AIR~\citep{eslami2016attend}, SQAIR~\citep{kosiorek2018sequential}, and related approaches~\citep{stelzner2019faster,crawford2019spatially,jiang2019scalable,lin2020space}, as an alternative to the typically employed autoregressive encoder~\citep{eslami2016attend,kosiorek2018sequential}. Our approach is orthogonal to methods using adversarial training~\citep{van2018case,chen2019unsupervised,yang2020learning} or contrastive learning~\citep{kipf2019contrastive} for object discovery: utilizing Slot Attention in such a setting is an interesting avenue for future work.

\textbf{Neural networks for sets} \
\looseness=-1A range of recent methods explore set encoding~\citep{lin2017structured,zaheer2017deep,zhang2019fspool}, generation~\citep{zhang2019deep,rezatofighi2020learn}, and set-to-set mappings~\citep{vaswani2017attention,lee2018set}. Graph neural networks~\citep{scarselli2008graph,li2015gated,kipf2016semi,battaglia2018relational} and in particular the self-attention mechanism of the Transformer model~\citep{vaswani2017attention} are frequently used to transform sets of elements with constant cardinality (i.e., number of set elements). Slot Attention addresses the problem of mapping from one set to another set of different cardinality while respecting permutation symmetry of both the input and the output set. The Deep Set Prediction Network (DSPN)~\citep{zhang2019deep,huang2020set} respects permutation symmetry by running an inner gradient descent loop for each example, which requires many steps for convergence and careful tuning of several loss hyperparmeters. Instead, \sam directly maps from set to set using only a few attention iterations and a single task-specific loss function. In concurrent work, both the DETR~\citep{carion2020detr} and the TSPN~\citep{kosiorek2020conditional} model propose to use a Transformer~\citep{vaswani2017attention} for conditional set generation. Most related approaches, including DiffPool~\citep{ying2018hierarchical}, Set Transformers~\citep{lee2018set}, DSPN~\citep{zhang2019deep}, and DETR~\citep{carion2020detr} use a learned per-element initialization (i.e., separate parameters for each set element), which prevents these approaches from generalizing to more set elements at test time.

\textbf{Iterative routing} \
Our iterative attention mechanism shares similarlities with iterative \textit{routing} mechanisms typically employed in variants of Capsule Networks~\citep{sabour2017dynamic, hinton2018matrix,tsai2020capsules}. The closest such variant is inverted dot-product attention routing~\citep{tsai2020capsules} which similarly uses a dot product attention mechanism to obtain assignment coefficients between representations. Their method (in line with other capsule models) however does not have permutation symmetry as each input-output pair is assigned a separately parameterized transformation. The low-level details in how the attention mechanism is normalized and how updates are aggregated, and the considered applications also differ significantly between the two approaches.

\textbf{Interacting memory models} \ Slot Attention can be seen as a variant of interacting memory models~\citep{watters2017visual,van2018relational,santoro2018relational,zambaldi2018relational,watters2019cobra,stanic2019r,goyal2019recurrent,kipf2019contrastive,veerapaneni2020entity}, which utilize a set of slots and their pairwise interactions to reason about elements in the input (e.g.~objects in a video). Common components of these models are (i) a recurrent update function that acts independently on individual slots and (ii) an interaction function that introduces communication between slots. Typically, slots in these models are fully symmetric with shared recurrent update functions and interaction functions for all slots, with the exception of the RIM model~\citep{goyal2019recurrent}, which uses a separate set of parameters for each slot. Notably, RMC~\citep{santoro2018relational} and RIM~\citep{goyal2019recurrent} introduce an attention mechanism to aggregate information from inputs to slots. In Slot Attention, the attention-based assignment from inputs to slots is normalized over the slots (as opposed to solely over the inputs), which introduces competition between the slots to perform a clustering of the input. Further, we do not consider temporal data in this work and instead use the recurrent update function to iteratively refine predictions for a single, static input.

\looseness=-1\textbf{Mixtures of experts} \ Expert models~\citep{jacobs1991adaptive,parascandolo2018learning,locatello2018competitive,goyal2019recurrent,von2020towards} are related to our slot-based approach, but do not fully share parameters between individual experts. This results in the specialization of individual experts to, e.g., different tasks or object types. In Slot Attention, slots use a common representational format and each slot can bind to any part of the input.

\textbf{Soft clustering} \
\looseness=-1Our routing procedure is related to soft k-means clustering~\cite{bauckhage2015lecture} (where slots corresponds to cluster centroids) with two key differences: We use a dot product similarity with learned linear projections and we use a parameterized, learnable update function. Variants of soft k-means clustering with learnable, cluster-specific parameters have been introduced in the computer vision~\citep{arandjelovic2016netvlad} and speech recognition communities~\citep{cai2018novel}, but they differ from our approach in that they do not use a recurrent, multi-step update, and do not respect permutation symmetry (cluster centers act as a fixed, ordered dictionary after training). The inducing point mechanism of the Set Transformer~\citep{lee2018set} and the image-to-slot attention mechanism in DETR~\citep{carion2020detr} can be seen as extensions of these ordered, single-step approaches using multiple attention heads (i.e., multiple similarity functions) for each cluster assignment.

\textbf{Recurrent attention} \
Our method is related to recurrent attention models used in image modeling and scene decomposition~\citep{mnih2014recurrent,gregor2015draw,eslami2016attend,ren2017end,kosiorek2018sequential}, and for set prediction~\citep{welleck2017saliency}. Recurrent models for set prediction have also been considered in this context without using attention mechanisms~\citep{stewart2016end,romera2016recurrent}. This line of work frequently uses permutation-invariant loss functions~\citep{stewart2016end,welleck2017saliency,welleck2018loss}, but relies on inferring one slot, representation, or label per time step in an auto-regressive manner, whereas Slot Attention updates all slots simultaneously at each step, hence fully respecting permutation symmetry.

\section{Experiments}\label{sec:experiments}
The goal of this section is to evaluate the Slot Attention module on two object-centric tasks---one being supervised and the other one being unsupervised---as described in Sections~\ref{sec:object_discovery} and~\ref{sec:set_prediction}. We compare against specialized state-of-the-art methods \citep{zhang2019deep,greff2019multi,burgess2019monet} for each respective task. We provide further details on experiments and implementation, and additional qualitative results and ablation studies in the supplementary material.

\looseness=-1\textbf{Baselines} \
In the unsupervised object discovery experiments, we compare against two recent state-of-the-art models: IODINE~\citep{greff2019multi} and MONet~\citep{burgess2019monet}.
For supervised object property prediction, we compare against Deep Set Prediction Networks (DSPN)~\citep{zhang2019deep}. DSPN is the only set prediction model that respects permutation symmetry that we are aware of, other than our proposed model.
In both tasks, we further compare against a simple MLP-based baseline that we term Slot MLP. This model replaces Slot Attention with an MLP that maps from the CNN feature maps (resized and flattened) to the (now ordered) slot representation. For the MONet, IODINE, and DSPN baselines, we compare with the published numbers in~\citep{greff2019multi,zhang2019deep} as we use the same experimental setup.

\textbf{Datasets} \
For the object discovery experiments, we use the following three multi-object datasets \citep{multiobjectdatasets19}: CLEVR (with masks), Multi-dSprites, and Tetrominoes. CLEVR (with masks) is a version of the CLEVR dataset with segmentation mask annotations. Similar to IODINE~\citep{greff2019multi}, we only use the first 70K samples from the CLEVR (with masks) dataset for training and we crop images to highlight objects in the center. For Multi-dSprites and Tetrominoes, we use the first 60K samples. As in~\citep{greff2019multi}, we evaluate on 320 test examples for object discovery.
For set prediction, we use the original CLEVR dataset~\citep{johnson2017clevr} which contains a training-validation split of 70K and 15K images of rendered objects respectively. Each image can contain between three and ten objects and has property annotations for each object (position, shape, material, color, and size).
In some experiments, we filter the CLEVR dataset to contain only scenes with at maximum 6 objects; we call this dataset CLEVR6 and we refer to the original full dataset as CLEVR10 for clarity.

\subsection{Object Discovery}\label{sec:exp_obj_disc}
\textbf{Training} \
The training setup is unsupervised: the learning signal is provided by the (mean squared) image reconstruction error.
We train the model using the Adam optimizer~\citep{kingma2014adam} with a learning rate of $4\times10^{-4}$ and a batch size of 64 (using a single GPU). We further make use of learning rate warmup~\citep{goyal2017accurate} to prevent early saturation of the attention mechanism and an exponential decay schedule in the learning rate, which we found to reduce variance. At training time, we use $T=3$ iterations of Slot Attention. We use the same training setting across all datasets, apart from the number of slots $K$: we use $K=7$ slots for CLEVR6, $K=6$ slots for Multi-dSprites (max.~5 objects per scene), and $K=4$ for Tetrominoes (3 objects per scene). Even though the number of slots in Slot Attention can be set to a different value for each input example, we use the same value $K$ for all examples in the training set to allow for easier batching.

\textbf{Metrics} \
In line with previous works~\citep{greff2019multi,burgess2019monet}, we compare the alpha masks produced by the decoder (for each individual object slot) with the ground truth segmentation (excluding the background) using the Adjusted Rand Index (ARI) score~\citep{rand1971objective,hubert1985comparing}. ARI is a score to measure clustering similarity, ranging from 0 (random) to 1 (perfect match). To compute the ARI score, we use the implementation provided by \citet{multiobjectdatasets19}.

\begin{figure}[t]
\centering
\begin{minipage}{0.58\textwidth}
\resizebox{\textwidth}{!}{
\begin{tabular}{l c c c c c }
\toprule
& CLEVR6 & Multi-dSprites & Tetrominoes \\
\midrule
\sam & $\mathbf{98.8 \pm 0.3}$ & $\mathbf{91.3 \pm 0.3}$ & $\mathbf{99.5 \pm 0.2}$*\\
IODINE~\citep{greff2019multi} & $\mathbf{98.8 \pm 0.0}$ & $76.7 \pm 5.6$ & $\mathbf{99.2 \pm 0.4}$\\
MONet~\citep{burgess2019monet} & $96.2 \pm 0.6$ & $\mathbf{90.4 \pm 0.8}$ & ---\\
Slot MLP & $60.4 \pm 6.6$ & $60.3 \pm 1.8$ & $25.1 \pm 34.3$\\
\bottomrule
\end{tabular}
\captionlistentry[table]{}
\label{table:ari_object_discovery}
}
\end{minipage}
~\quad
\begin{minipage}{0.32\textwidth}
\includegraphics[width=\textwidth,trim={0 0.4cm 0 0.2cm},clip]{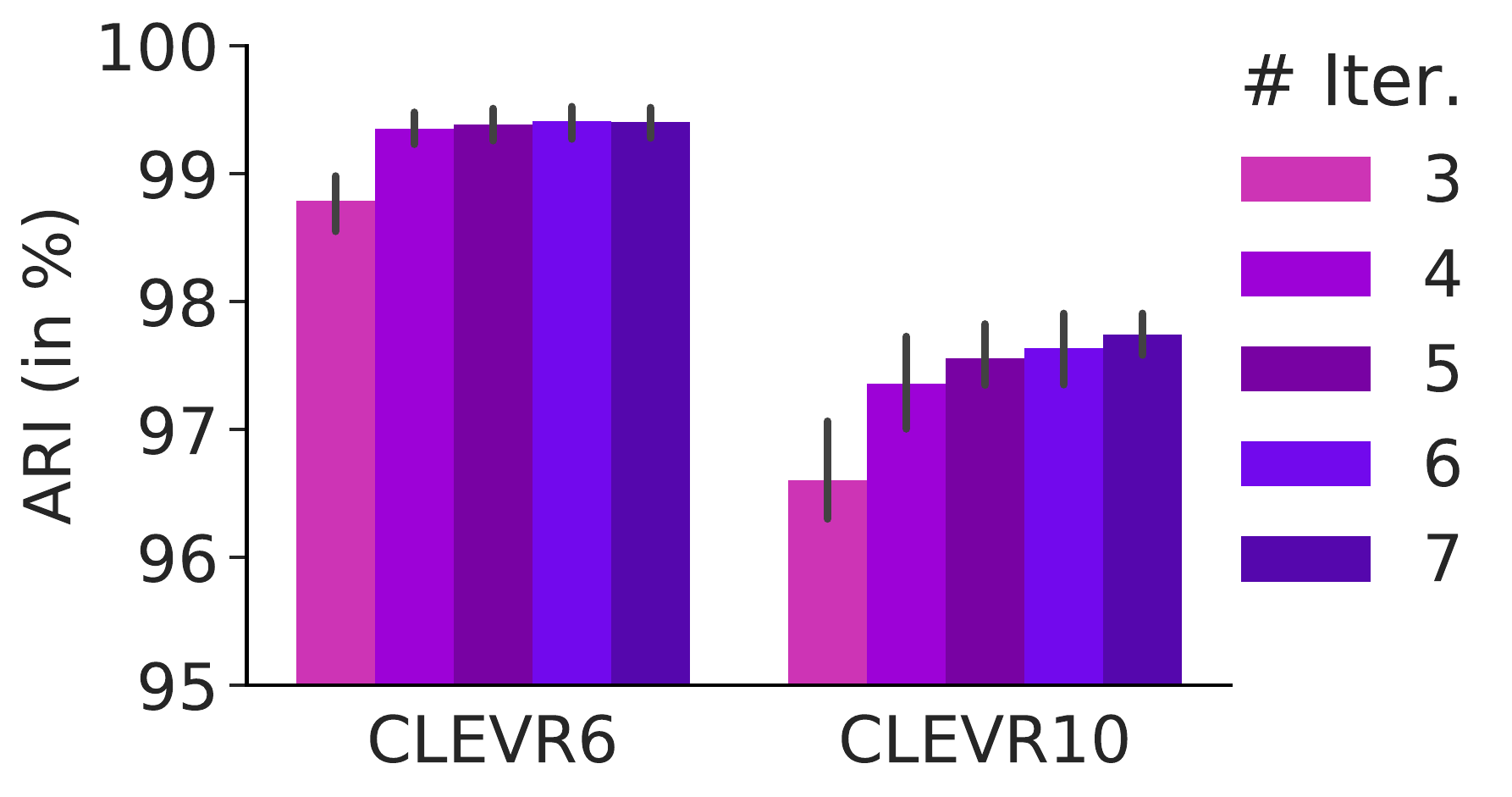}
\end{minipage}
\captionsetup{labelformat=andtable}
\caption{(\textbf{Left}) Adjusted Rand Index (ARI) scores (in $\%$, mean $\pm$ stddev for 5 seeds) for unsupervised object discovery in multi-object datasets. In line with previous works~\citep{greff2019multi,burgess2019monet,engelcke2019genesis}, we exclude background labels in ARI evaluation. *denotes that one outlier was excluded from evaluation. (\textbf{Right}) Effect of increasing the number of Slot Attention iterations $T$ at test time (for a model trained on CLEVR6 with $T=3$ and $K=7$ slots), tested on CLEVR6 ($K=7$) and CLEVR10 ($K=11$).}
\label{fig:ari_object_discovery_iters}
\vspace{-3mm}
\end{figure}

\looseness=-1\textbf{Results} \
Quantitative results are summarized in Table \ref{table:ari_object_discovery} and Figure \ref{fig:ari_object_discovery_iters}. In general, we observe that our model compares favorably against two recent state-of-the-art baselines: IODINE~\citep{greff2019multi} and MONet~\citep{burgess2019monet}. We also compare against a simple MLP-based baseline (Slot MLP) which performs better than chance, but due to its ordered representation is unable to model the compositional nature of this task. We note a failure mode of our model: In rare cases it can get stuck in a suboptimal solution on the Tetrominoes dataset, where it segments the image into stripes. This leads to a significantly higher reconstruction error on the training set, and hence such an outlier can easily be identified at training time. We excluded a single such outlier (1 out of 5 seeds) from the final score in Table \ref{table:ari_object_discovery}. We expect that careful tuning of the training hyperparameters particularly for this dataset could alleviate this issue, but we opted for a single setting shared across all datasets for simplicity.

Compared to IODINE~\citep{greff2019multi}, Slot Attention is significantly more efficient in terms of both memory consumption and runtime. On CLEVR6, we can use a batch size of up to 64 on a single V100 GPU with 16GB of RAM as opposed to 4 in~\citep{greff2019multi} using the same type of hardware. Similarly, when using 8 V100 GPUs in parallel, model training on CLEVR6 takes approximately 24hrs for Slot Attention as opposed to approximately 7 days for IODINE~\citep{greff2019multi}.

\begin{figure}[t]
    \centering
    \begin{subfigure}[b]{0.61\textwidth}
    \includegraphics[width=\textwidth,trim={0 0.35cm 0 0.25cm},clip]{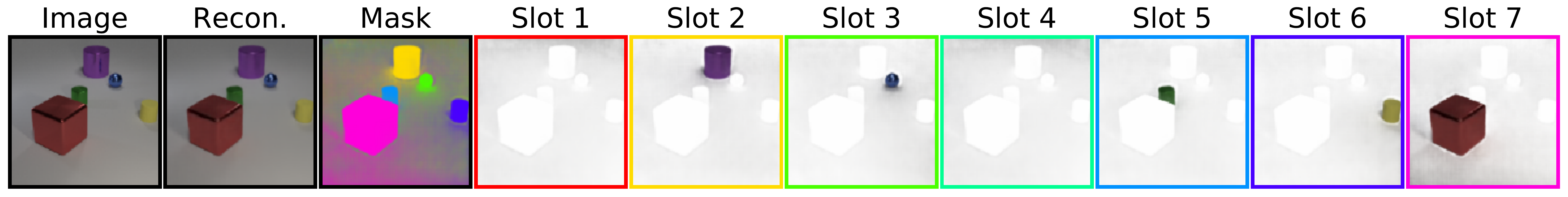}
    \includegraphics[width=\textwidth,trim={0 0.35cm 0 0.35cm},clip]{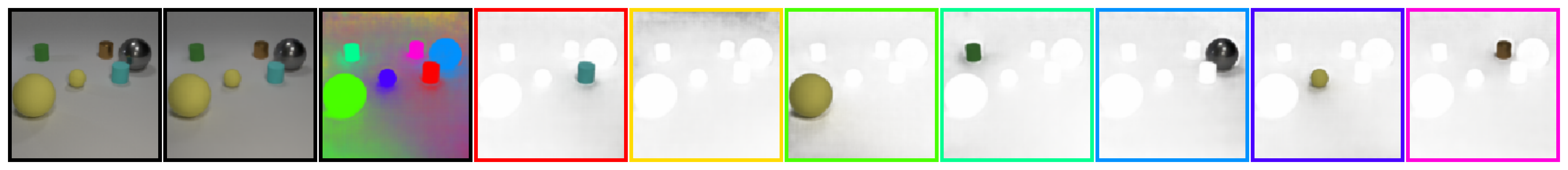}
    \includegraphics[width=0.9\textwidth,trim={0 0.35cm 0 0.35cm},clip]{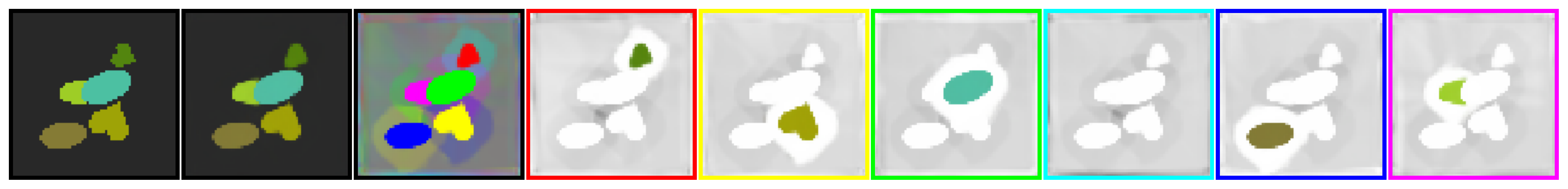}
    \includegraphics[width=0.9\textwidth,trim={0 0.35cm 0 0.35cm},clip]{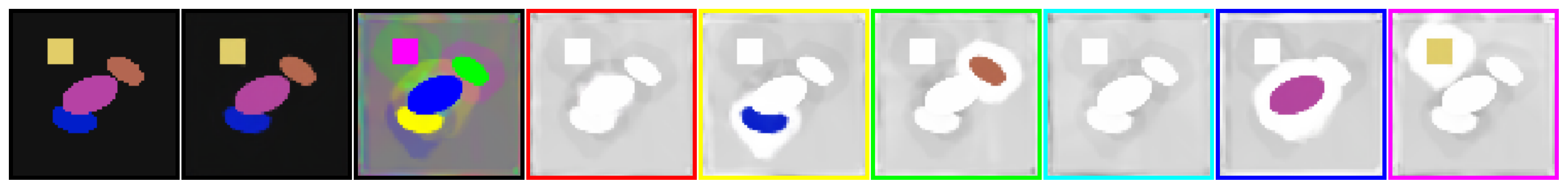}
    \includegraphics[width=0.7\textwidth,trim={0 0.35cm 0 0.35cm},clip]{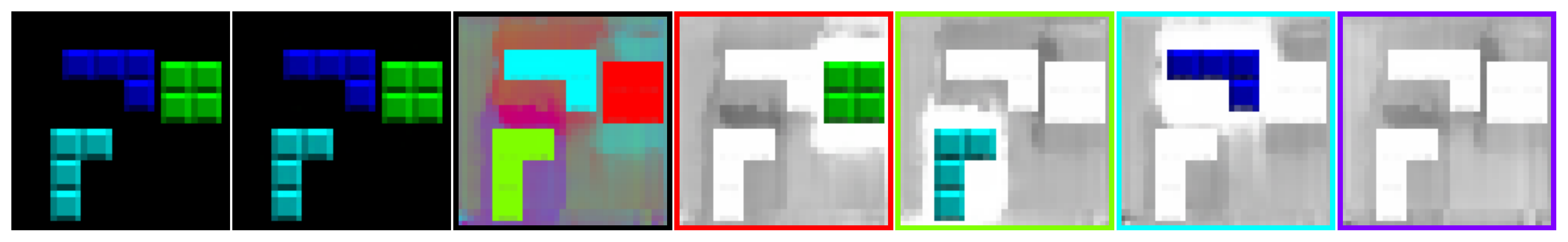}
    \includegraphics[width=0.7\textwidth,trim={0 0.35cm 0 0.35cm},clip]{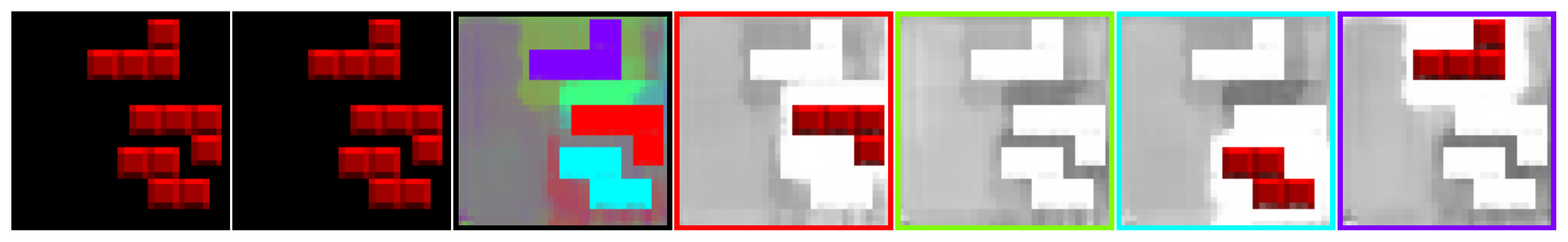}
    \caption{Decomposition across datasets.}
    \end{subfigure}
    ~
    \begin{subfigure}[b]{0.37\textwidth}
    {
        \centering
        \includegraphics[width=0.84\textwidth,trim={0 0.35cm 0 0.25cm},clip]{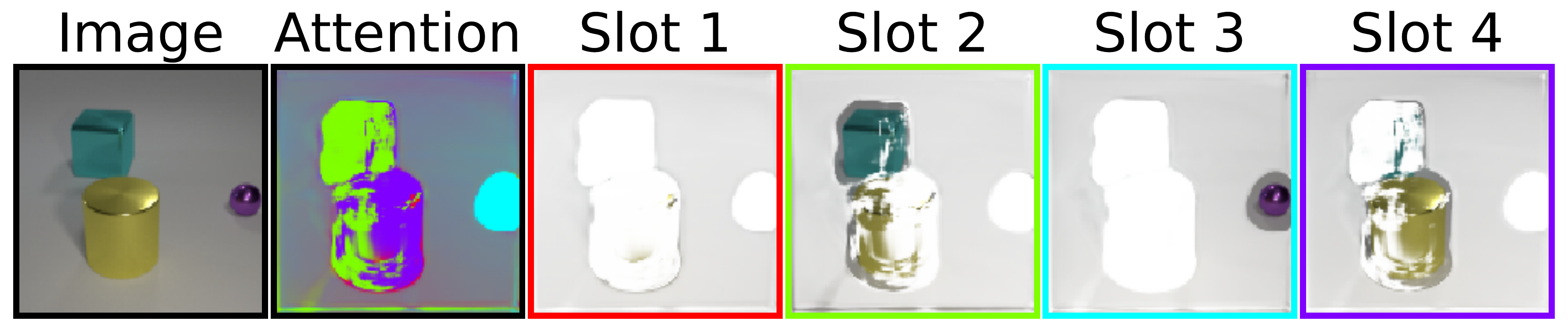}
        \includegraphics[width=0.84\textwidth,trim={0 0.35cm 0 0.35cm},clip]{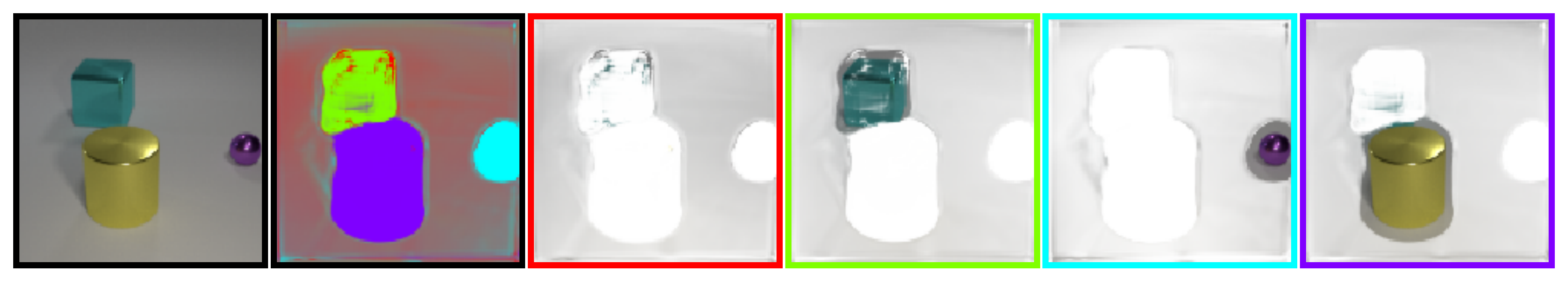}
        \includegraphics[width=0.84\textwidth,trim={0 0.35cm 0 0.35cm},clip]{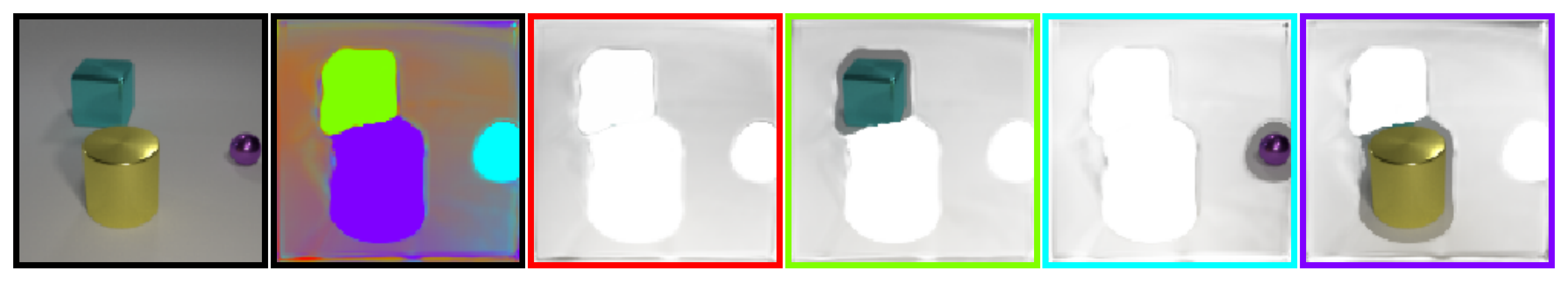}
        \caption{Attention iterations.}
    }\vspace{0.1em}
    {
        \centering
        \includegraphics[width=\textwidth,trim={0 0.35cm 0 0.25cm},clip]{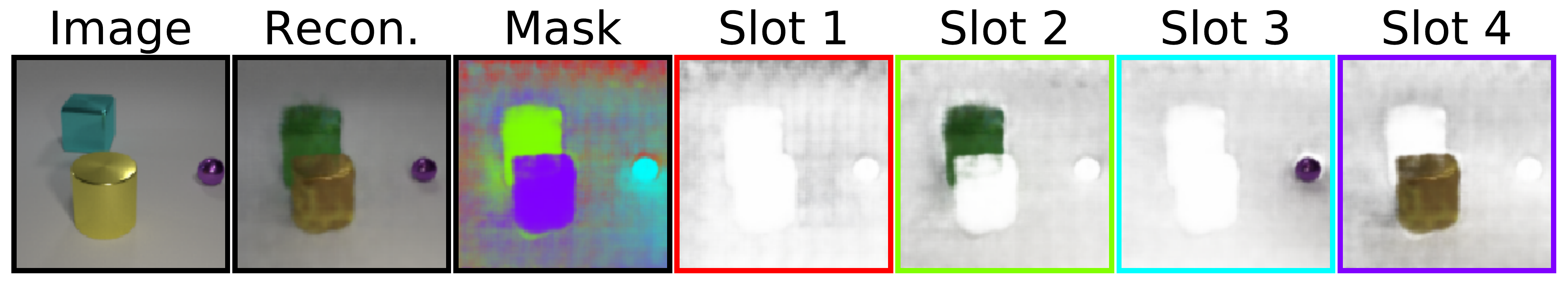}
        \includegraphics[width=\textwidth,trim={0 0.35cm 0 0.35cm},clip]{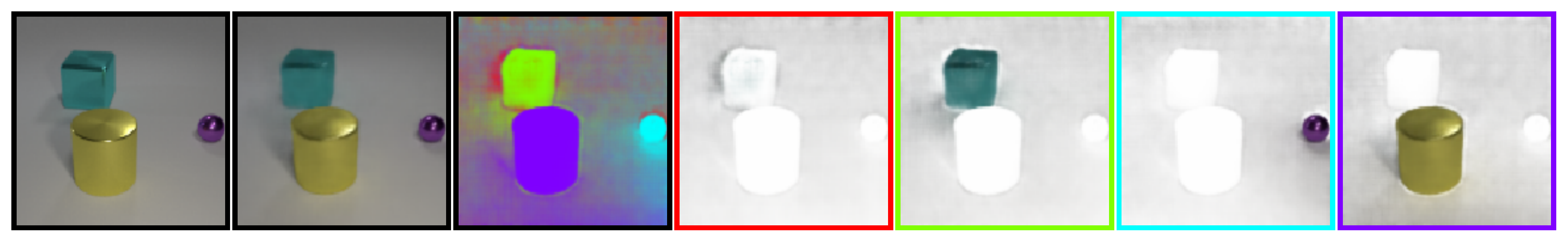}
        \includegraphics[width=\textwidth,trim={0 0.35cm 0 0.35cm},clip]{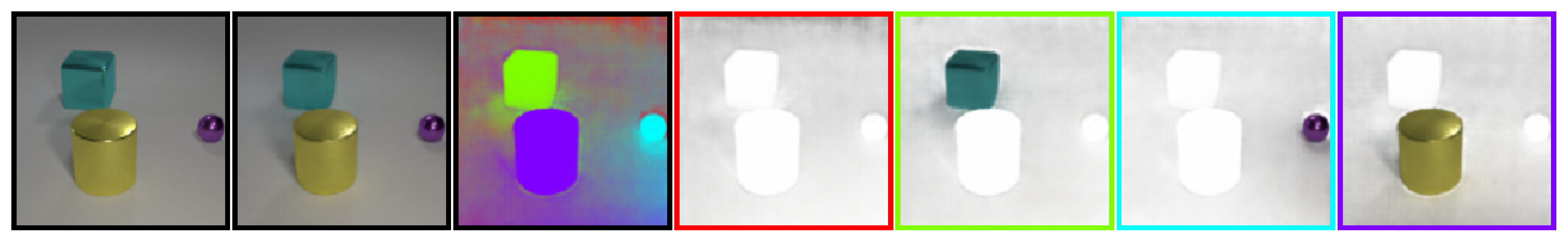}
        \caption{Reconstructions per iteration.}
    }
    \end{subfigure}
    \caption{(\textbf{a}) Visualization of per-slot reconstructions and alpha masks in the unsupervised training setting (object discovery). Top rows: CLEVR6, middle rows: Multi-dSprites, bottom rows: Tetrominoes. (\textbf{b}) Attention masks (\texttt{attn}) for each iteration, only using four object slots at test time on CLEVR6. (\textbf{c}) Per-iteration reconstructions and reconstruction masks (from decoder). Border colors for slots correspond to colors of segmentation masks used in the combined mask visualization (third column). We visualize individual slot reconstructions multiplied with their respective alpha mask, using the visualization script from~\citep{greff2019multi}.}
    \label{fig:masks_obj_disc}
\end{figure}
\begin{figure}[t]
    \centering
    \includegraphics[width=0.7\textwidth,trim={0 0.3cm 0 0.25cm},clip]{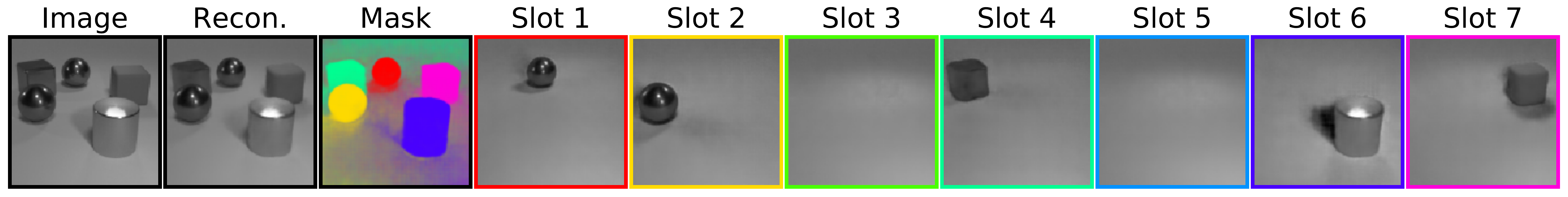}
    \includegraphics[width=0.7\textwidth,trim={0 0.3cm 0 0.4cm},clip]{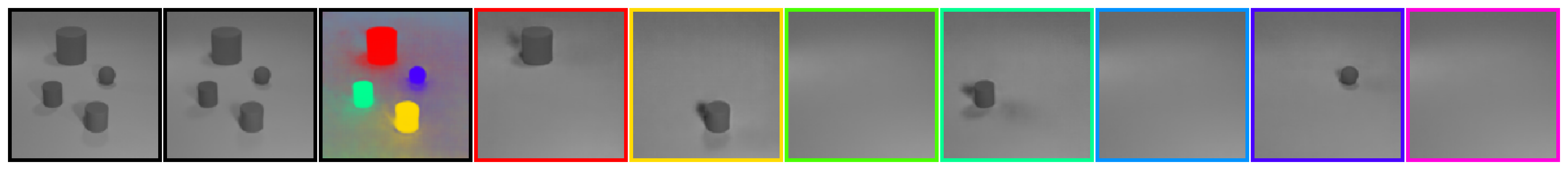}
    \caption{Visualization of (per-slot) reconstructions and masks of a Slot Attention model trained on a greyscale version of CLEVR6, where it achieves $98.5\pm0.3\%$ ARI. Here, we show the full reconstruction of each slot (i.e., without multiplication with their respective alpha mask).}
    \label{fig:greyscale_clevr}
\end{figure}

In Figure \ref{fig:ari_object_discovery_iters}, we investigate to what degree our model generalizes when using more Slot Attention iterations at test time, while being trained with a fixed number of $T=3$ iterations. We further evaluate generalization to more objects (CLEVR10) compared to the training set (CLEVR6). We observe that segmentation scores significantly improve beyond the numbers reported in Table \ref{table:ari_object_discovery} when using more iterations. This improvement is stronger when testing on CLEVR10 scenes with more objects. For this experiment, we increase the number of slots from $K=7$ (training) to $K=11$ at test time. Overall, segmentation performance remains strong even when testing on scenes that contain more objects than seen during training.

We visualize discovered object segmentations in Figure~\ref{fig:masks_obj_disc} for all three datasets. The model learns to keep slots empty (only capturing the background) if there are more slots than objects. We find that Slot Attention typically spreads the uniform background across all slots instead of capturing it in just a single slot, which is likely an artifact of the attention mechanism that does not harm object disentanglement or reconstruction quality. We further visualize how the attention mechanism segments the scene over the individual attention iterations, and we inspect scene reconstructions from each individual iteration (the model has been trained to reconstruct only after the final iteration). It can be seen that the attention mechanism learns to specialize on the extraction of individual objects already at the second iteration, whereas the attention map of the first iteration still maps parts of multiple objects into a single slot.

To evaluate whether Slot Attention can perform segmentation without relying on color cues, we further run experiments on a binarized version of multi-dSprites with white objects on black background, and on a greyscale version of CLEVR6. We use the binarized multi-dSprites dataset from~\citet{multiobjectdatasets19}, for which Slot Attention achieves $69.4 \pm 0.9\%$ ARI using $K=4$ slots, compared to $64.8 \pm 17.2\%$ for IODINE~\citep{greff2019multi} and $68.5 \pm 1.7\%$ for R-NEM~\citep{van2018relational}, as reported in~\citep{greff2019multi}. Slot Attention performs competitively in decomposing scenes into objects based on shape cues only.
We visualize discovered object segmentations for the Slot Attention model trained on greyscale CLEVR6 in Figure~\ref{fig:greyscale_clevr}, which Slot Attention handles without issue despite the lack of object color as a distinguishing feature.

As our object discovery architecture uses the same decoder and reconstruction loss as IODINE~\citep{greff2019multi}, we expect it to similarly struggle with scenes containing more complicated backgrounds and textures. Utilizing different perceptual~\citep{goodfellow2014generative,yang2020learning} or contrastive losses~\citep{kipf2019contrastive} could help overcome this limitation. We discuss further limitations and future work in Section~\ref{sec:conclusions} and in the supplementary material.

\textbf{Summary} \ Slot Attention is highly competitive with prior approaches on unsupervised scene decomposition, both in terms of quality of object segmentation and in terms of training speed and memory efficiency. At test time, Slot Attention can be used without a decoder to obtain object-centric representations from an unseen scene.

\subsection{Set Prediction}\label{sec:exp_set_pred}
\begin{figure}[t]
    \centering
    \begin{subfigure}[b]{0.29\textwidth}
    \includegraphics[width=\textwidth]{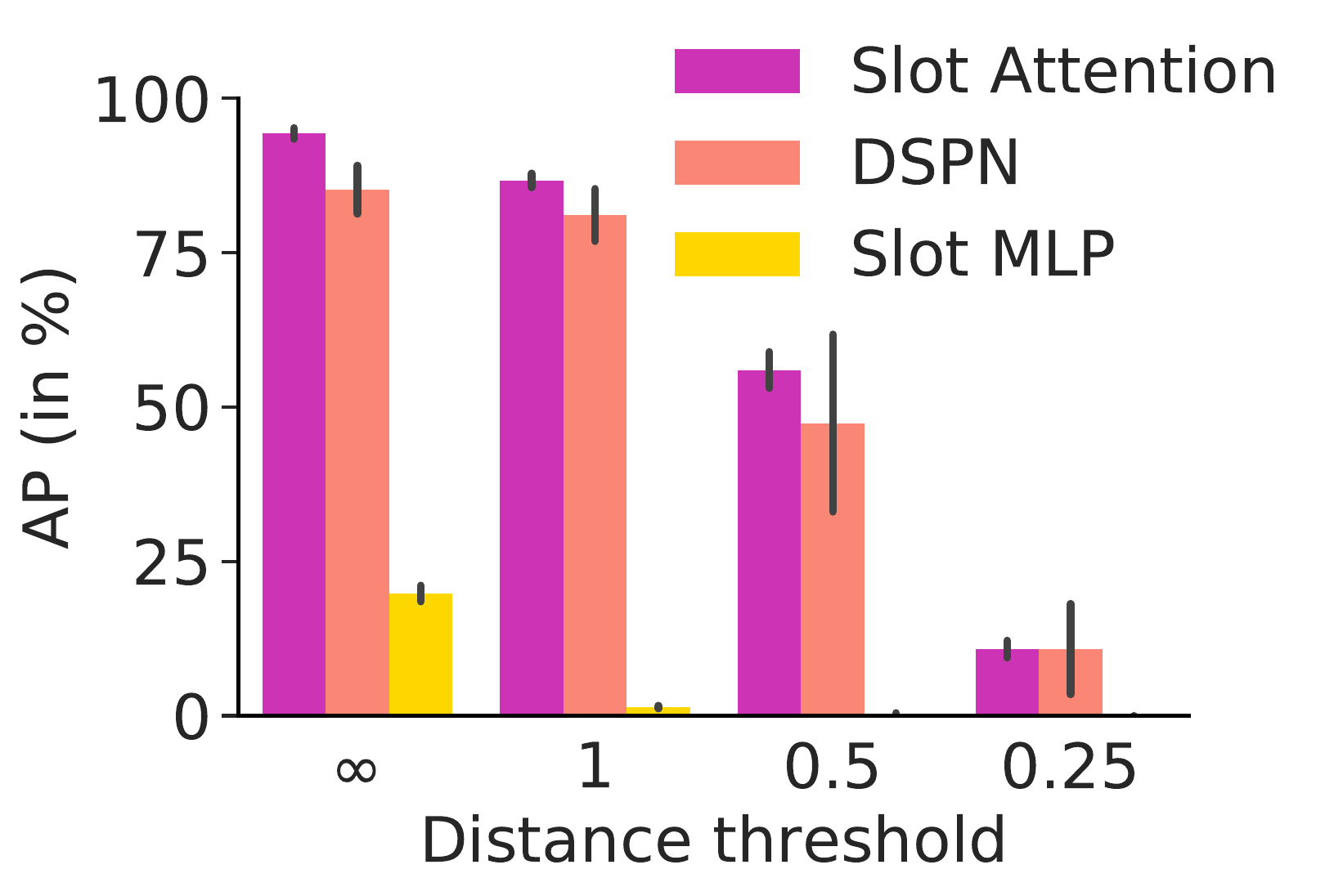}
    \end{subfigure}
    ~\quad
    \begin{subfigure}[b]{0.31\textwidth}
    \includegraphics[width=\textwidth]{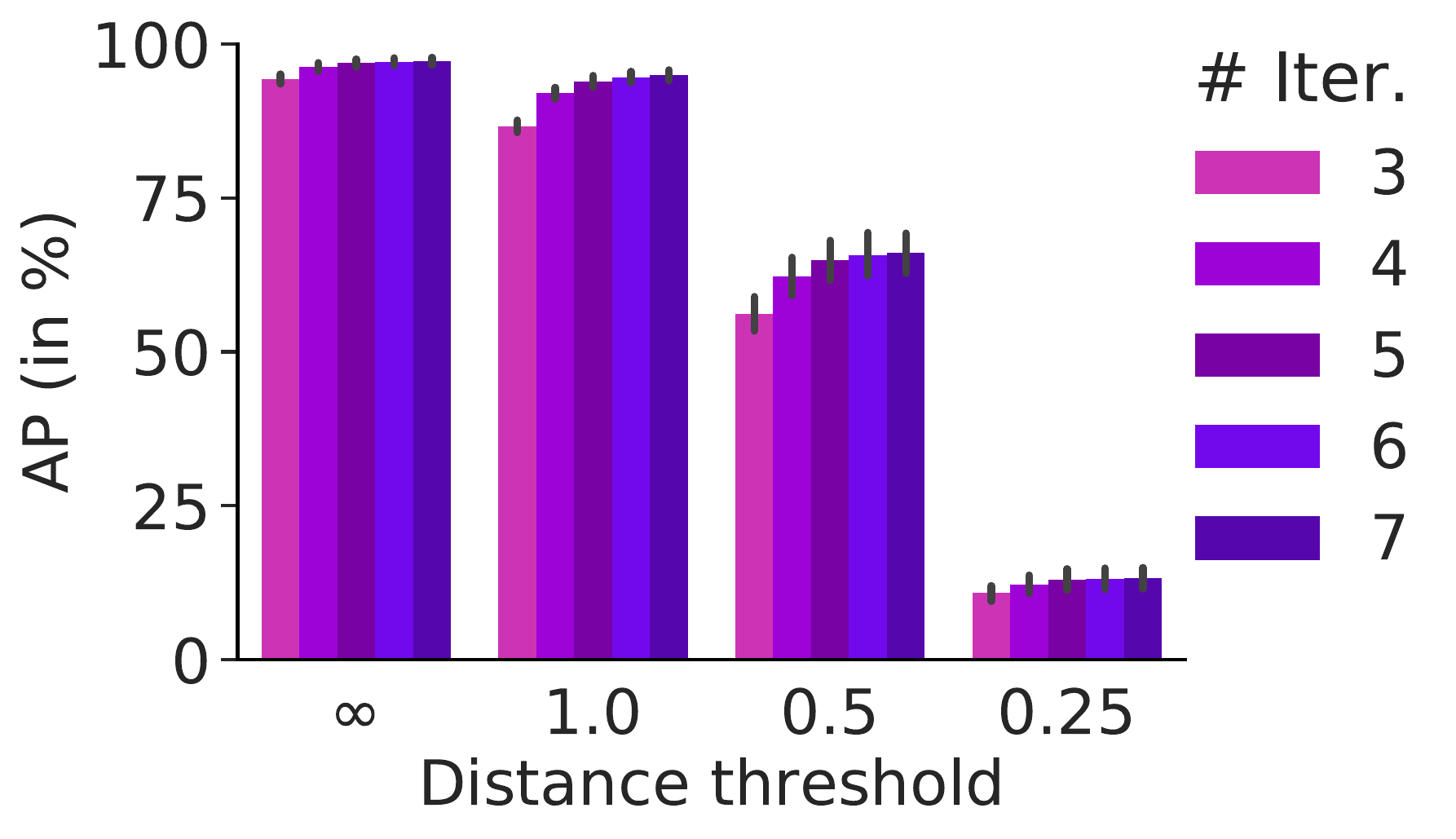}
    \end{subfigure}
    ~\quad
    \begin{subfigure}[b]{0.31\textwidth}
    \includegraphics[width=\textwidth]{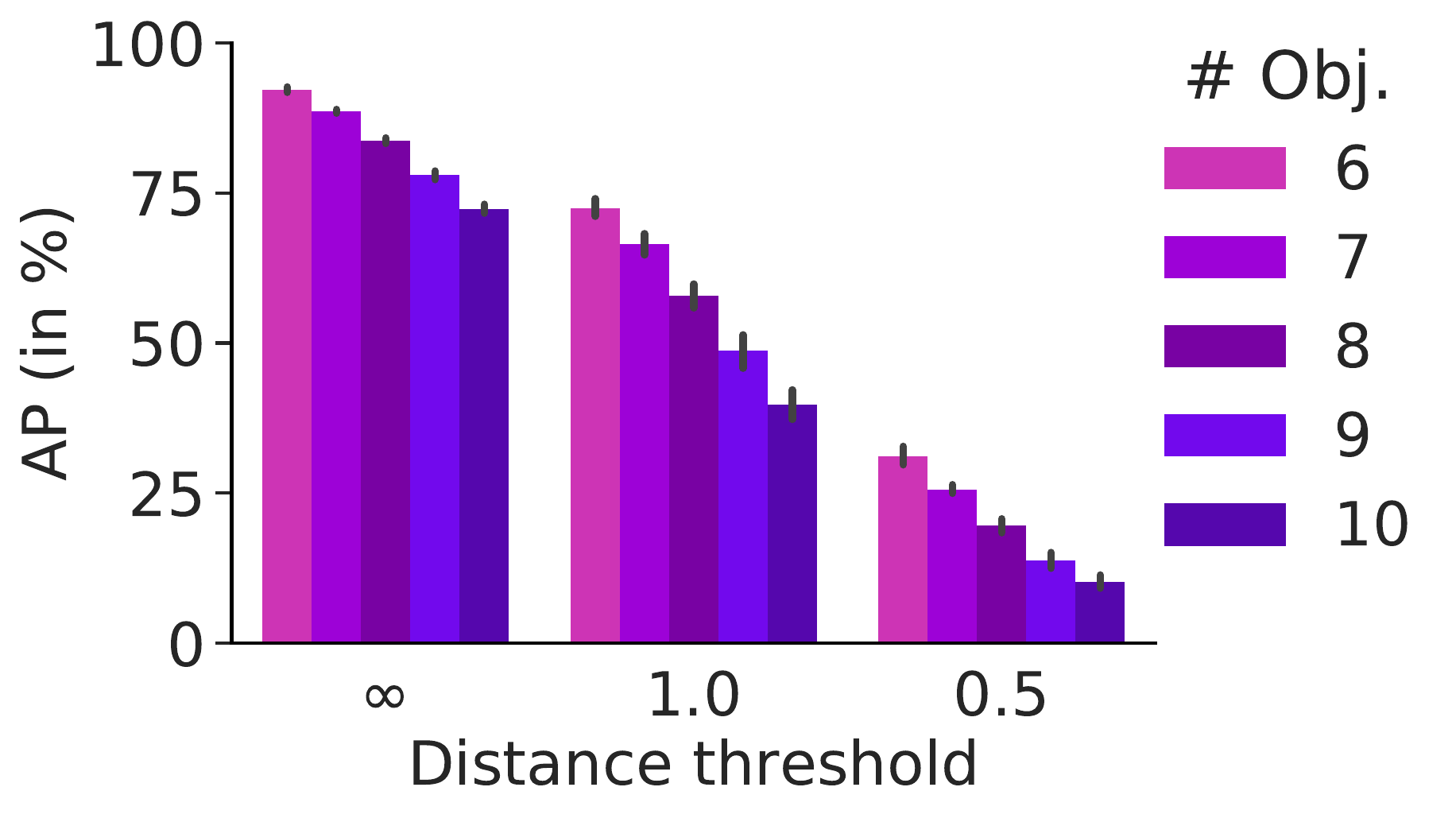}
    \end{subfigure}
    \vspace{-1mm}
    \caption{(\textbf{Left}) AP at different distance thresholds on CLEVR10 (with $K=10$).
    (\textbf{Center}) AP for the Slot Attention model with different number of iterations. The models are trained with 3 iterations and tested with iterations ranging from 3 to 7. (\textbf{Right}) AP for Slot Attention trained on CLEVR6 ($K=6$) and tested on scenes containing exactly $N$ objects (with $N=K$ from $6$ to $10$).}
    \label{fig:prop_pred}
\end{figure}

\begin{figure}[t]
    \centering
    \includegraphics[width=0.75\textwidth,trim={0 0.3cm 0 0.25cm},clip]{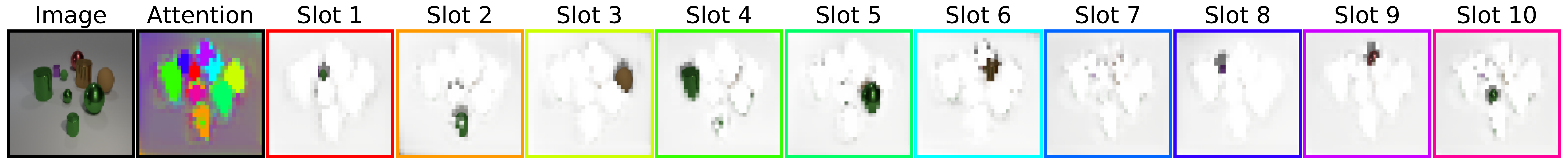}
    \includegraphics[width=0.75\textwidth,trim={0 0.3cm 0 1.3cm},clip]{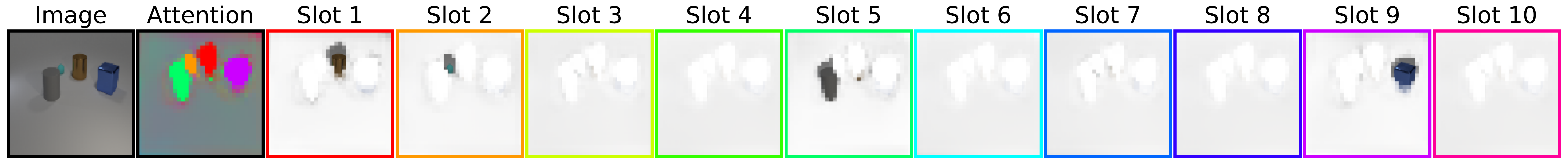}
    \caption{Visualization of the attention masks on CLEVR10 for two examples with 9 and 4 objects, respectively, for a model trained on the property prediction task. The masks are upsampled to $128\times 128$ for this visualization to match the resolution of input image.}
    \label{fig:masks_prop_pred}
\end{figure}

\textbf{Training} \
We train our model using the same hyperparameters as in Section~\ref{sec:exp_obj_disc} except we use a batch size of $512$ and striding in the encoder. On CLEVR10, we use $K=10$ object slots to be in line with~\citep{zhang2019deep}. The Slot Attention model is trained using a single NVIDIA Tesla V100 GPU with 16GB of RAM.

\looseness=-1\textbf{Metrics} \
Following~\citet{zhang2019deep}, we compute the Average Precision (AP) as commonly used in object detection~\citep{everingham2015pascal}. A prediction (object properties and position) is considered correct if there is a matching object with exactly the same properties (shape, material, color, and size) within a certain distance threshold ($\infty$ means we do not enforce any threshold).
The predicted position coordinates are scaled to $[-3, 3]$. We zero-pad the targets and predict an additional indicator score in $[0, 1]$ corresponding to the presence probability of an object (1 means there is an object) which we then use as prediction confidence to compute the AP.

\textbf{Results} \
In Figure~\ref{fig:prop_pred} (left) we report results in terms of Average Precision for supervised object property prediction on CLEVR10 (using $T=3$ for Slot Attention at both train and test time). We compare to both the DSPN results of~\citep{zhang2019deep} and the Slot MLP baseline. Overall, we observe that our approach matches or outperforms the DSPN baseline. The performance of our method degrades gracefully at more challenging distance thresholds (for the object position feature) maintaining a reasonably small variance. Note that the DSPN baseline~\citep{zhang2019deep} uses a significantly deeper ResNet 34~\citep{he2016deep} image encoder. In Figure~\ref{fig:prop_pred} (center) we observe that increasing the number of attention iterations at test time generally improves performance. Slot Attention  can naturally handle more objects at test time by changing the number of slots. In Figure~\ref{fig:prop_pred} (right) we observe that the AP degrades gracefully if we train a model on CLEVR6 (with $K=6$ slots) and test it with more objects.

Intuitively, to solve this set prediction task each slot should attend to a different object. In Figure~\ref{fig:masks_prop_pred}, we visualize the attention maps of each slot for two CLEVR images. In general, we observe that the attention maps naturally segment the objects. We remark that the method is only trained to predict the property of the objects, without any segmentation mask. Quantitatively, we can evaluate the Adjusted Rand Index (ARI) scores of the attention masks. On CLEVR10 (with masks), the attention masks produced by Slot Attention achieve an ARI of $78.0\% \pm 2.9$ (to compute the ARI we downscale the input image to $32 \times 32$). Note that the masks evaluated in Table~\ref{table:ari_object_discovery} are not the attention maps but are predicted by the object discovery decoder.

\textbf{Summary} \ Slot Attention learns a representation of objects for set-structured property prediction tasks and achieves results competitive with a prior state-of-the-art approach while being significantly easier to implement and tune. Further, the attention masks naturally segment the scene, which can be valuable for debugging and interpreting the predictions of the model.

\section{Conclusion}\label{sec:conclusions}
\looseness=-1We have presented the Slot Attention module, a versatile architectural component that learns object-centric abstract representations from low-level perceptual input. The iterative attention mechanism used in Slot Attention allows our model to learn a grouping strategy to decompose input features into a set of slot representations. In experiments on unsupervised visual scene decomposition and supervised object property prediction we have shown that Slot Attention is highly competitive with prior related approaches, while being more efficient in terms of memory consumption and computation.

A natural next step is to apply \sam to video data or to other data modalities, e.g.~for clustering of nodes in graphs, on top of a point cloud processing backbone or for textual or speech data. It is also promising to investigate other downstream tasks, such as reward prediction, visual reasoning, control, or planning.

\section*{Broader Impact}
\looseness=-1The Slot Attention module allows to learn object-centric representations from perceptual input. As such, it is a general module that can be used in a wide range of domains and applications. In our paper, we only consider artificially generated datasets under well-controlled settings where slots are expected to specialize to objects. However, the specialization of our model is implicit and fully driven by the downstream task.
We remark that as a concrete measure to assess whether the module specialized in unwanted ways, one can visualize the attention masks to understand how the input features are distributed across the slots (see Figure~\ref{fig:masks_prop_pred}). While more work is required to properly address the usefulness of the attention coefficients in explaining the overall predictions of the network (especially if the input features are not human interpretable), we argue that they may serve as a step towards more transparent and interpretable predictions.

\begin{ack}
We would like to thank Nal Kalchbrenner for general advise and feedback on the paper, Mostafa Dehghani, Klaus Greff, Bernhard Schölkopf, Klaus-Robert Müller, Adam Kosiorek, and Peter Battaglia for helpful discussions, and Rishabh Kabra for advise regarding the DeepMind Multi-Object Datasets.
\end{ack}

{\small
\bibliographystyle{unsrtnat}
\bibliography{references}}

\newpage
\appendix

\begin{center}
\parbox{.8\linewidth}{%
	\centering\Large
	\textbf{Supplementary Material for \\Object-Centric Learning with Slot Attention}
}
\end{center}
In Section~\ref{sec:app-limitations}, we highlight some limitations of our work as well as potential directions for future work. In Section~\ref{app:ablations}, we report results of an ablation study on Slot Attention. In Section~\ref{sec:app_other_rez}, we report further qualitative and quantitative results. In Section~\ref{app:proof}, we give the proof for Proposition 1. In Section~\ref{app:implementation_details}, we report details on our implementation and experimental setting.

\section{Limitations}\label{sec:app-limitations}
\looseness=-1We highlight several limitations of the Slot Attention module that could potentially be addressed in future work:

\textbf{Background treatment} \ The background of a scene receives no special treatment in Slot Attention as all slots use the same representational format. Providing special treatment for backgrounds (e.g., by assigning a separate background slot) is interesting for future work.

\textbf{Translation symmetry} \ The positional encoding used in our experiments is absolute and hence our module is not equivariant to translations. Using a patch-based object extraction process as in~\citep{lin2020space} or an attention mechanism with relative positional encoding~\citep{shaw2018self} are promising extensions.

\textbf{Type of clustering} \ Slot Attention does not know about objects per-se: segmentation is solely driven by the downstream task, i.e., Slot Attention does not distinguish between clustering objects, colors, or simply spatial regions, but relies on the downstream task to drive the specialization to objects.

\textbf{Communication between slots} \ In Slot Attention, slots only communicate via the softmax attention over the input keys, which is normalized across slots. This choice of normalization establishes competition between the slots for explaining parts of the input, which can drive specialization of slots to objects. In some scenarios it can make sense to introduce a more explicit form of communication between slots, e.g. via slot-to-slot message passing in the form of a graph neural network as in~\citep{watters2017visual,kipf2019contrastive} or self-attention~\citep{vaswani2017attention,lee2018set,santoro2018relational,goyal2019recurrent,kosiorek2020conditional}. This can be beneficial for modeling systems of objects that interact dynamically~\citep{watters2017visual,kipf2019contrastive,goyal2019recurrent} or for set generation conditioned on a single vector (as opposed to an image or a set of vectors)~\citep{kosiorek2020conditional}.

\section{Model Ablations}\label{app:ablations}
In this section, we investigate the importance of individual components and modeling choices in the Slot Attention module and compare our default choice to a variety of reasonable alternatives. For simplicity, we report all results on a smaller validation set consisting of $500$ validation images for property prediction (instead of 15K) and on $320$ training images for object discovery. In the unsupervised case, results on the training set and on held-out validation examples are nearly identical.

\begin{figure}[h!]
\centering
    \quad
    \begin{subfigure}[t]{0.47\textwidth}
    \includegraphics[scale=0.225,trim={0 -1.19cm 0 0},clip]{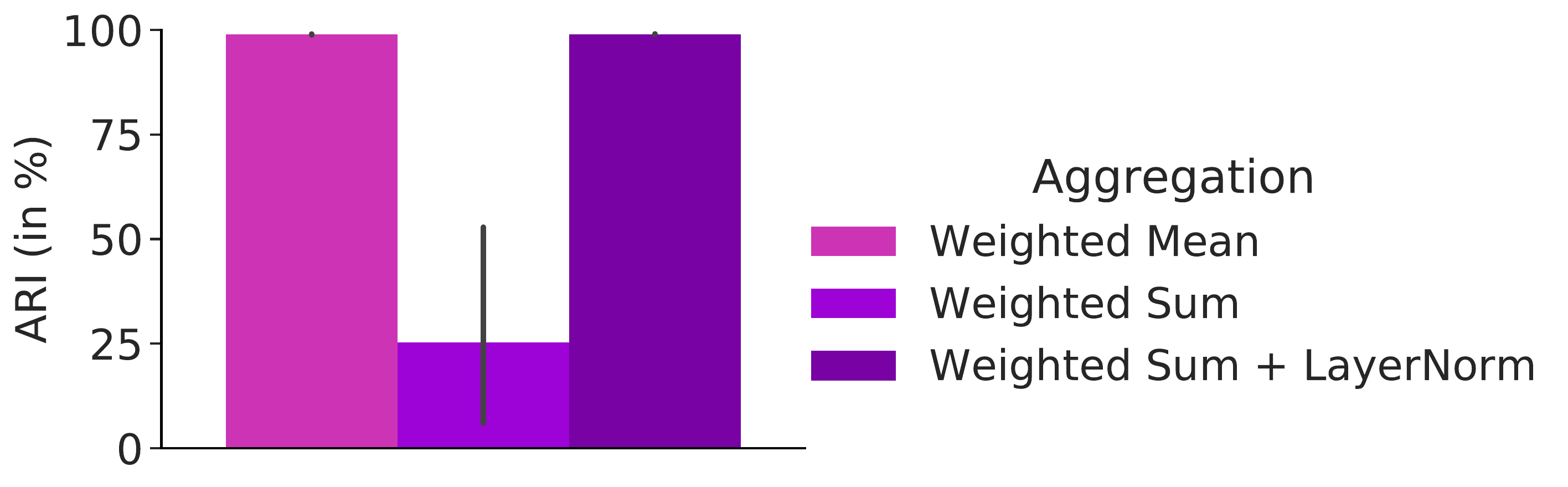}
    \end{subfigure}
    ~
    \begin{subfigure}[t]{0.47\textwidth}
    \includegraphics[scale=0.225,trim={0 0.4cm 0 0},clip]{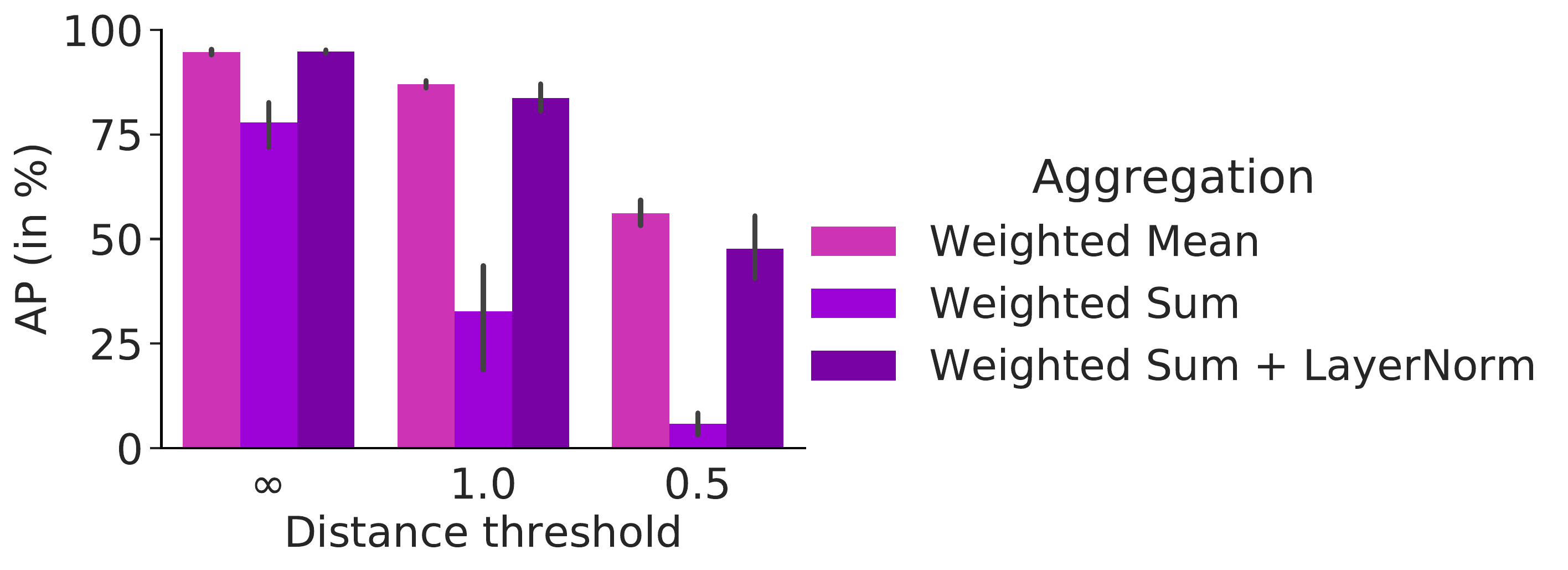}
    \end{subfigure}
    \caption{Aggregation function variants (Line~8) for object discovery on CLEVR6 (left) and property prediction on CLEVR10 (right).}
    \label{fig:pp_input_norm}
\end{figure}

\textbf{Value aggregation} \ In Figure~\ref{fig:pp_input_norm}, we show the effect of taking a weighted sum as opposed to a weighted average in Line~8 of Algorithm~1. The average stabilizes training and yields significantly higher ARI and Average Precision scores (especially at the more strict distance thresholds). We can obtain a similar effect by replacing the weighted mean with a weighted sum followed by layer normalization (LayerNorm)~\citep{ba2016layer}.

\begin{figure}[h!]
\centering
    \quad
    \begin{subfigure}[t]{0.47\textwidth}
    \includegraphics[scale=0.225,trim={0 -1.19cm 0 0},clip]{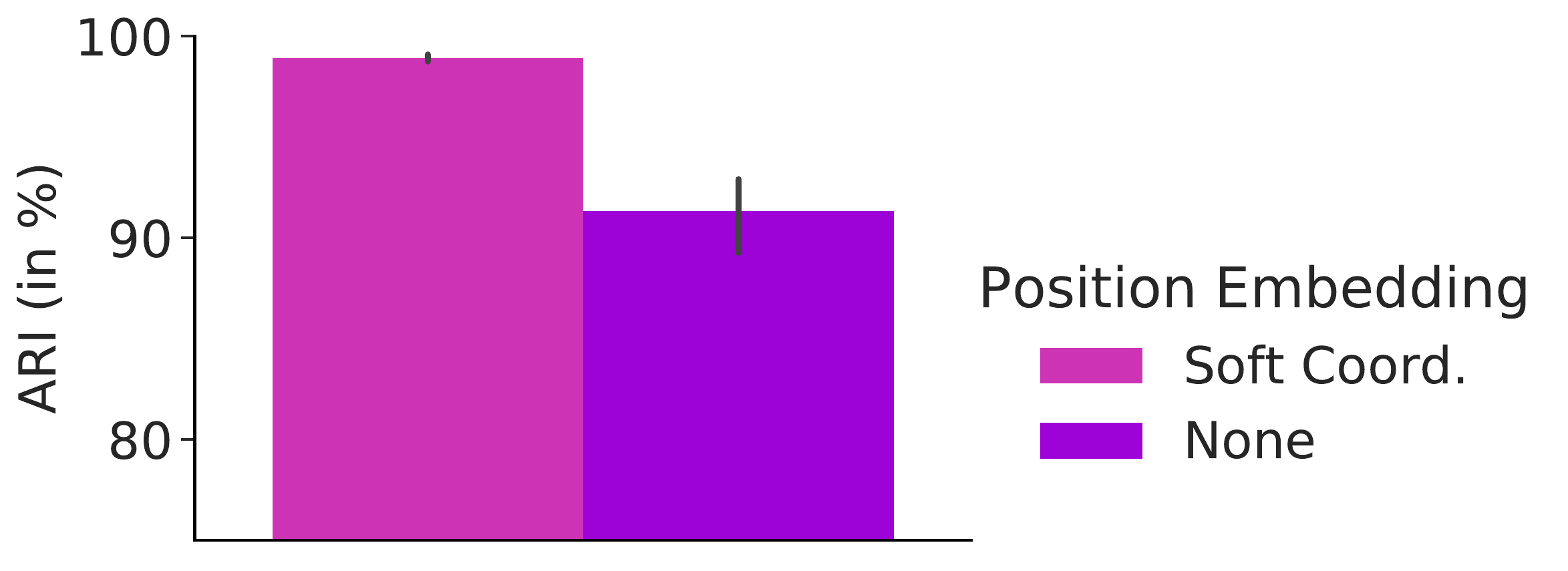}
    \end{subfigure}
    ~
    \begin{subfigure}[t]{0.47\textwidth}
    \includegraphics[scale=0.225,trim={0 0.4cm 0 0},clip]{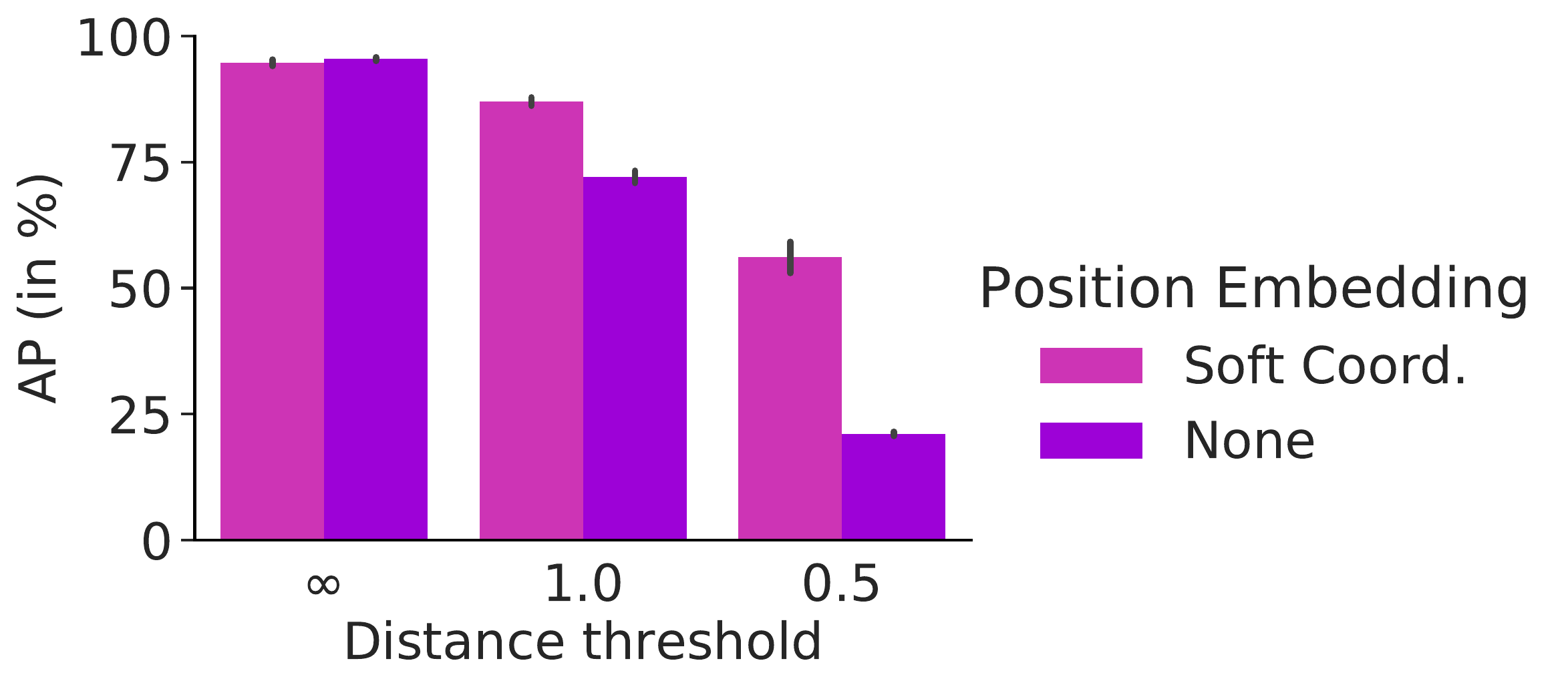}
    \end{subfigure}
    \caption{Ablation on the position embedding for object discovery on CLEVR6 (left) and property prediction on CLEVR10 (right).}
    \label{fig:pp_pos_emb}
\end{figure}

\textbf{Position embedding} \ In Figure~\ref{fig:pp_pos_emb}, we observe that the position embedding is not necessary for predicting categorical object properties. However, the performance in predicting the object position clearly decreases if we remove the position embedding. Similarly, the ARI score in unsupervised object discovery is significantly lower when not adding positional information to the CNN feature maps before the Slot Attention module.

\begin{figure}[h!]
\centering
    \quad
    \begin{subfigure}[t]{0.47\textwidth}
    \includegraphics[scale=0.225,trim={0 -1.19cm 0 0},clip]{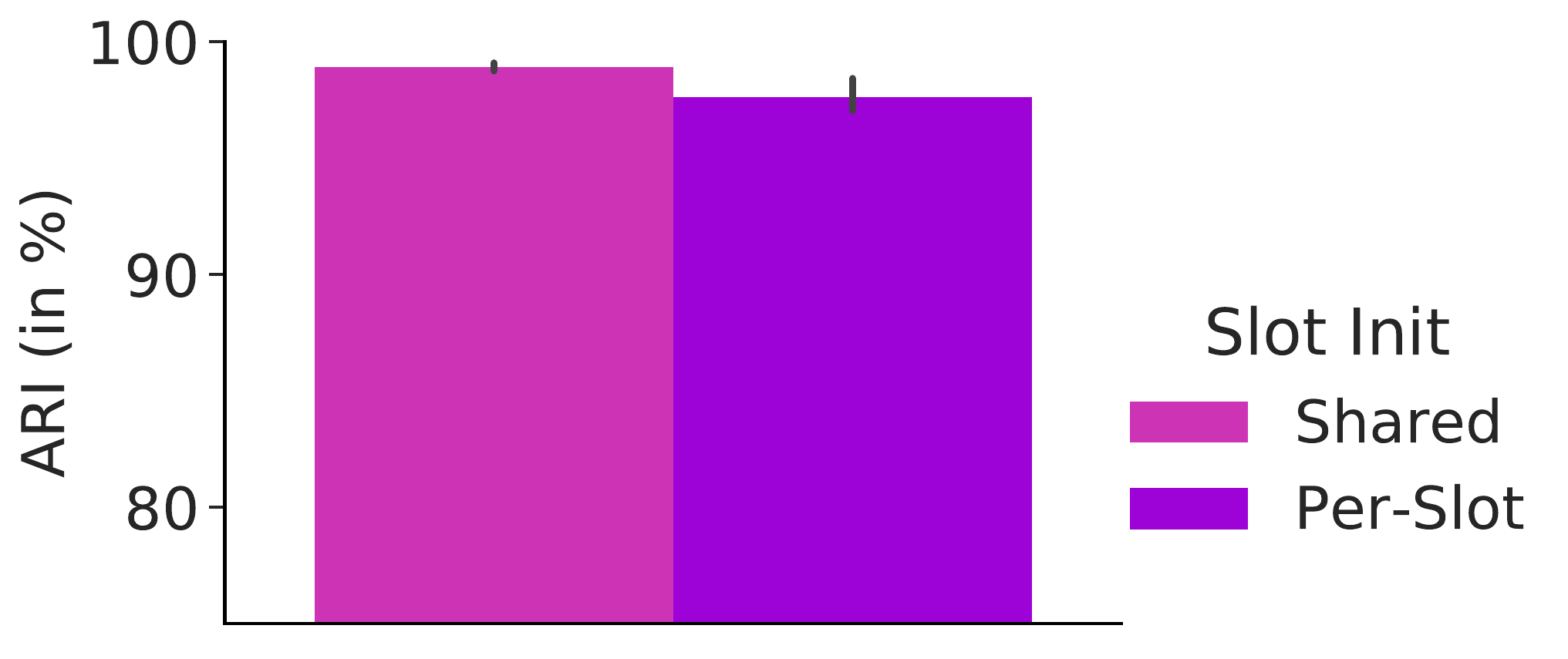}
    \end{subfigure}
    ~
    \begin{subfigure}[t]{0.47\textwidth}
    \includegraphics[scale=0.225,trim={0 0.4cm 0 0},clip]{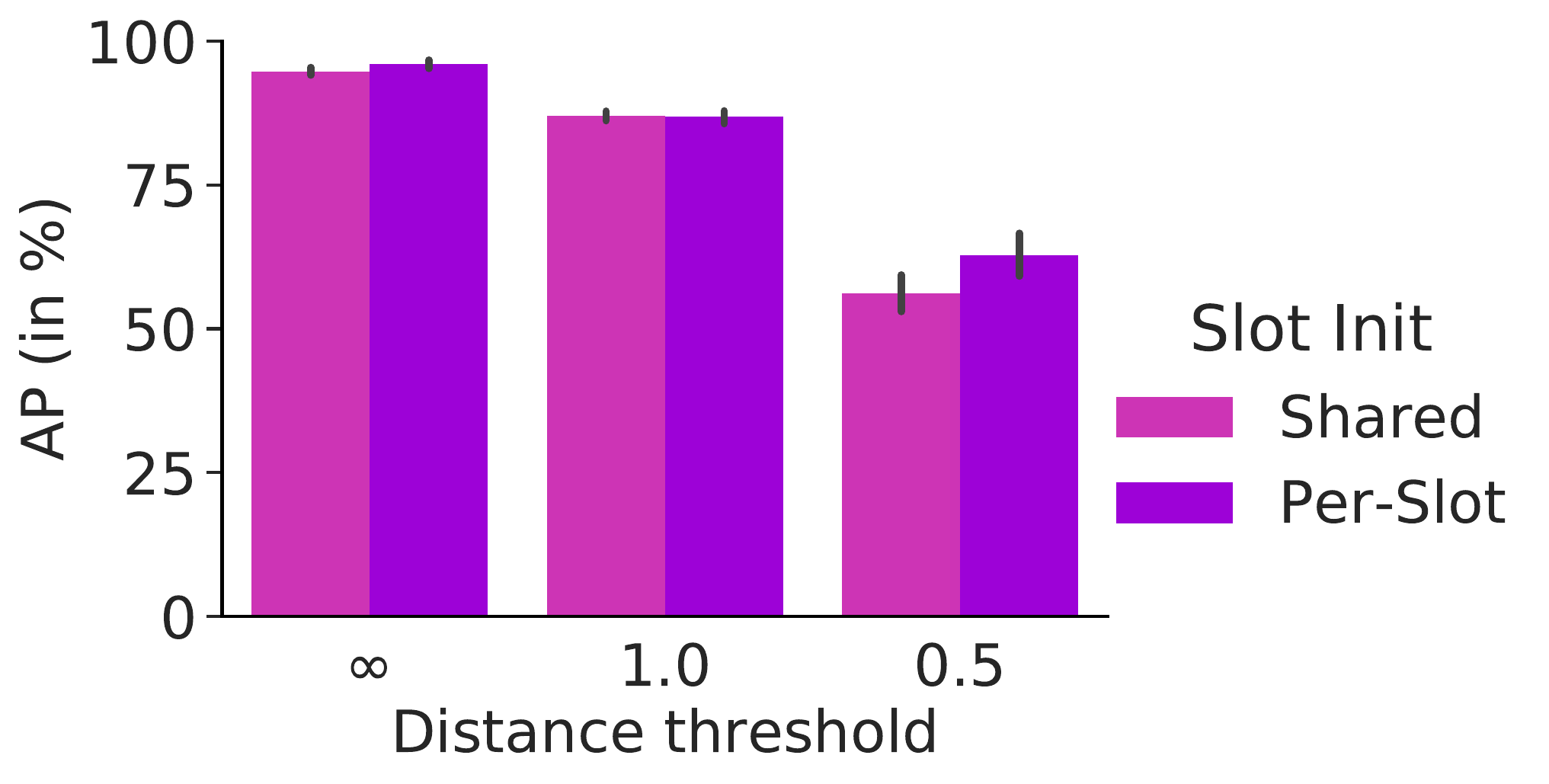}
    \end{subfigure}
    \caption{Slot initialization variants for object discovery on CLEVR6 (left) and property prediction on CLEVR10 (right).}
    \label{fig:pp_learned_init}
\end{figure}

\textbf{Slot initialization} \ In Figure~\ref{fig:pp_learned_init}, we show the effect of learning a separate set of Gaussian mean and variance parameters for the initialization of each slot compared to the default setting of using a shared set of parameters for all slots. We observe that a per-slot parameterization can increase performance slightly for the supervised task, but decreases performance on the unsupervised task, compared to the default shared parameterization. We remark that when learning a separate set of parameters for each slot, adding additional slots at test time is not possible without re-training.

\begin{figure}[h!]
\centering
    \quad
    \begin{subfigure}[t]{0.47\textwidth}
    \includegraphics[scale=0.225,trim={0 -1.19cm 0 0},clip]{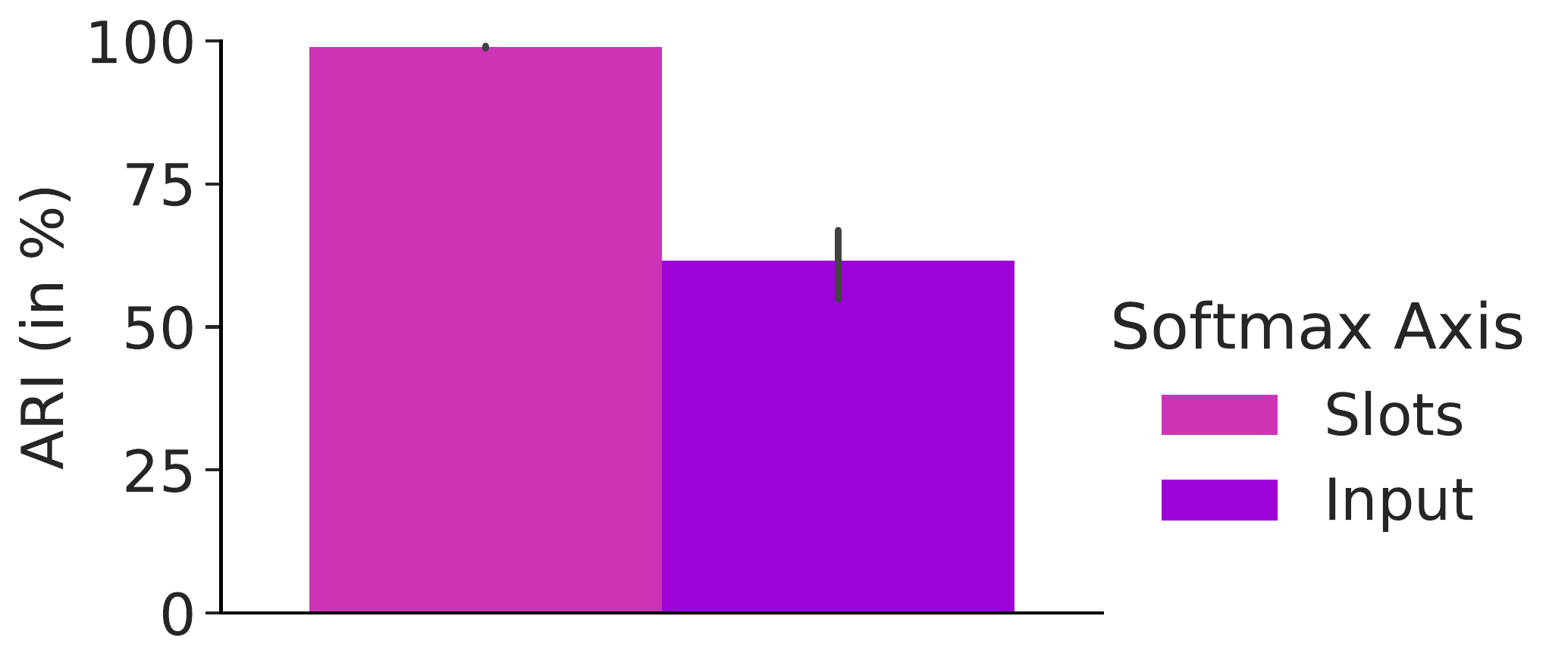}
    \end{subfigure}
    ~
    \begin{subfigure}[t]{0.47\textwidth}
    \includegraphics[scale=0.225,trim={0 0.4cm 0 0},clip]{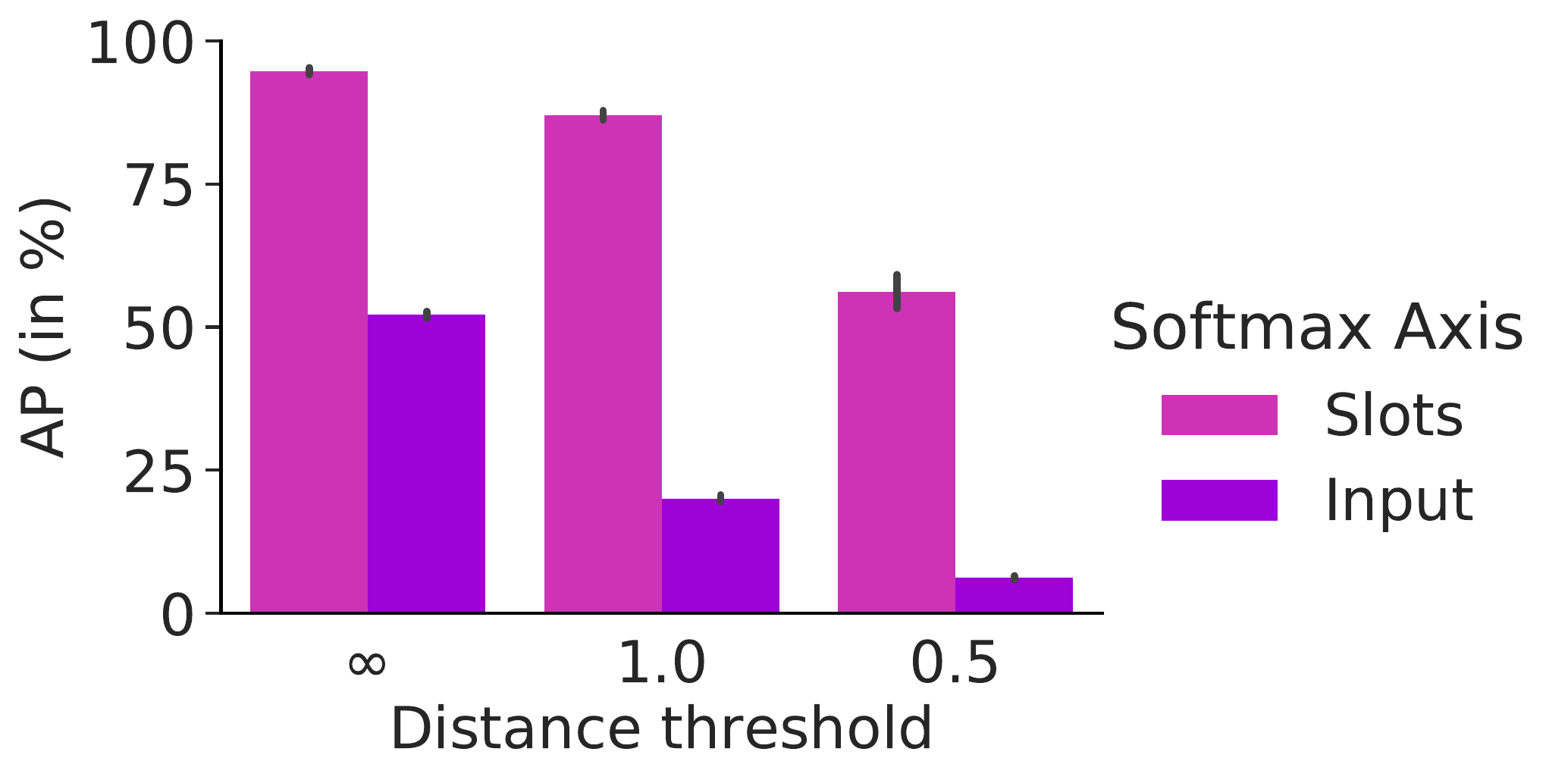}
    \end{subfigure}
    \caption{Choice of softmax axis for object discovery on CLEVR6 (left) and property prediction on CLEVR10 (right).}
    \label{fig:pp_partition_att}
\end{figure}

\textbf{Attention normalization axis} \ In Figure~\ref{fig:pp_partition_att}, we highlight the role of the softmax axis in the attention mechanism, i.e., over which dimension the normalization is performed. Taking the softmax over the slot axis induces competition among the slots for explaining parts of the input. When we take the softmax over the input axis instead\footnote{In this case we aggregate using a weighted sum, as the attention coefficients are normalized over the inputs.} (as done in regular self-attention), the attention coefficients for each slot will be independent of all other slots, and hence slots have no means of exchanging information, which significantly harms performance on both tasks.

\begin{figure}[h!]
\centering
    \quad
    \begin{subfigure}[t]{0.47\textwidth}
    \includegraphics[scale=0.225,trim={0 -1.19cm 0 0},clip]{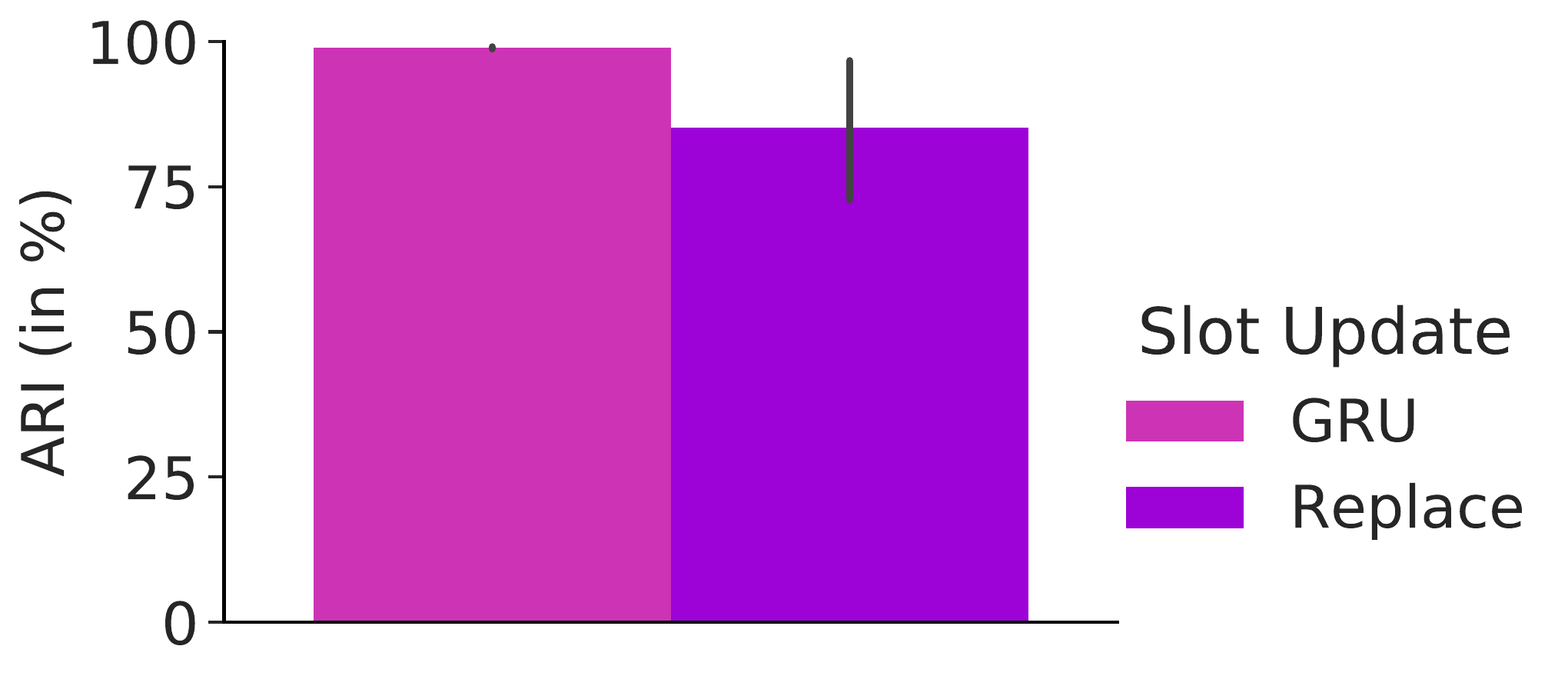}
    \end{subfigure}
    ~
    \begin{subfigure}[t]{0.47\textwidth}
    \includegraphics[scale=0.225,trim={0 0.4cm 0 0},clip]{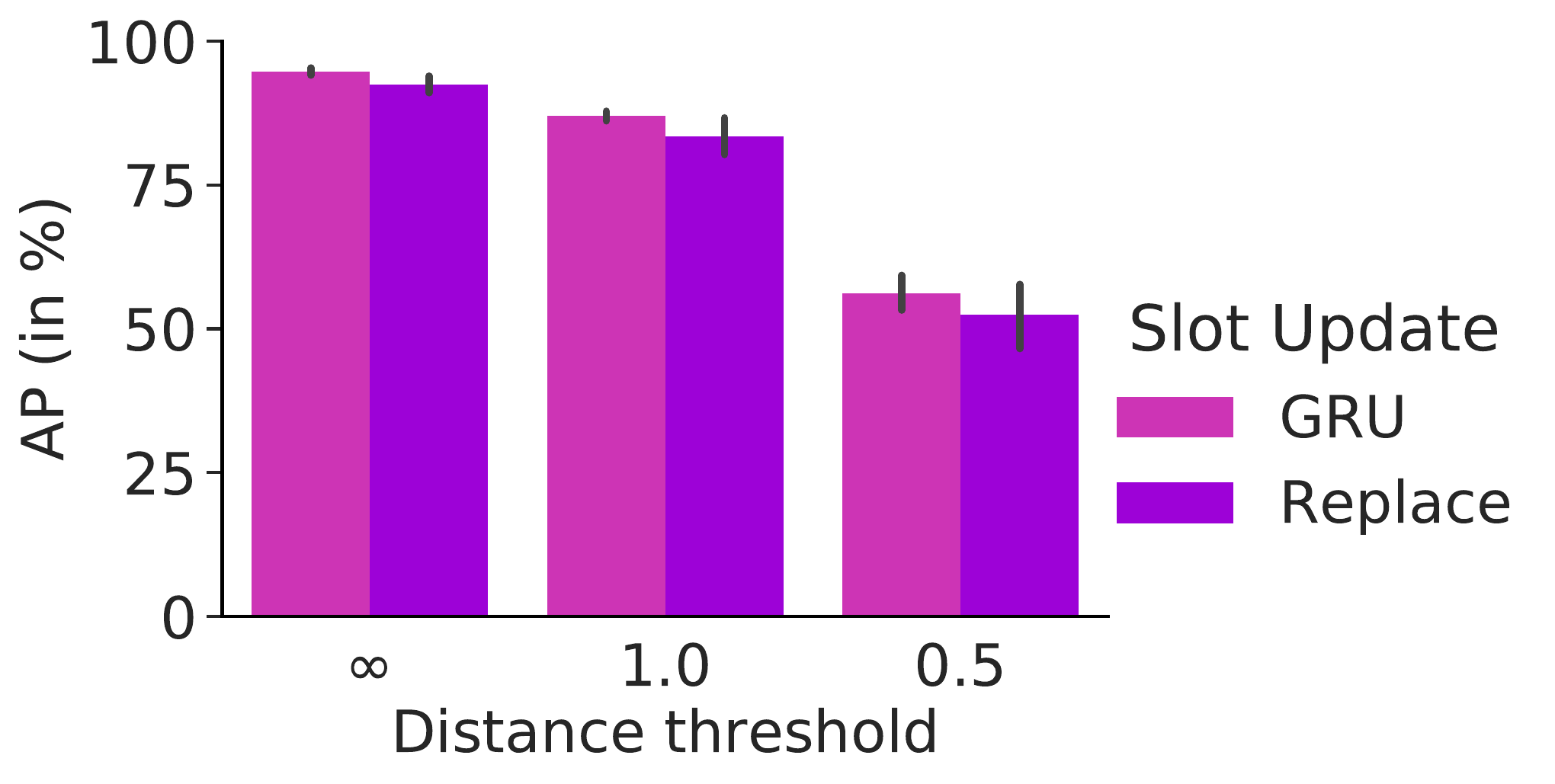}
    \end{subfigure}
    \caption{Slot update function variants for object discovery on CLEVR6 (left) and property prediction on CLEVR10 (right).}
    \label{fig:pp_update_c}
\end{figure}

\textbf{Recurrent update function} \ In Figure~\ref{fig:pp_update_c}, we highlight the role of the GRU in learning the update function for the slots as opposed to simply taking the output of Line~8 as the next value for the slots. We observe that the learned update function yields a noticable improvement.

\begin{figure}[h!]
\centering
    \quad
    \begin{subfigure}[t]{0.47\textwidth}
    \includegraphics[scale=0.225,trim={0 -1.19cm 0 0},clip]{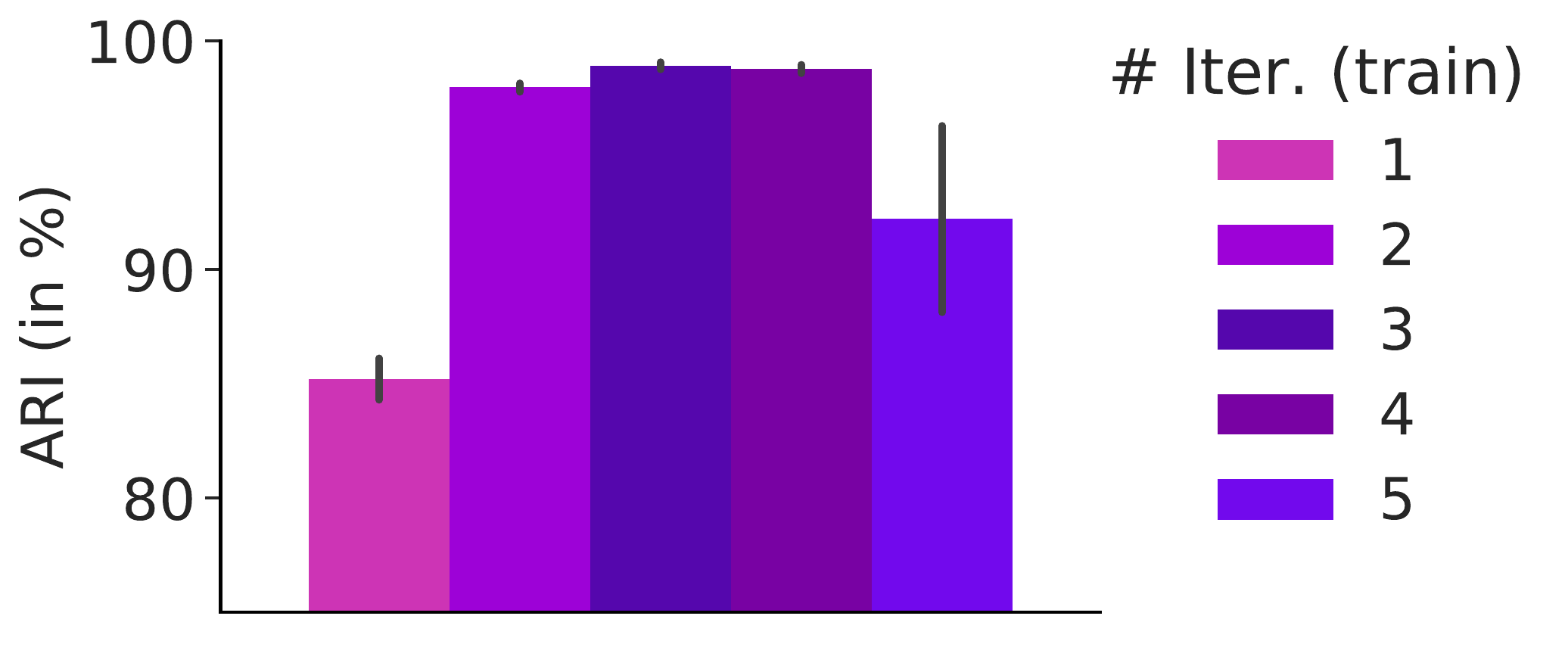}
    \end{subfigure}
    ~
    \begin{subfigure}[t]{0.47\textwidth}
    \includegraphics[scale=0.225,trim={0 0.4cm 0 0},clip]{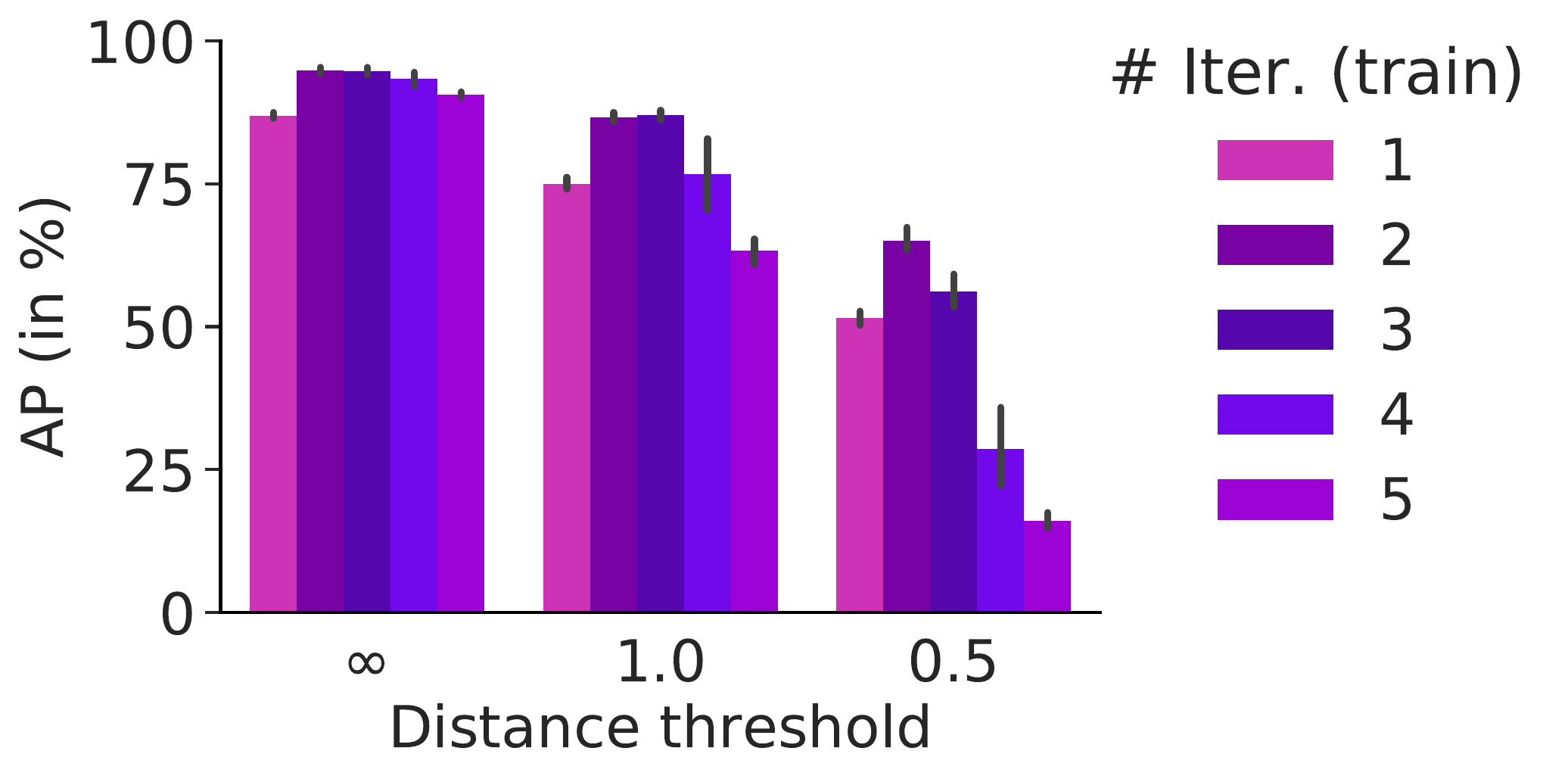}
    \end{subfigure}
    \caption{Number of attention iterations during training for object discovery on CLEVR6 (left) and property prediction on CLEVR10 (right).}
    \label{fig:pp_iters}
\end{figure}

\textbf{Attention iterations} \ In Figure~\ref{fig:pp_iters}, we show the impact of the number of attention iterations while training. We observe a clear benefit in having more than a single attention iteration. Having more than 3 attention iterations significantly slows down training convergence, which results in lower performance when trained for the same number of steps. This can likely be mitigated by instead decoding and applying a loss at every attention iteration as opposed to only the last one. We note that at test time, using more than 3 attention iterations (even when trained with only 3 iterations) generally improves performance.

\begin{figure}[h!]
\centering
    \quad
    \begin{subfigure}[t]{0.47\textwidth}
    \includegraphics[scale=0.225,trim={0 -1.19cm 0 0},clip]{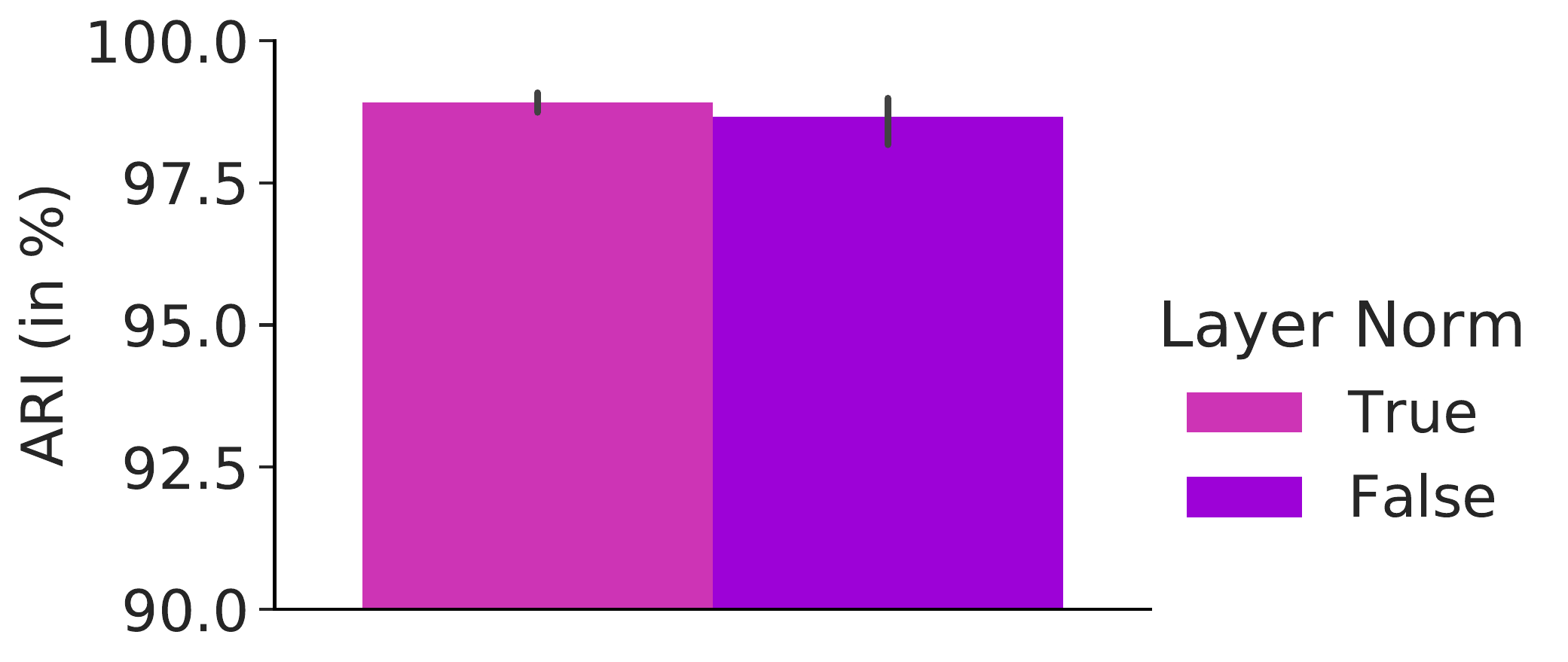}
    \end{subfigure}
    ~
    \begin{subfigure}[t]{0.47\textwidth}
    \includegraphics[scale=0.225,trim={0 0.4cm 0 0},clip]{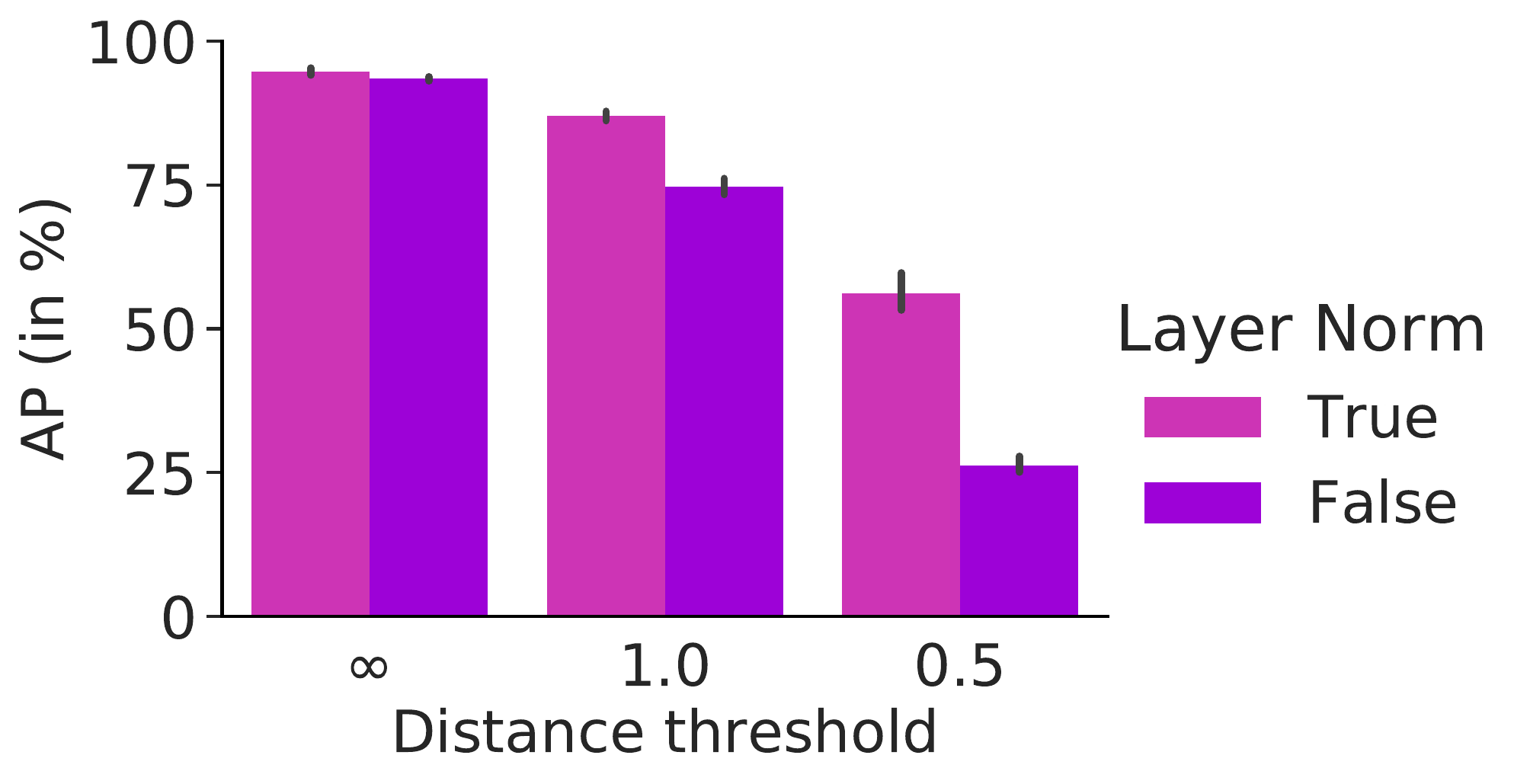}
    \end{subfigure}
    \caption{LayerNorm in the Slot Attention Module for object discovery on CLEVR6 (left) and property prediction on CLEVR10 (right).}
    \label{fig:pp_layernorm}
\end{figure}

\textbf{Layer normalization} \ In Figure~\ref{fig:pp_layernorm}, we show that applying layer normalization (LayerNorm)~\citep{ba2016layer} to the inputs and to the slot representations at each iteration in the Slot Attention module improves predictive performance. For set prediction, it particularly improves its ability to predict position accurately, likely because it leads to faster convergence at training time.

\begin{figure}[h!]
\centering
    \quad
    \begin{subfigure}[t]{0.47\textwidth}
    \includegraphics[scale=0.225,trim={0 -1.19cm 0 0},clip]{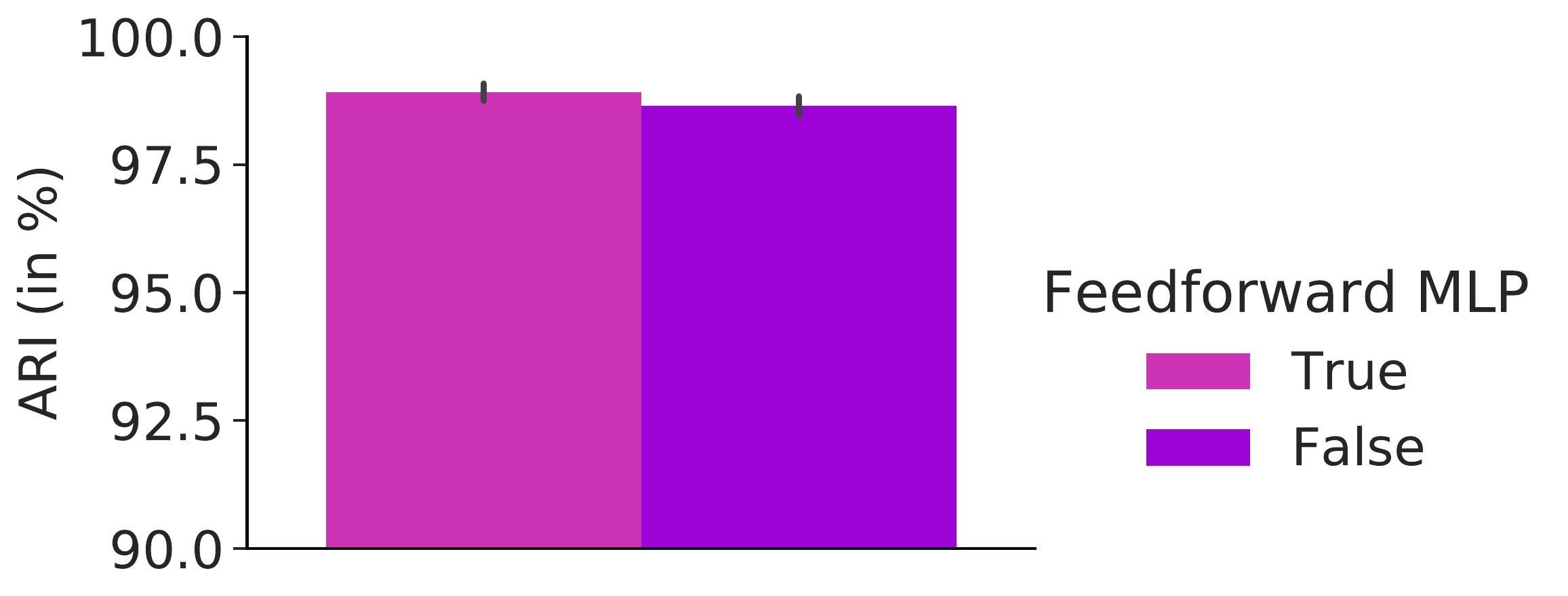}
    \end{subfigure}
    ~
    \begin{subfigure}[t]{0.47\textwidth}
    \includegraphics[scale=0.225,trim={0 0.4cm 0 0},clip]{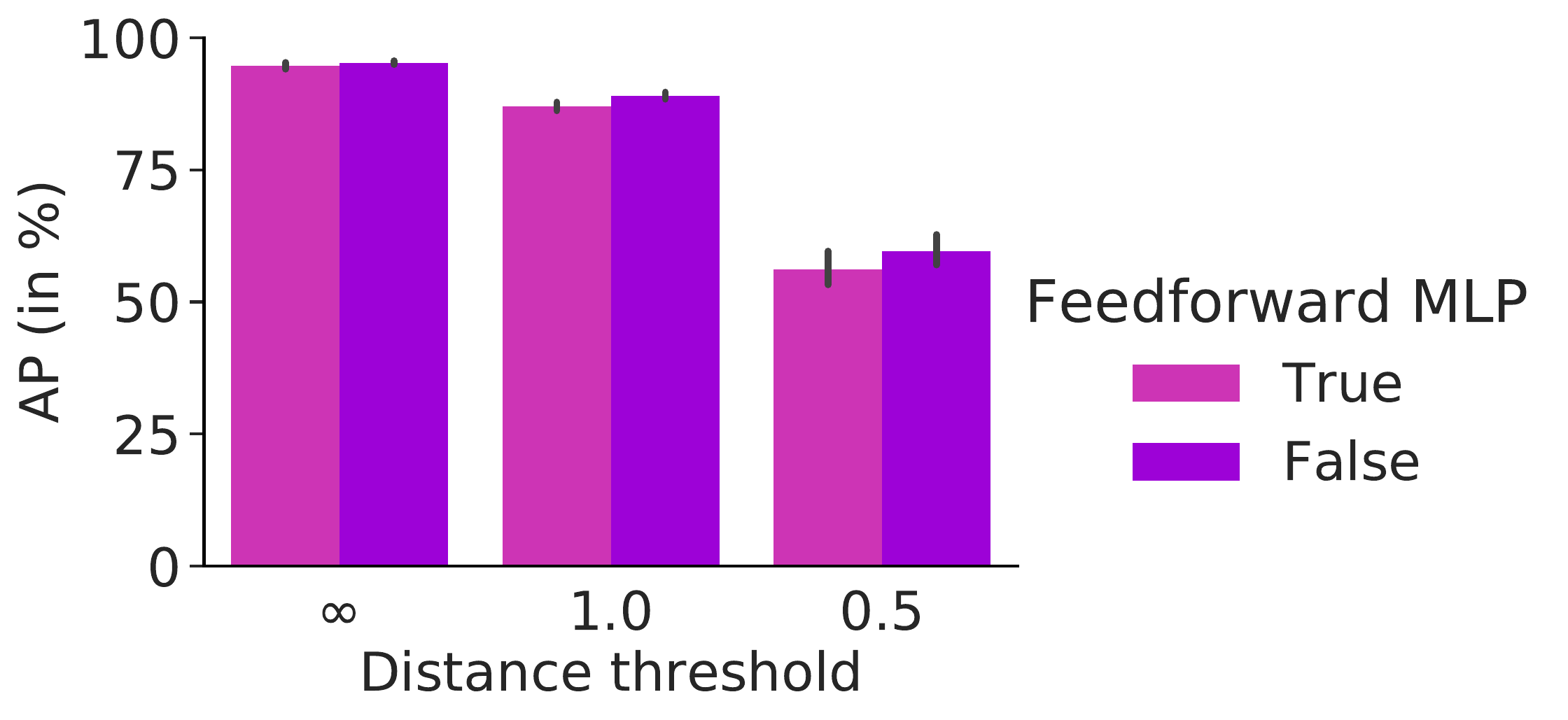}
    \end{subfigure}
    \caption{Optional feedforward MLP for object discovery on CLEVR6 (left) and property prediction on CLEVR10 (right).}
    \label{fig:pp_ff}
\end{figure}

\textbf{Feedforward network} \ In Figure~\ref{fig:pp_ff}, we show that the residual MLP after the GRU is optional and may slow down convergence in property prediction, but may slightly improve performance on object discovery.

\begin{figure}[h!]
\centering
    \quad
    \begin{subfigure}[t]{0.47\textwidth}
    \includegraphics[scale=0.225,trim={0 -1.19cm 0 0},clip]{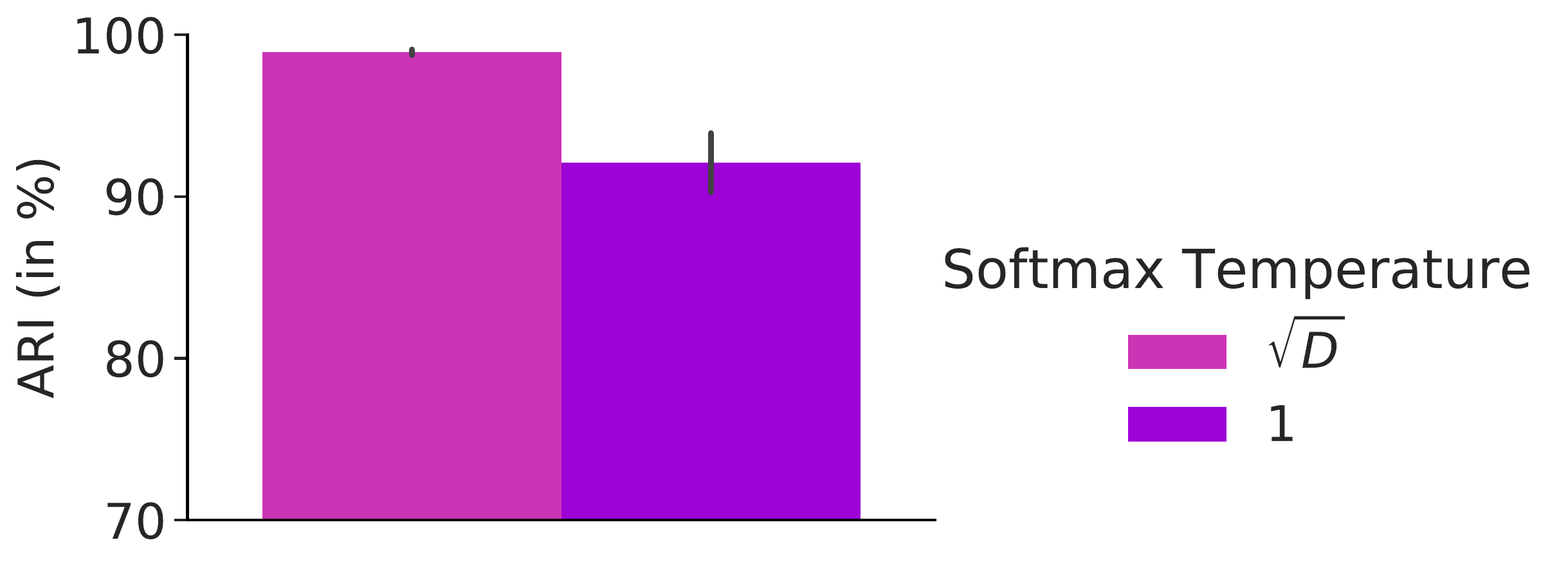}
    \end{subfigure}
    ~
    \begin{subfigure}[t]{0.47\textwidth}
    \includegraphics[scale=0.225,trim={0 0.4cm 0 0},clip]{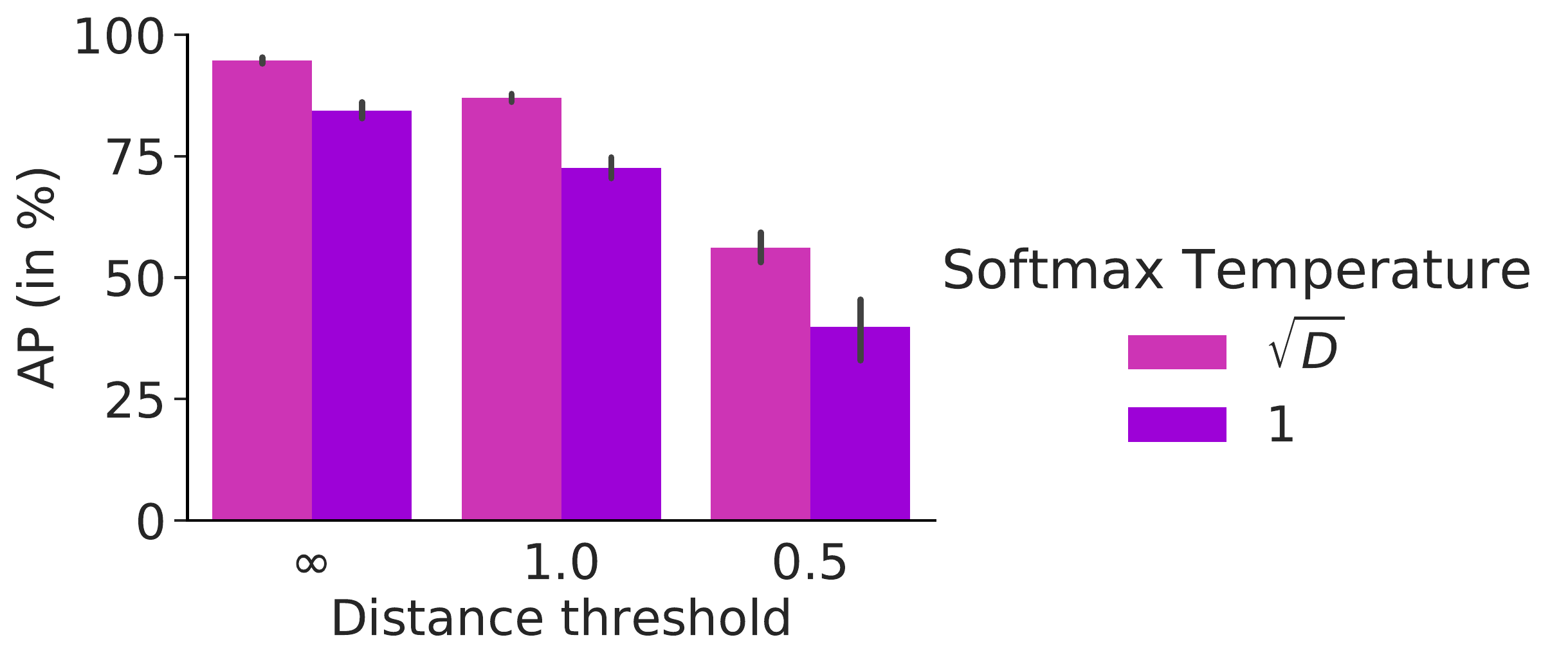}
    \end{subfigure}
    \caption{Softmax temperature in the Slot Attention Module for object discovery on CLEVR6 (left) and property prediction on CLEVR10 (right).}
    \label{fig:pp_D}
\end{figure}

\textbf{Softmax temperature} \ In Figure~\ref{fig:pp_D}, we show the effect of the softmax temperature. The scaling of  $\sqrt{D}$ clearly improves the performance on both tasks.

\begin{figure}[h!]
\centering
    \quad
    \begin{subfigure}[t]{0.47\textwidth}
    \includegraphics[scale=0.225,trim={0 -1.19cm 0 0},clip]{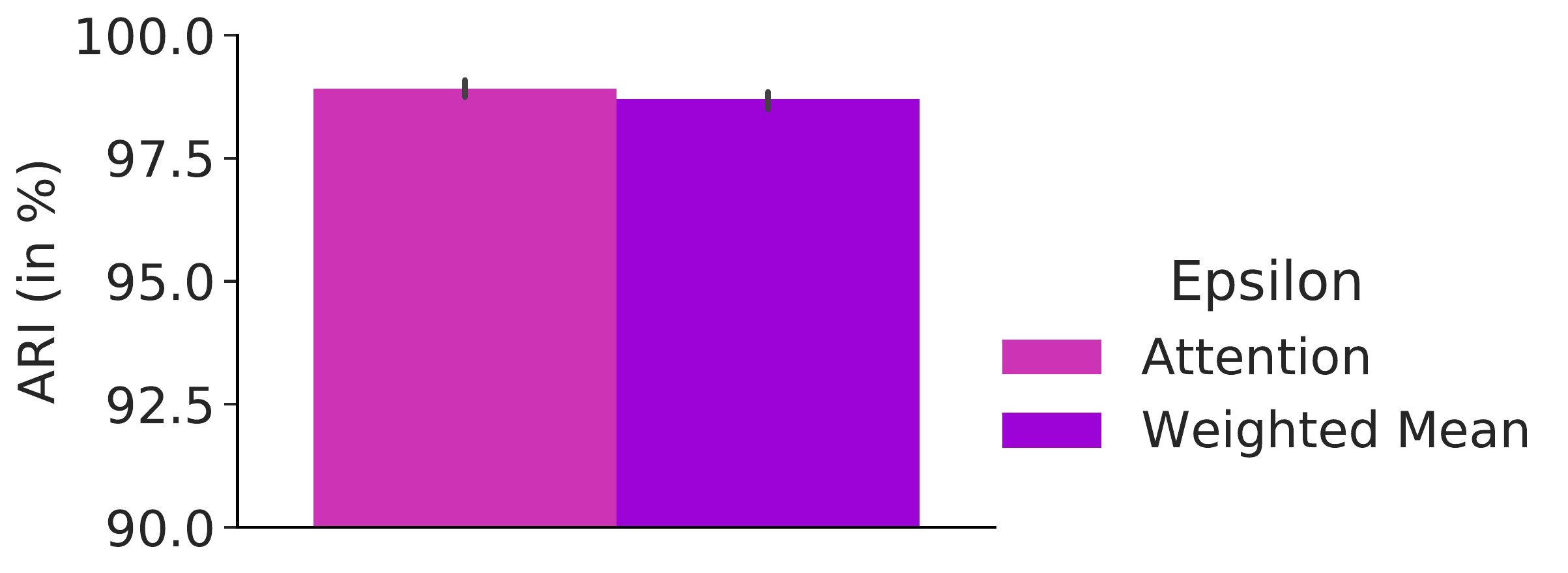}
    \end{subfigure}
    ~
    \begin{subfigure}[t]{0.47\textwidth}
    \includegraphics[scale=0.225,trim={0 0.4cm 0 0},clip]{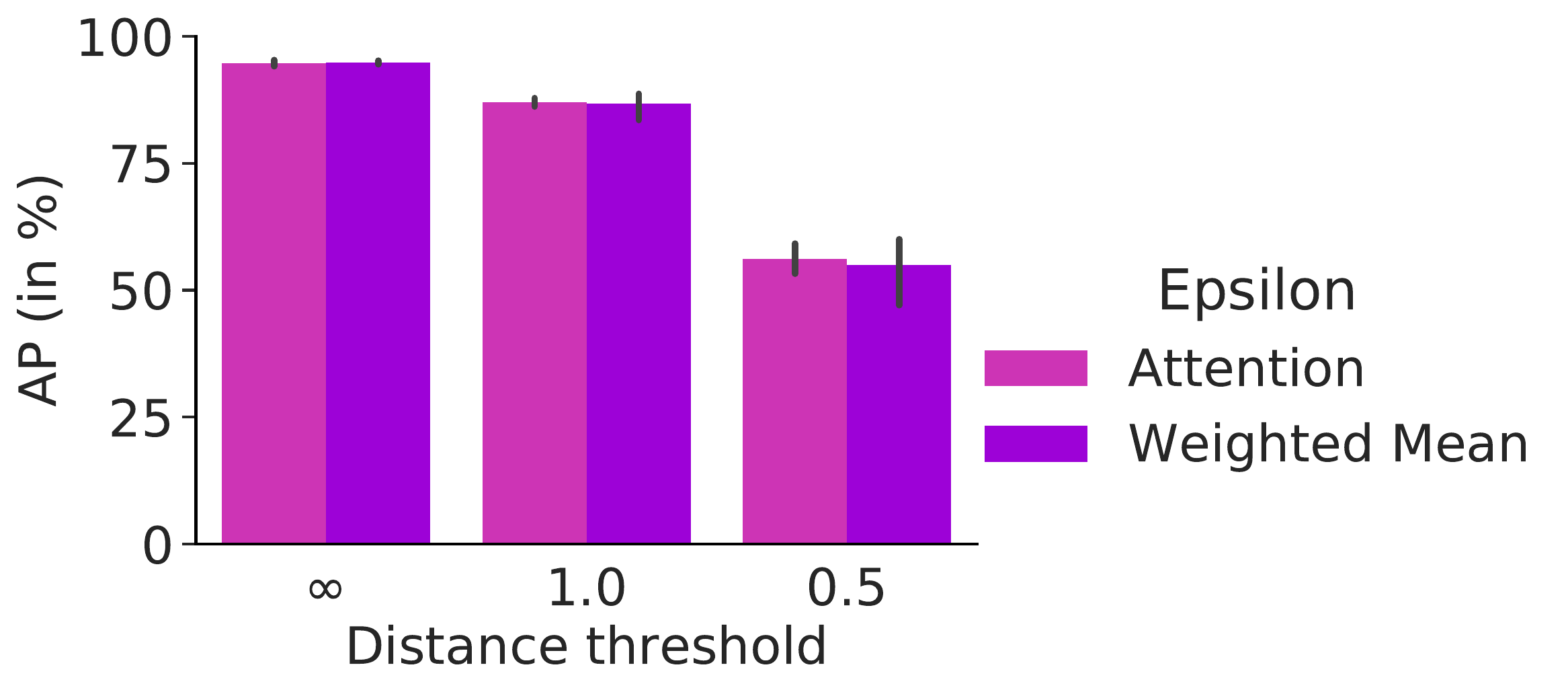}
    \end{subfigure}
    \caption{Offset in the attention maps or the denominator of the weighted mean for object discovery on CLEVR6 (left) and property prediction on CLEVR10 (right). }
    \label{fig:pp_epsilon}
\end{figure}

\textbf{Offset for numerical stability} \ In Figure~\ref{fig:pp_epsilon}, we show that adding a small offset to the attention maps (for numerical stability) as in Algorithm~1 compared to the alternative of adding an offset to the denominator in the weighted mean does not significantly change the result in either task.

\begin{figure}[h!]
\centering
    \quad
    \begin{subfigure}[t]{0.47\textwidth}
    \includegraphics[scale=0.225,trim={0 -1.19cm 0 0},clip]{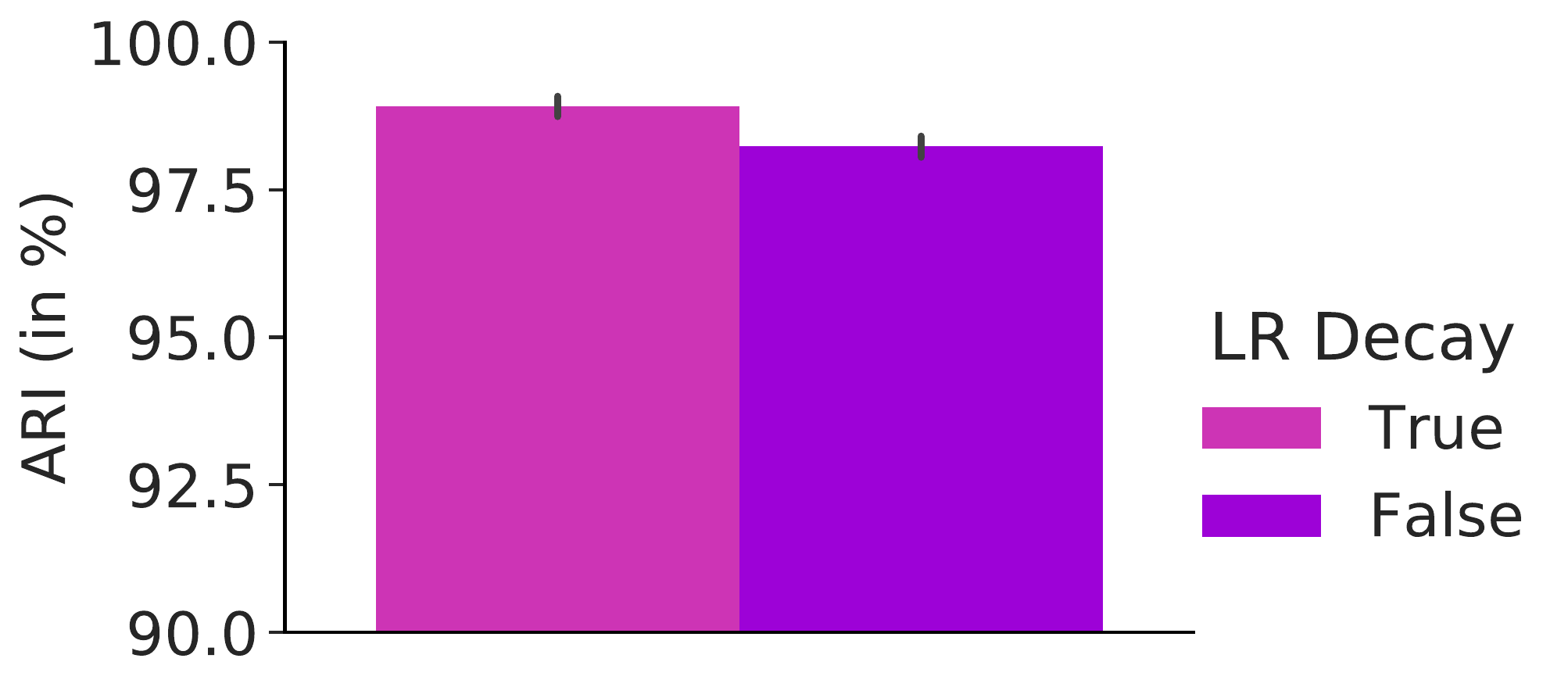}
    \end{subfigure}
    ~
    \begin{subfigure}[t]{0.47\textwidth}
    \includegraphics[scale=0.225,trim={0 0.4cm 0 0},clip]{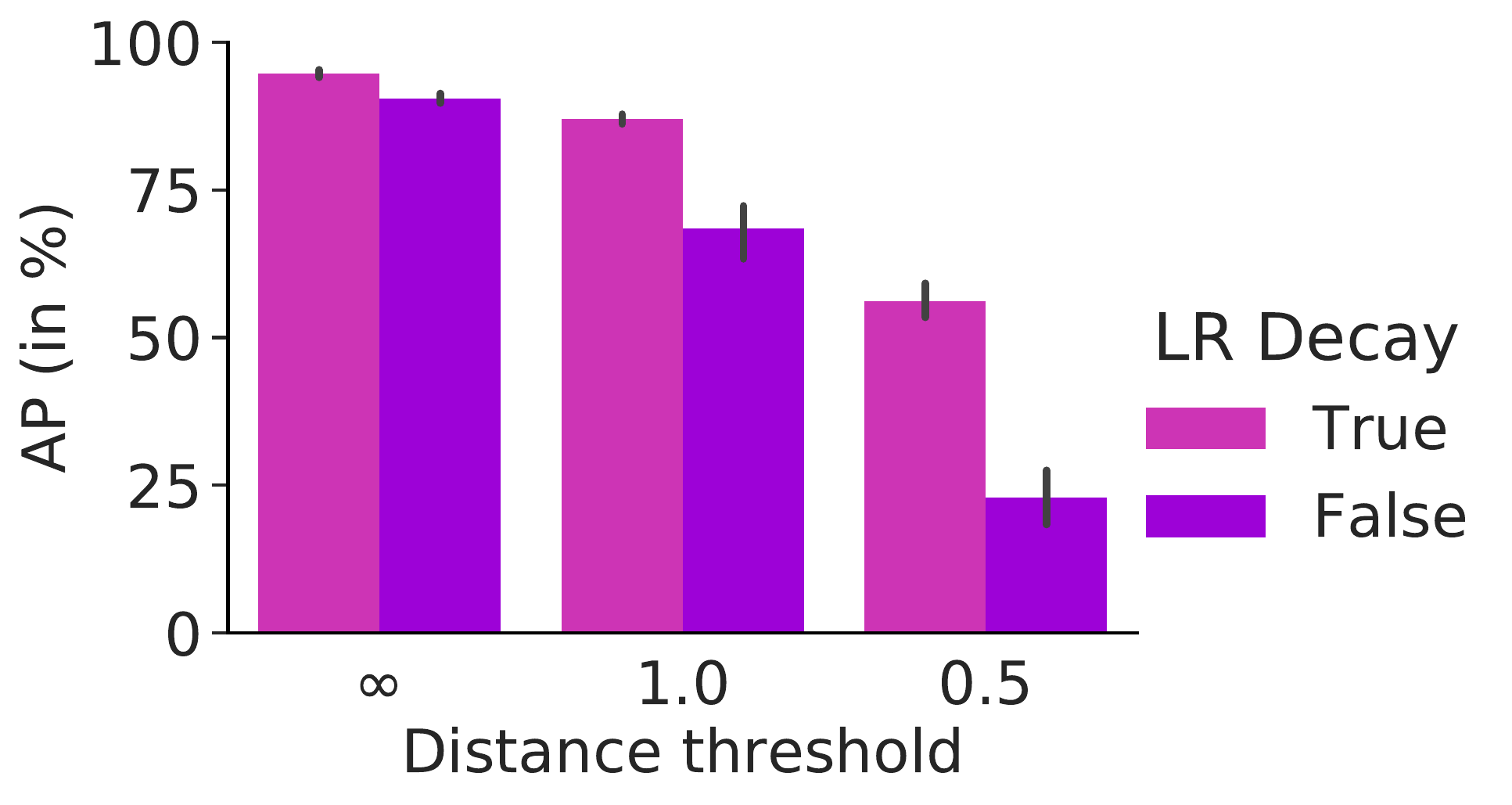}
    \end{subfigure}
    \caption{Learning rate decay for object discovery on CLEVR6 (left) and property prediction on CLEVR10 (right).}
    \label{fig:pp_lr_decay}
\end{figure}

\begin{figure}[h!]
\centering
    \quad
    \begin{subfigure}[t]{0.47\textwidth}
    \includegraphics[scale=0.225,trim={0 -1.19cm 0 0},clip]{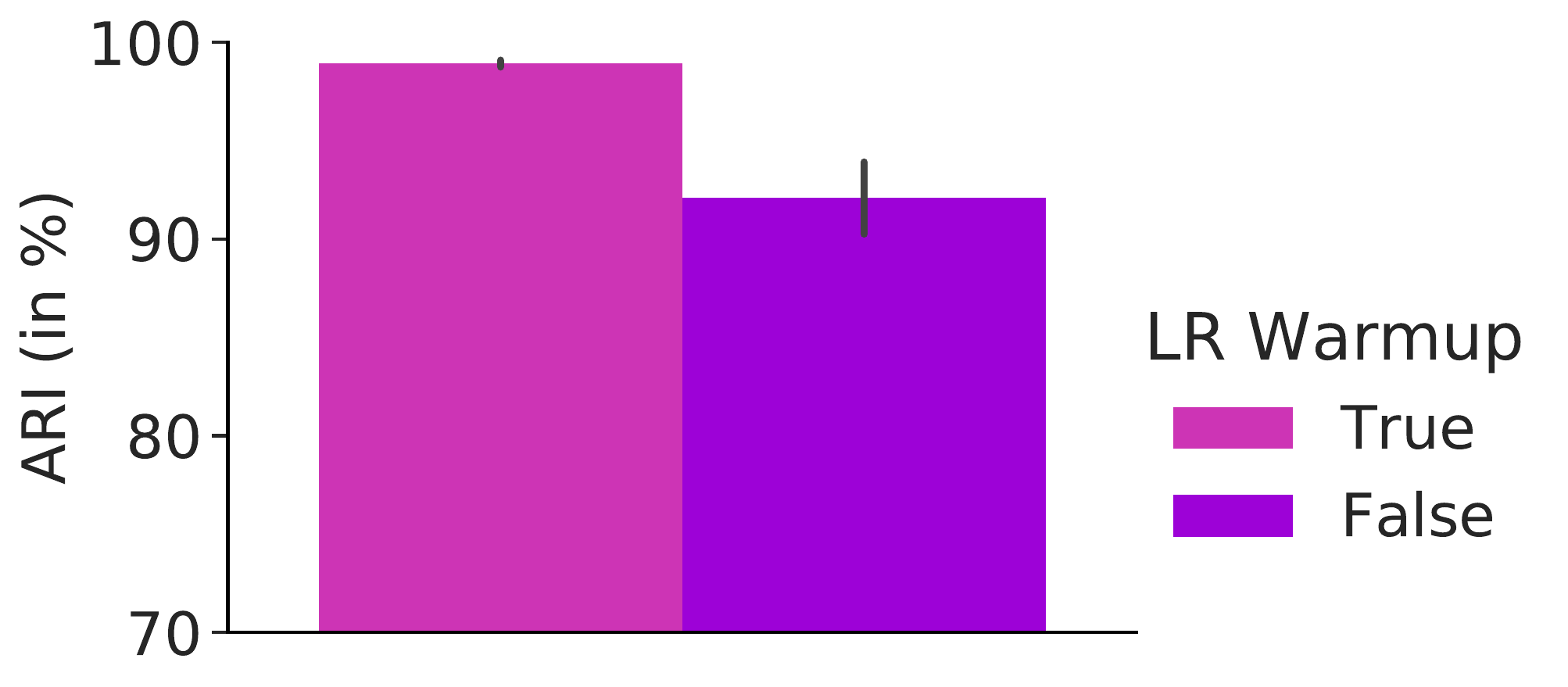}
    \end{subfigure}
    ~
    \begin{subfigure}[t]{0.47\textwidth}
    \includegraphics[scale=0.225,trim={0 0.4cm 0 0},clip]{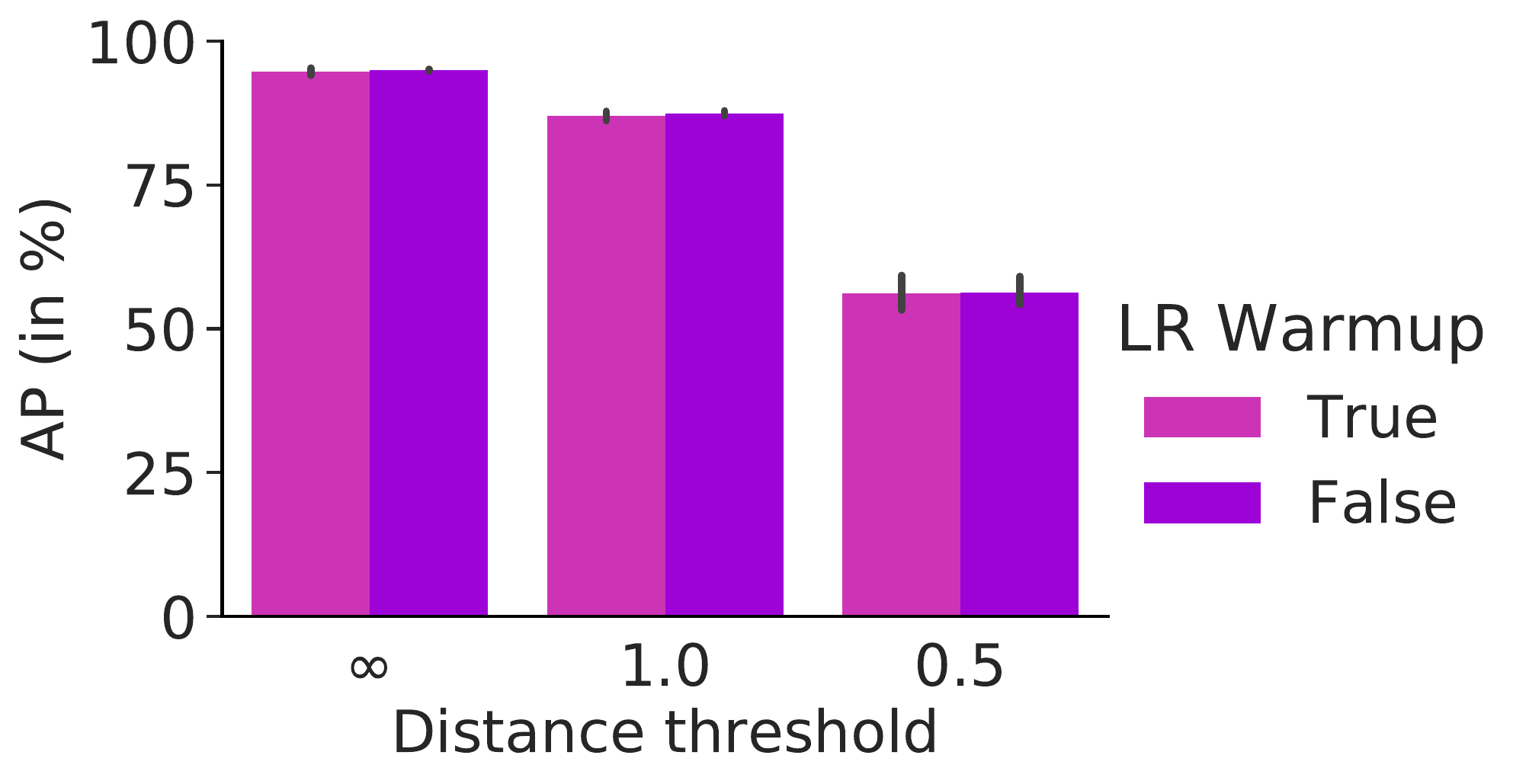}
    \end{subfigure}
    \caption{Learning rate warmup for object discovery on CLEVR6 (left) and property prediction on CLEVR10 (right).}
    \label{fig:pp_lr_warmup}
\end{figure}

\textbf{Learning rate schedules} \ In Figure~\ref{fig:pp_lr_decay} and~\ref{fig:pp_lr_warmup}, we show the effect of our decay and warmup schedules. While we observe a clear benefit from the decay schedule, the warmup seem to be mostly useful in the object discovery setting, where it helps avoid failure cases of getting stuck in a suboptimal solution (e.g., clustering the image into stripes as opposed to objects).

\begin{figure}[h!]
\centering
    \quad
    \begin{subfigure}[t]{0.47\textwidth}
    \includegraphics[scale=0.225,trim={0 -1.19cm 0 0},clip]{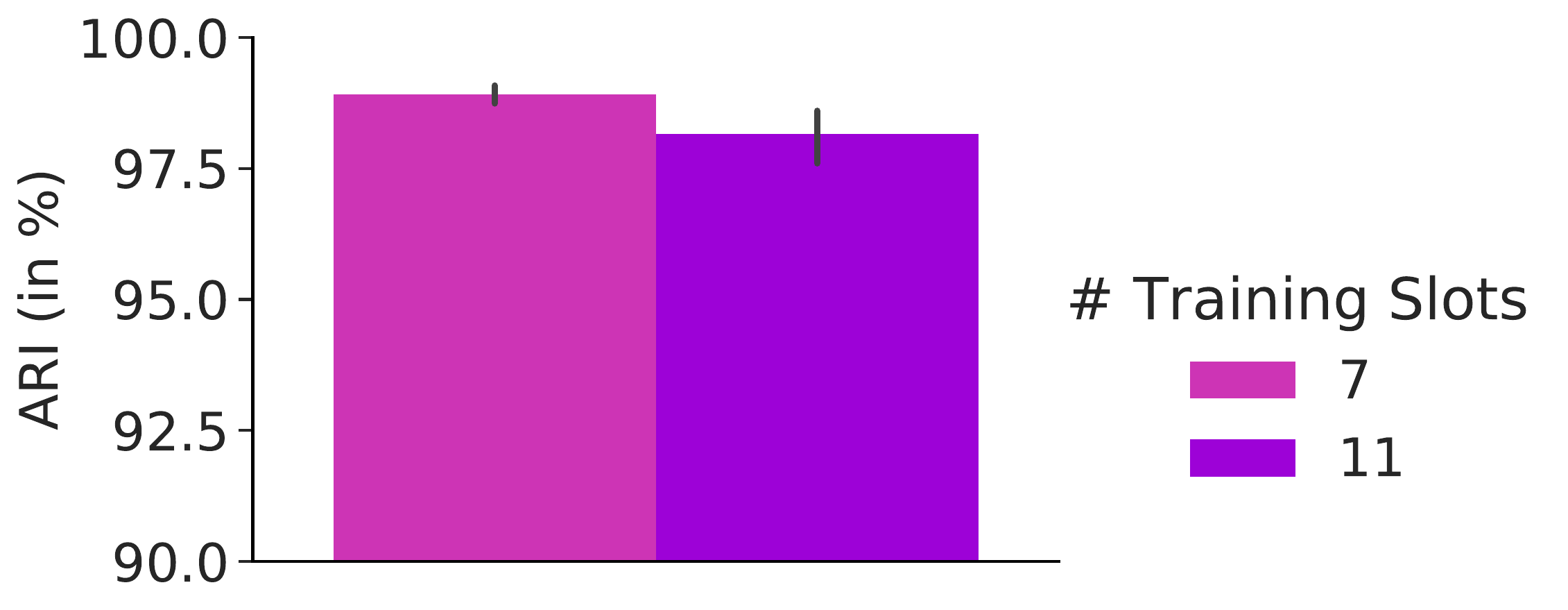}
    \end{subfigure}
    ~
    \begin{subfigure}[t]{0.47\textwidth}
    \includegraphics[scale=0.225,trim={0 0.4cm 0 0},clip]{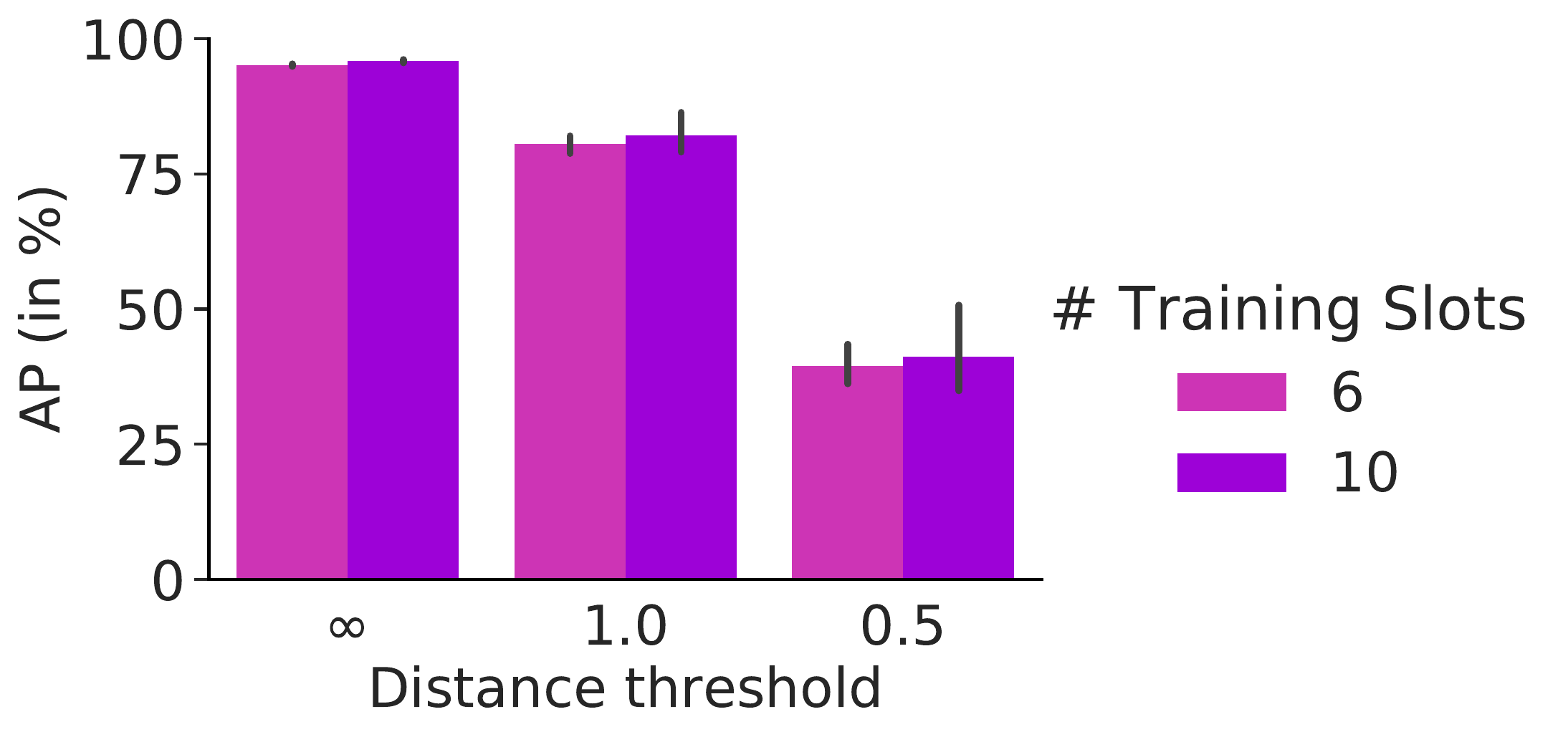}
    \end{subfigure}
    \caption{Number of training slots on CLEVR6 for object discovery (left) and property prediction (right).}
    \label{fig:pp_more_slots}
\end{figure}

\textbf{Number of training slots} \ In Figure~\ref{fig:pp_more_slots}, we show the effect of training with a larger number of slots than necessary. We train both the object discovery and property prediction methods on CLEVR6 with the number of slots we used in our CLEVR10 experiments (note that for object discovery we use an additional slot to be consistent with the baselines). We observe that knowing the precise number of objects in the dataset is generally not required. Training with more slots may even help in the property prediction experiments and is slightly harmful in the object discovery. Overall, this indicates that the model is rather robust to the number of slots (given enough slots to model each object independently). Using a (rough) upper bound to the number of objects in the dataset seem to be a reasonable selection strategy for the number of slots.

\textbf{Soft k-means} \ Slot Attention can be seen as a generalized version of the soft k-means algorithm~\citep{bauckhage2015lecture}. We can reduce Slot Attention to a version of soft k-means with a dot-product scoring function (as opposed to the negative Euclidean distance) by simultaneously replacing the GRU update, all LayerNorm functions and the key/query/value projections with the identity function. Specifically, instead of the GRU update, we simply take the output of Line 8 in Algorithm 1 as the next value for the slots. With these ablations, the model achieves $75.5\pm3.8\%$ ARI on CLEVR6, compared to $98.8\pm0.3\%$ for the full version of Slot Attention.

\section{Further Experimental Results}\label{sec:app_other_rez}
\subsection{Object Discovery}

\paragraph{Runtime} Experiments on a single V100 GPU with 16GB of RAM with 500k steps and a batch size of 64 ran for approximately 7.5hrs for Tetrominoes, 24hrs for multi-dSprites, and 5 days, 13hrs for CLEVR6 (wall-clock time).

\begin{figure}[htp!]
    \centering
    \includegraphics[width=\textwidth,trim={0 0.3cm 0 0.25cm},clip]{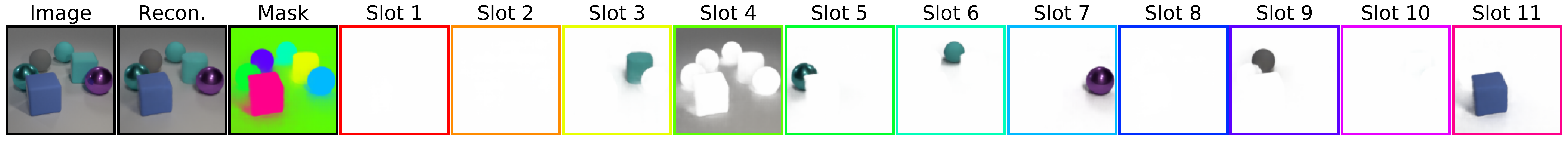}
    \includegraphics[width=\textwidth,trim={0 0.3cm 0 0.3cm},clip]{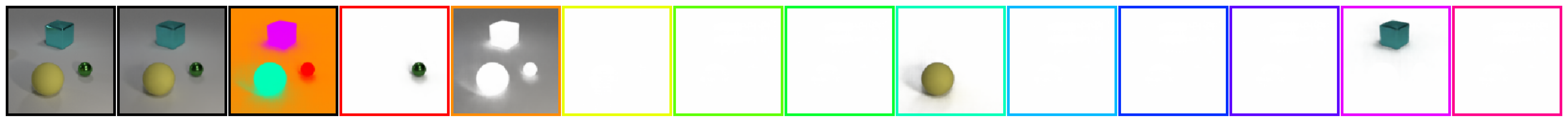}
    \includegraphics[width=\textwidth,trim={0 0.3cm 0 0.3cm},clip]{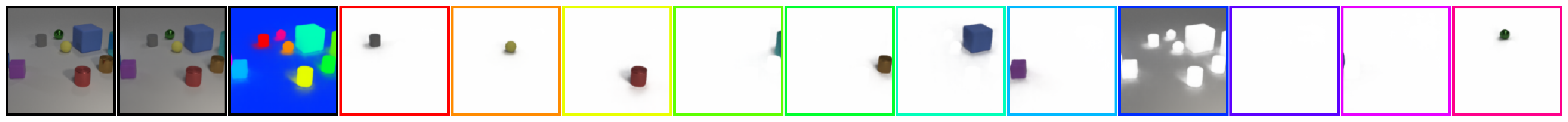}
    \includegraphics[width=\textwidth,trim={0 0.3cm 0 0.3cm},clip]{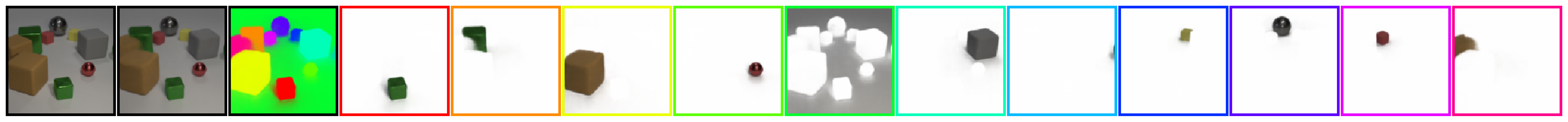}
    \caption{Visualization of overall reconstructions, alpha masks, and per-slot reconstructions for a Slot Attention model trained on CLEVR6 (i.e., on scenes with a maximum number of 6 objects), but tested on scenes with up to 10 objects, using $K=11$ slots both at training and at test time, and $T=3$ iterations at training time and $T=5$ iterations at test time. We only visualize examples where the objects were successfully clustered after $T=5$ iterations. For some random slot initializations, clustering results still improve when run for more iterations. We note that this particular model learned to separate the background out into a separate (but random) slot instead of spreading it out evenly over all slots.}
    \label{fig:app_clevr_11_slots}
\end{figure}

\paragraph{Qualitative results} In Figure~\ref{fig:app_clevr_11_slots}, we show qualitative segmentation results for a Slot Attention model trained on the object discovery task. This model is trained on CLEVR6 but uses $K=11$ instead of the default setting of $K=7$ slots during both training and testing, while all other settings remain unchanged. In this particular experiment, we trained 5 models using this setting with 5 different random seeds for model parameter initialization. Out of these 5 models, we found that a single model learned the solution of placing the background into a separate slot (which is the one we visualize). The typical solution that a Slot Attention-based model finds (for most random seeds) is to distribute the background equally over all slots, which is the solution we highlight in the main paper. In Figure~\ref{fig:app_clevr_11_slots} (bottom two rows), we further show how the model generalizes to scenes with more objects (up to 10) despite being trained on CLEVR6, i.e., on scenes containing a maximum of 6 objects.

\subsection{Set Prediction}
\paragraph{Runtime} Experiments on a single V100 GPU with 16GB of RAM with 150k steps and a batch size of 512 ran for approximately 2 days and 3hrs for CLEVR (wall-clock time).

\paragraph{Qualitative results} In Table~\ref{table:visual_pred}, we show the predictions and attention coefficients of a Slot Attention model on several challenging test examples for the supervised property prediction task. The model was trained with default settings ($T=3$ attention iterations) and the images are selected by hand to highlight especially difficult cases (e.g., multiple identical objects or many partially overlapping objects in one scene). Overall, we can see that the property prediction typically becomes more accurate with more iterations, although the accuracy of the position prediction may decrease. This is not surprising as we only apply the loss at $t=3$, and generalization to more time steps at test time is not guaranteed. We note that one could alternatively apply the loss at every iteration during training, which has the potential to improve accuracy, but would increase computational cost. We observe that the model appears to handle multiple copies of the same object well (top). On very crowded scenes (middle and bottom), we note that the slots have a harder time segmenting the scene, which can lead to errors in the prediction. However, more iterations seem to sharpen the segmentation which in turns improves predictions.

\begin{table}[ht!]
    \centering
    \vspace{-1em}
    \caption{Example predictions of a Slot Attention model trained with $T=3$ on a challenging example with $4$ objects (two of which are identical and partially overlapping) and crowded scenes with $10$ objects. We highlight wrong prediction of attributes and distances greater than $0.5$.}
    \vspace{1em}
    \label{table:visual_pred}
    \includegraphics[width=0.5\textwidth,trim={0 0.4cm 0 0},clip]{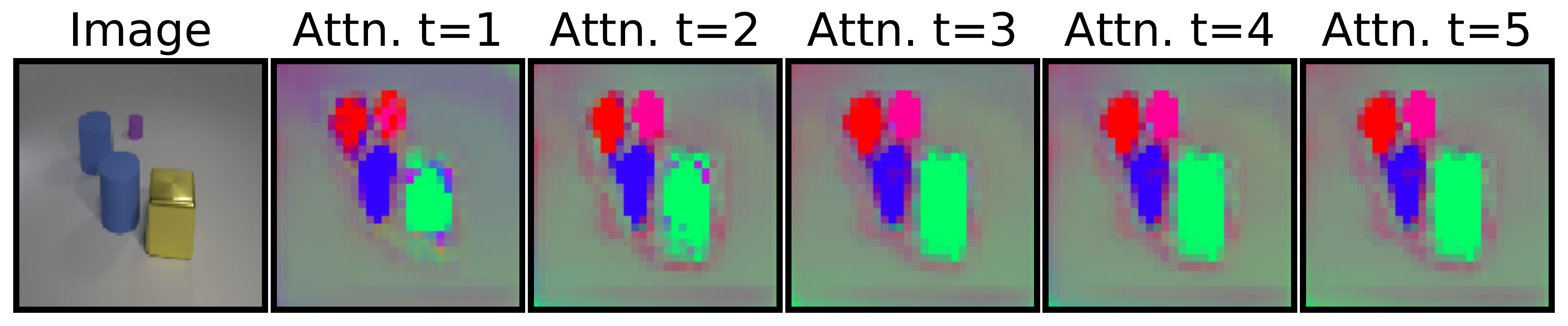}
\resizebox{\textwidth}{!}{%
\begin{tabular}{cccccc}
\toprule
                      True Y &                                               Pred. t=1 &                                               Pred. t=2 &                     Pred. t=3 &                     Pred. t=4 &                     Pred. t=5 \\
\midrule
         (-2.11, -0.69, 0.70) &  (-2.82, -0.19, 0.71), \textcolor{red}{\textbf{d=0.87}} &  (-2.43, -0.22, 0.70), \textcolor{red}{\textbf{d=0.56}} &  (-2.42, -0.55, 0.71), d=0.34 &  (-2.41, -0.35, 0.71), d=0.45 &  (-2.42, -0.48, 0.71), d=0.36 \\
  large blue rubber cylinder &                              large blue rubber cylinder &                              large blue rubber cylinder &    large blue rubber cylinder &    large blue rubber cylinder &    large blue rubber cylinder \\
          (2.41, -0.82, 0.70) &                             (2.52, -0.35, 0.66), d=0.48 &                             (2.57, -0.64, 0.72), d=0.24 &   (2.53, -0.83, 0.71), d=0.12 &   (2.55, -0.79, 0.70), d=0.14 &   (2.54, -0.82, 0.71), d=0.13 \\
      large yellow metal cube &                                 large yellow metal cube &                                 large yellow metal cube &       large yellow metal cube &       large yellow metal cube &       large yellow metal cube \\
          (-2.57, 1.88, 0.35) &   (-2.19, 2.23, 0.37), \textcolor{red}{\textbf{d=0.52}} &                             (-2.70, 2.31, 0.34), d=0.45 &   (-2.57, 2.35, 0.33), d=0.47 &   (-2.58, 2.35, 0.34), d=0.48 &   (-2.58, 2.35, 0.34), d=0.48 \\
 small purple rubber cylinder &                            small purple rubber cylinder &                            small purple rubber cylinder &  small purple rubber cylinder &  small purple rubber cylinder &  small purple rubber cylinder \\
          (0.69, -1.51, 0.70) &   (0.26, -2.05, 0.67), \textcolor{red}{\textbf{d=0.69}} &                             (0.72, -1.47, 0.70), d=0.05 &   (0.72, -1.57, 0.69), d=0.07 &   (0.71, -1.54, 0.69), d=0.03 &   (0.72, -1.55, 0.69), d=0.04 \\
  large blue rubber cylinder &                              large blue rubber cylinder &                              large blue rubber cylinder &    large blue rubber cylinder &    large blue rubber cylinder &    large blue rubber cylinder \\
\bottomrule
\vspace{5mm}
\end{tabular}
}
\includegraphics[width=0.5\textwidth,trim={0 0.4cm 0 0},clip]{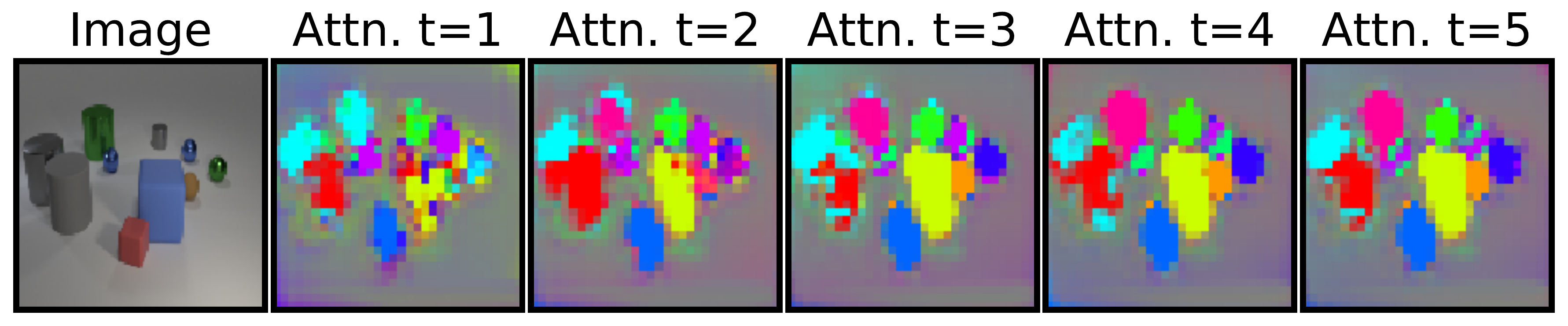}
\resizebox{\textwidth}{!}{%
\begin{tabular}{cccccc}
\toprule
                     True Y &                                                                                                Pred. t=1 &                                                                      Pred. t=2 &                                               Pred. t=3 &                                               Pred. t=4 &                                              Pred. t=5 \\
\midrule
        (-2.92, 0.03, 0.70) &                                                   (-2.36, -1.24, 0.68), \textcolor{red}{\textbf{d=1.39}} &                          (-0.90, 0.35, 0.57), \textcolor{red}{\textbf{d=2.05}} &   (-2.24, 0.16, 0.71), \textcolor{red}{\textbf{d=0.70}} &  (-2.38, -0.04, 0.68), \textcolor{red}{\textbf{d=0.55}} &  (-2.36, 0.03, 0.69), \textcolor{red}{\textbf{d=0.56}} \\
 large green metal cylinder &                                                                               large green metal cylinder &                                                     large green metal cylinder &                              large green metal cylinder &                              large green metal cylinder &                             large green metal cylinder \\
        (-1.41, 2.57, 0.35) &                                                    (-0.47, 2.41, 0.40), \textcolor{red}{\textbf{d=0.95}} &                          (-1.58, 2.11, 0.25), \textcolor{red}{\textbf{d=0.50}} &   (-1.65, 1.99, 0.34), \textcolor{red}{\textbf{d=0.63}} &   (-1.98, 2.10, 0.36), \textcolor{red}{\textbf{d=0.74}} &  (-1.92, 2.24, 0.36), \textcolor{red}{\textbf{d=0.61}} \\
  small gray metal cylinder &                                                                                small gray metal cylinder &                                                      small gray metal cylinder &                               small gray metal cylinder &                               small gray metal cylinder &                              small gray metal cylinder \\
         (0.33, 2.72, 0.35) &                                                     (0.28, 1.34, 0.38), \textcolor{red}{\textbf{d=1.38}} &                           (0.27, 1.06, 0.33), \textcolor{red}{\textbf{d=1.66}} &   (-1.44, 1.56, 0.35), \textcolor{red}{\textbf{d=2.11}} &   (-0.56, 1.16, 0.23), \textcolor{red}{\textbf{d=1.80}} &  (-0.71, 1.24, 0.34), \textcolor{red}{\textbf{d=1.81}} \\
    small blue metal sphere &                                                                                  small blue metal sphere &                                                        small blue metal sphere &                                 small blue metal sphere &                                 small blue metal sphere &                                small blue metal sphere \\
        (2.22, -2.28, 0.35) &                                                                              (2.05, -1.94, 0.37), d=0.38 &                                                    (2.28, -2.02, 0.37), d=0.27 &                             (2.10, -2.03, 0.36), d=0.28 &                             (2.10, -2.01, 0.36), d=0.30 &                            (2.09, -2.01, 0.36), d=0.30 \\
      small red rubber cube &                                                                                    small red rubber cube &                                                          small red rubber cube &                                   small red rubber cube &                                   small red rubber cube &                                  small red rubber cube \\
        (1.99, -0.93, 0.70) &                                                    (2.03, -0.31, 0.68), \textcolor{red}{\textbf{d=0.62}} &                                                    (1.54, -1.04, 0.72), d=0.47 &                             (1.90, -0.95, 0.72), d=0.09 &                             (1.83, -0.90, 0.72), d=0.17 &                            (1.86, -0.91, 0.72), d=0.14 \\
     large blue rubber cube &                                                                                   large blue rubber cube &                                                         large blue rubber cube &                                  large blue rubber cube &                                  large blue rubber cube &                                 large blue rubber cube \\
      (-1.50, -0.34, 0.35) &                                                    (-2.50, 0.14, 0.40), \textcolor{red}{\textbf{d=1.11}} &                          (-2.06, 1.66, 0.28), \textcolor{red}{\textbf{d=2.08}} &   (-0.11, 0.72, 0.33), \textcolor{red}{\textbf{d=1.76}} &   (-0.60, 1.13, 0.35), \textcolor{red}{\textbf{d=1.72}} &  (-0.55, 1.04, 0.32), \textcolor{red}{\textbf{d=1.68}} \\
    small blue metal sphere &                            small \textcolor{red}{\textbf{gray}} metal \textcolor{red}{\textbf{cylinder}} &  small \textcolor{red}{\textbf{gray}} metal \textcolor{red}{\textbf{cylinder}} &                                 small blue metal sphere &                                 small blue metal sphere &                                small blue metal sphere \\
         (1.94, 2.51, 0.35) &                                                    (-1.54, 0.85, 0.41), \textcolor{red}{\textbf{d=3.86}} &                           (1.15, 2.35, 0.31), \textcolor{red}{\textbf{d=0.81}} &                              (1.85, 2.38, 0.37), d=0.16 &                              (1.76, 2.38, 0.38), d=0.23 &                             (1.81, 2.34, 0.37), d=0.22 \\
  small green metal sphere &  \textcolor{red}{\textbf{large}} \textcolor{red}{\textbf{gray}} metal \textcolor{red}{\textbf{cylinder}} &                                                       small green metal sphere &                                small green metal sphere &                                small green metal sphere &                               small green metal sphere \\
      (-2.05, -2.99, 0.70) &                                                   (-0.45, -1.37, 0.43), \textcolor{red}{\textbf{d=2.30}} &                         (-2.53, -2.33, 0.69), \textcolor{red}{\textbf{d=0.82}} &  (-1.53, -2.54, 0.72), \textcolor{red}{\textbf{d=0.69}} &                            (-2.30, -2.61, 0.71), d=0.45 &                           (-2.09, -2.51, 0.70), d=0.48 \\
  large gray metal cylinder &                           large \textcolor{red}{\textbf{blue}} \textcolor{red}{\textbf{rubber}} cylinder &                                                      large gray metal cylinder &    large gray \textcolor{red}{\textbf{rubber}} cylinder &                               large gray metal cylinder &                              large gray metal cylinder \\
      (-0.31, -2.95, 0.70) &                                                    (0.10, -2.59, 0.70), \textcolor{red}{\textbf{d=0.54}} &                                                   (-0.25, -2.50, 0.70), d=0.45 &                            (-0.26, -2.60, 0.69), d=0.35 &                            (-0.26, -2.69, 0.69), d=0.26 &                           (-0.29, -2.64, 0.69), d=0.31 \\
 large gray rubber cylinder &                                                         large gray rubber \textcolor{red}{\textbf{cube}} &                                                     large gray rubber cylinder &                              large gray rubber cylinder &                              large gray rubber cylinder &                             large gray rubber cylinder \\
         (1.81, 0.84, 0.35) &                                                    (1.16, -1.06, 0.37), \textcolor{red}{\textbf{d=2.01}} &                          (0.27, 0.66, -0.17), \textcolor{red}{\textbf{d=1.64}} &    (1.49, 0.32, 0.33), \textcolor{red}{\textbf{d=0.62}} &    (1.40, 0.53, 0.33), \textcolor{red}{\textbf{d=0.52}} &   (1.40, 0.50, 0.33), \textcolor{red}{\textbf{d=0.54}} \\
  small brown rubber sphere &     \textcolor{red}{\textbf{large}} \textcolor{red}{\textbf{blue}} rubber \textcolor{red}{\textbf{cube}} &                                                      small brown rubber sphere &                               small brown rubber sphere &                               small brown rubber sphere &                              small brown rubber sphere \\
\bottomrule
\vspace{5mm}
\end{tabular}
}
\includegraphics[width=0.5\textwidth,trim={0 0.4cm 0 0},clip]{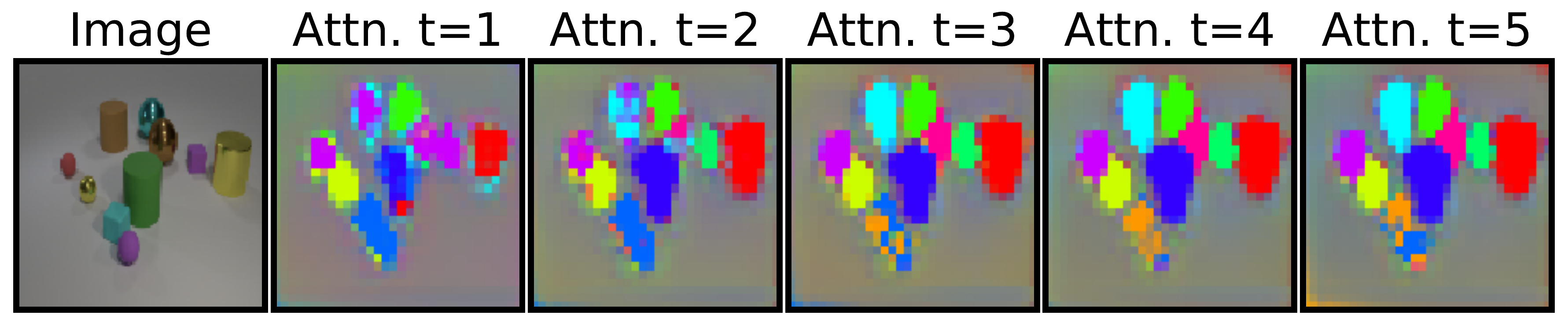}
\resizebox{\textwidth}{!}{%
\begin{tabular}{cccccc}
\toprule
                      True Y &                                                                                                                          Pred. t=1 &                                                Pred. t=2 &                                               Pred. t=3 &                                               Pred. t=4 &                                               Pred. t=5 \\
\midrule
         (-2.28, 2.76, 0.70) &                                                                                                        (-2.40, 2.30, 0.69), d=0.47 &    (-1.96, 2.15, 0.67), \textcolor{red}{\textbf{d=0.69}} &   (-2.01, 2.16, 0.66), \textcolor{red}{\textbf{d=0.66}} &   (-1.99, 2.12, 0.66), \textcolor{red}{\textbf{d=0.70}} &   (-1.98, 2.10, 0.66), \textcolor{red}{\textbf{d=0.73}} \\
     large cyan metal sphere &                                                                                                            large cyan metal sphere &                                  large cyan metal sphere &                                 large cyan metal sphere &                                 large cyan metal sphere &                                 large cyan metal sphere \\
          (0.93, 2.56, 0.35) &                                                                               (0.00, 1.31, 0.37), \textcolor{red}{\textbf{d=1.55}} &                               (0.60, 2.43, 0.33), d=0.35 &                              (0.85, 2.28, 0.33), d=0.29 &                              (0.76, 2.39, 0.33), d=0.24 &                              (0.71, 2.39, 0.33), d=0.28 \\
    small purple rubber cube &  \textcolor{red}{\textbf{large}} \textcolor{red}{\textbf{blue}} \textcolor{red}{\textbf{metal}} \textcolor{red}{\textbf{cylinder}} &                                 small purple rubber cube &                                small purple rubber cube &                                small purple rubber cube &                                small purple rubber cube \\
         (2.27, -2.44, 0.35) &                                                                             (-1.14, -1.29, 0.31), \textcolor{red}{\textbf{d=3.60}} &  (-2.61, -2.59, -2.50), \textcolor{red}{\textbf{d=5.65}} &                             (2.10, -2.58, 0.35), d=0.22 &  (0.81, -1.73, -0.11), \textcolor{red}{\textbf{d=1.69}} &                             (2.46, -2.34, 0.38), d=0.22 \\
  small purple rubber sphere &  \textcolor{red}{\textbf{large}} \textcolor{red}{\textbf{cyan}} \textcolor{red}{\textbf{metal}} \textcolor{red}{\textbf{cylinder}} &      small \textcolor{red}{\textbf{green}} rubber sphere &      small \textcolor{red}{\textbf{cyan}} rubber sphere &                              small purple rubber sphere &                              small purple rubber sphere \\
        (-0.70, -2.14, 0.35) &                                                                              (0.76, -1.79, 0.41), \textcolor{red}{\textbf{d=1.50}} &                             (-0.44, -1.97, 0.36), d=0.31 &  (-0.17, -1.94, 0.36), \textcolor{red}{\textbf{d=0.57}} &                            (-0.31, -1.93, 0.35), d=0.44 &                            (-0.28, -1.95, 0.35), d=0.46 \\
  small yellow metal sphere &                                                                                                          small yellow metal sphere &                                small yellow metal sphere &                               small yellow metal sphere &                               small yellow metal sphere &                               small yellow metal sphere \\
         (-0.46, 1.75, 0.70) &                                                                              (-1.03, 0.70, 0.25), \textcolor{red}{\textbf{d=1.27}} &                              (-0.46, 1.61, 0.52), d=0.23 &   (-0.45, 2.25, 0.66), \textcolor{red}{\textbf{d=0.50}} &   (-0.83, 2.20, 0.65), \textcolor{red}{\textbf{d=0.58}} &   (-0.70, 2.29, 0.65), \textcolor{red}{\textbf{d=0.59}} \\
    large brown metal sphere &                                                                                large \textcolor{red}{\textbf{yellow}} metal sphere &                                 large brown metal sphere &                                large brown metal sphere &                                large brown metal sphere &                                large brown metal sphere \\
         (1.14, -0.91, 0.70) &                                                                                                        (0.73, -0.68, 0.68), d=0.46 &                              (0.66, -0.88, 0.71), d=0.48 &                             (0.74, -0.96, 0.70), d=0.40 &                             (0.73, -0.91, 0.70), d=0.41 &                             (0.74, -0.95, 0.69), d=0.40 \\
 large green rubber cylinder &                                                                                                        large green rubber cylinder &                              large green rubber cylinder &                             large green rubber cylinder &                             large green rubber cylinder &                             large green rubber cylinder \\
         (-2.98, 0.78, 0.70) &                                                                              (-0.94, 1.14, 0.55), \textcolor{red}{\textbf{d=2.08}} &    (-2.56, 1.25, 0.66), \textcolor{red}{\textbf{d=0.63}} &                             (-2.71, 0.56, 0.68), d=0.35 &                             (-2.73, 1.01, 0.68), d=0.34 &                             (-2.73, 0.67, 0.68), d=0.27 \\
 large brown rubber cylinder &                                                                               large brown \textcolor{red}{\textbf{metal}} cylinder &                              large brown rubber cylinder &                             large brown rubber cylinder &                             large brown rubber cylinder &                             large brown rubber cylinder \\
        (-2.51, -2.02, 0.35) &                                                                              (-2.39, 2.31, 0.57), \textcolor{red}{\textbf{d=4.34}} &   (-2.81, -0.77, 0.36), \textcolor{red}{\textbf{d=1.29}} &  (-2.35, -1.36, 0.33), \textcolor{red}{\textbf{d=0.68}} &  (-2.25, -1.45, 0.32), \textcolor{red}{\textbf{d=0.63}} &  (-2.26, -1.45, 0.32), \textcolor{red}{\textbf{d=0.63}} \\
     small red rubber sphere &                                                          \textcolor{red}{\textbf{large}} red rubber \textcolor{red}{\textbf{cube}} &                                  small red rubber sphere &                                 small red rubber sphere &                                 small red rubber sphere &                                 small red rubber sphere \\
         (1.30, -2.20, 0.35) &                                                                              (2.27, -2.65, 0.37), \textcolor{red}{\textbf{d=1.07}} &    (2.24, -2.76, 0.36), \textcolor{red}{\textbf{d=1.09}} &   (1.86, -2.38, 0.37), \textcolor{red}{\textbf{d=0.58}} &   (2.03, -2.66, 0.35), \textcolor{red}{\textbf{d=0.85}} &                             (1.41, -2.55, 0.34), d=0.36 \\
      small cyan rubber cube &                                                                                                             small cyan rubber cube &                                   small cyan rubber cube &                                  small cyan rubber cube &                                  small cyan rubber cube &                                  small cyan rubber cube \\
          (2.50, 2.80, 0.70) &                                                                               (2.59, 1.99, 0.72), \textcolor{red}{\textbf{d=0.81}} &                               (2.57, 2.72, 0.75), d=0.12 &                              (2.61, 2.52, 0.75), d=0.30 &                              (2.60, 2.52, 0.74), d=0.30 &                              (2.59, 2.51, 0.73), d=0.30 \\
 large yellow metal cylinder &                                                                                                        large yellow metal cylinder &                              large yellow metal cylinder &                             large yellow metal cylinder &                             large yellow metal cylinder &                             large yellow metal cylinder \\
\bottomrule
\end{tabular}
}
\end{table}

\begin{table*}[htp!]
\centering
\vspace{1em}
\caption{Average Precision at different distance thresholds on CLEVR10 (in $\%$, mean $\pm$ std for 5 seeds). We highlighted the best result for each threshold within confidence intervals.}
\resizebox{0.75\textwidth}{!}{%
\begin{tabular}{l c c c c c }
\toprule
& $AP_\infty$ & $AP_1$ & $AP_{0.5}$ & $AP_{0.25}$ & $AP_{0.125}$ \\
\midrule
\sam & \textbf{94.3} $\pm$ \textbf{1.1} & \textbf{86.7} $\pm$ \textbf{1.4} & \textbf{56.0} $\pm$ \textbf{3.6} & \textbf{10.8} $\pm$ \textbf{1.7} & \textbf{0.9} $\pm$ \textbf{0.2}\\
DSPN T=30 & 85.2 $\pm$ 4.8 & \textbf{81.1} $\pm$ \textbf{5.2} & \textbf{47.4} $\pm$ \textbf{17.6} & \textbf{10.8} $\pm$ \textbf{9.0} & \textbf{0.6} $\pm$ \textbf{0.7}\\
DSPN T=10 & 72.8 $\pm$ 2.3 & 59.2 $\pm$ 2.8 & 39.0 $\pm$ 4.4 & \textbf{12.4} $\pm$ \textbf{2.5} & \textbf{1.3} $\pm$ \textbf{0.4}\\
 Slot MLP & 19.8 $\pm$ 1.6 & 1.4 $\pm$ 0.3 & 0.3 $\pm$ 0.2 & 0.0 $\pm$ 0.0 & 0.0 $\pm$ 0.0\\
\bottomrule
\end{tabular}
}
\label{table:set_pred_clevr}
\end{table*}

\paragraph{Numerical results} To facilitate comparison with our method, we report the results of Figure~5 of the main paper (left subfigure) in numerical form in Table~\ref{table:set_pred_clevr} as well as the performance of DSPN~\citep{zhang2019deep} with 10 iterations (as opposed to 30). We note that our approach has generally higher average $AP$ compared to DSPN and lower variance. We remark that the published implementation of DSPN uses a significantly deeper image encoder than our model: ResNet 34~\citep{he2016deep} vs.~a CNN with 4 layers. Further, we use the same scale for all properties (each coordinate in the prediction vector is in $[0, 1]$), while in DSPN the object-coordinates are rescaled to $[-1, 1]$ and every other property is in $[0, 1]$.

\begin{wrapfigure}{r}{0.35\textwidth}
    \vspace{-0.5em}
    \centering
    \includegraphics[width=0.35\textwidth,trim={0 0.3cm 0 0.25cm},clip]{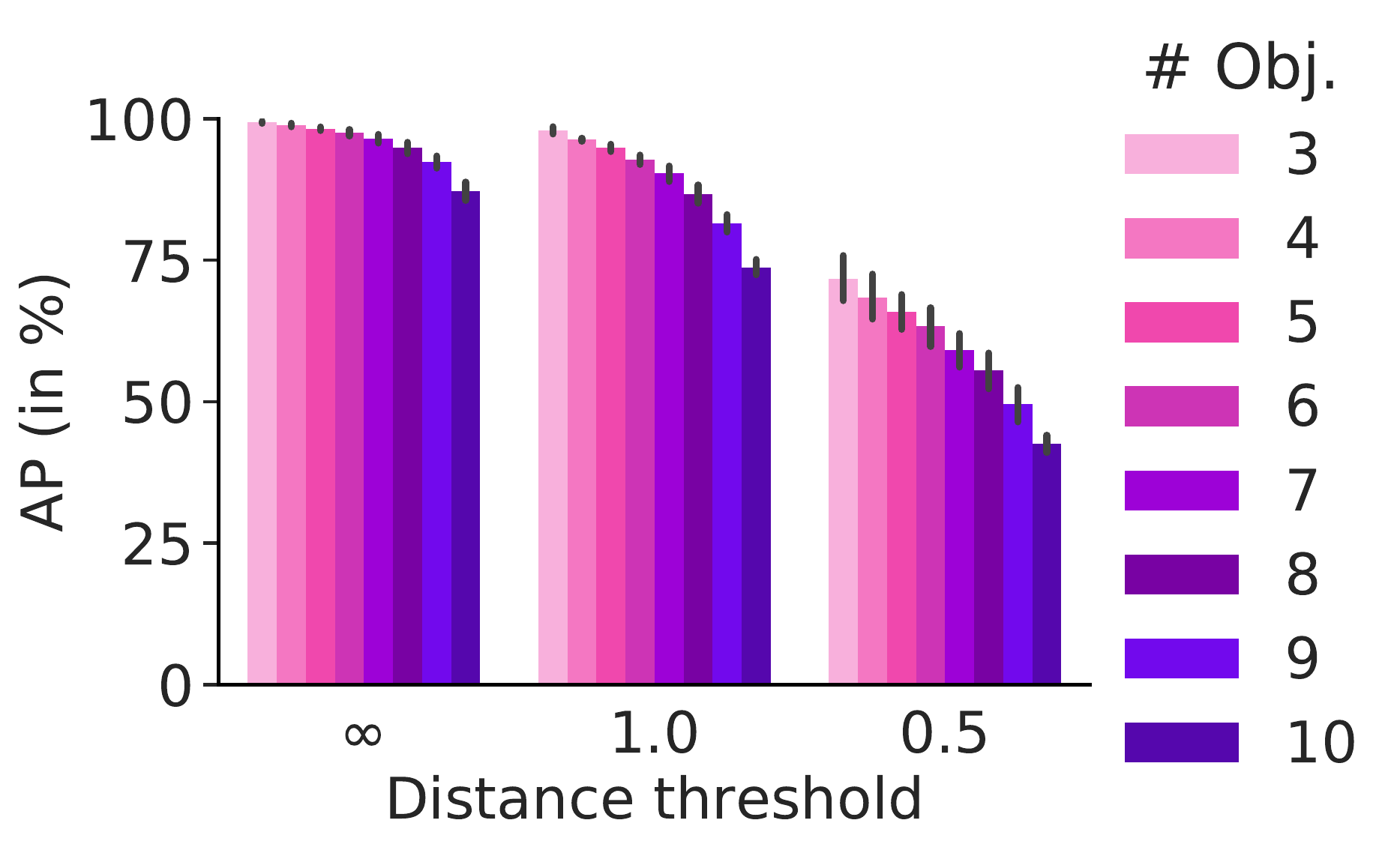}
    \caption{AP score by \# of objects.}
    \label{fig:stratified_obj}
    \vspace{-1em}
\end{wrapfigure}
\paragraph{Results partitioned by number of objects} Here, we break down the results from Table~\ref{table:set_pred_clevr} for the Slot Attention model into separate bins that measure the AP score solely for images with a certain fixed number of objects. This is different from Figure 5 (right subfigure) in the main paper, where we test generalization to more objects at test time. We can observe that the rate of mistakes increases with the number of objects in the scene.

To analyse to what degree this can be addressed by increasing the number of iterations that are used in the Slot Attention module, we run the same experiment where we increase the number of iterations at test time from 3 to 5 iterations for a model trained with 3 iterations. We can see that increasing the number of iterations significantly improves results for difficult scenes with many objects, whereas this has little effect for scenes with only a small number of objects.

\begin{figure}[h!]
\centering
    \begin{subfigure}[t]{0.32\textwidth}
    \includegraphics[scale=0.245,trim={0 -1.19cm 0 0},clip]{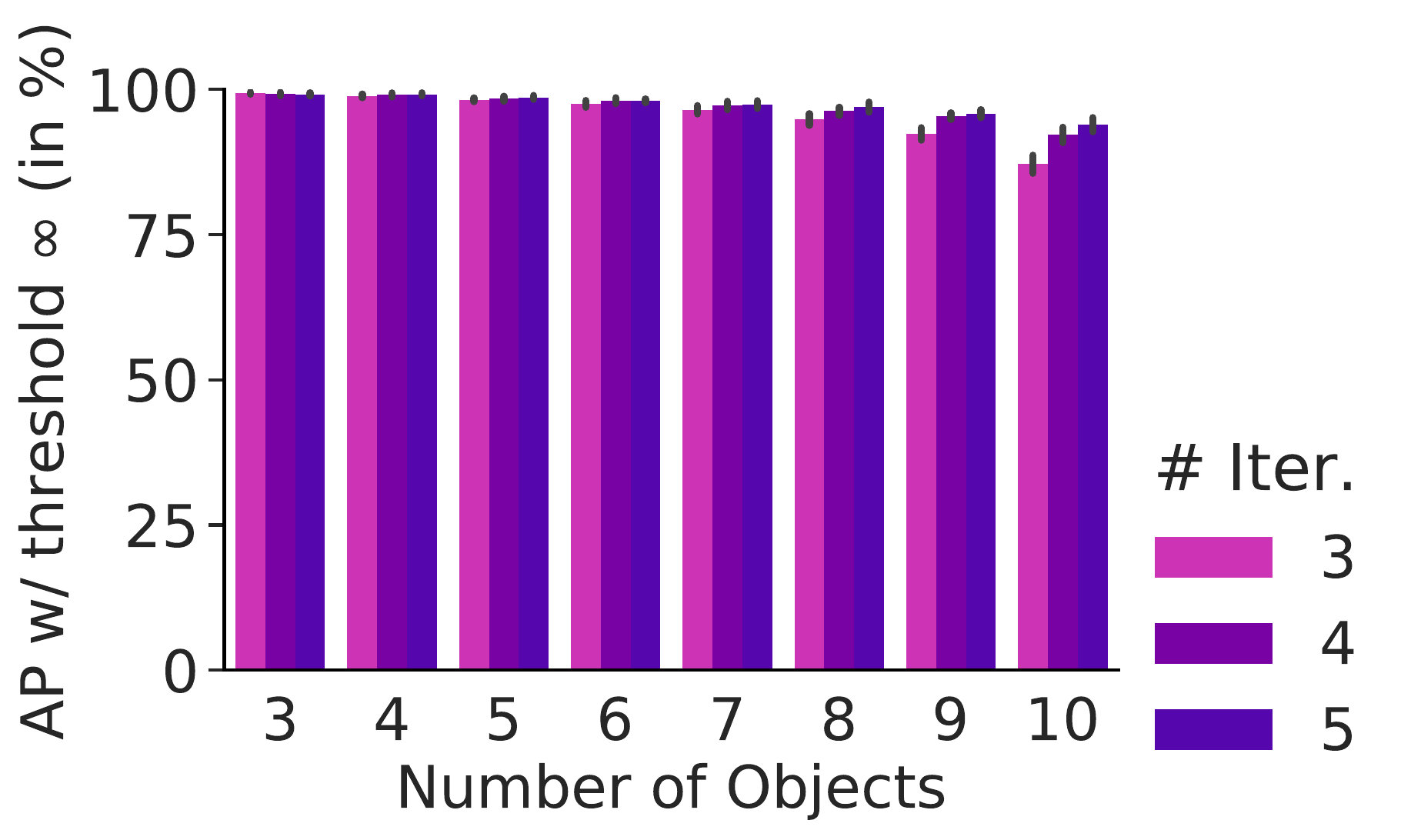}
    \end{subfigure}
    ~
    \begin{subfigure}[t]{0.32\textwidth}
    \includegraphics[scale=0.245,trim={0 -1.19cm 0 0},clip]{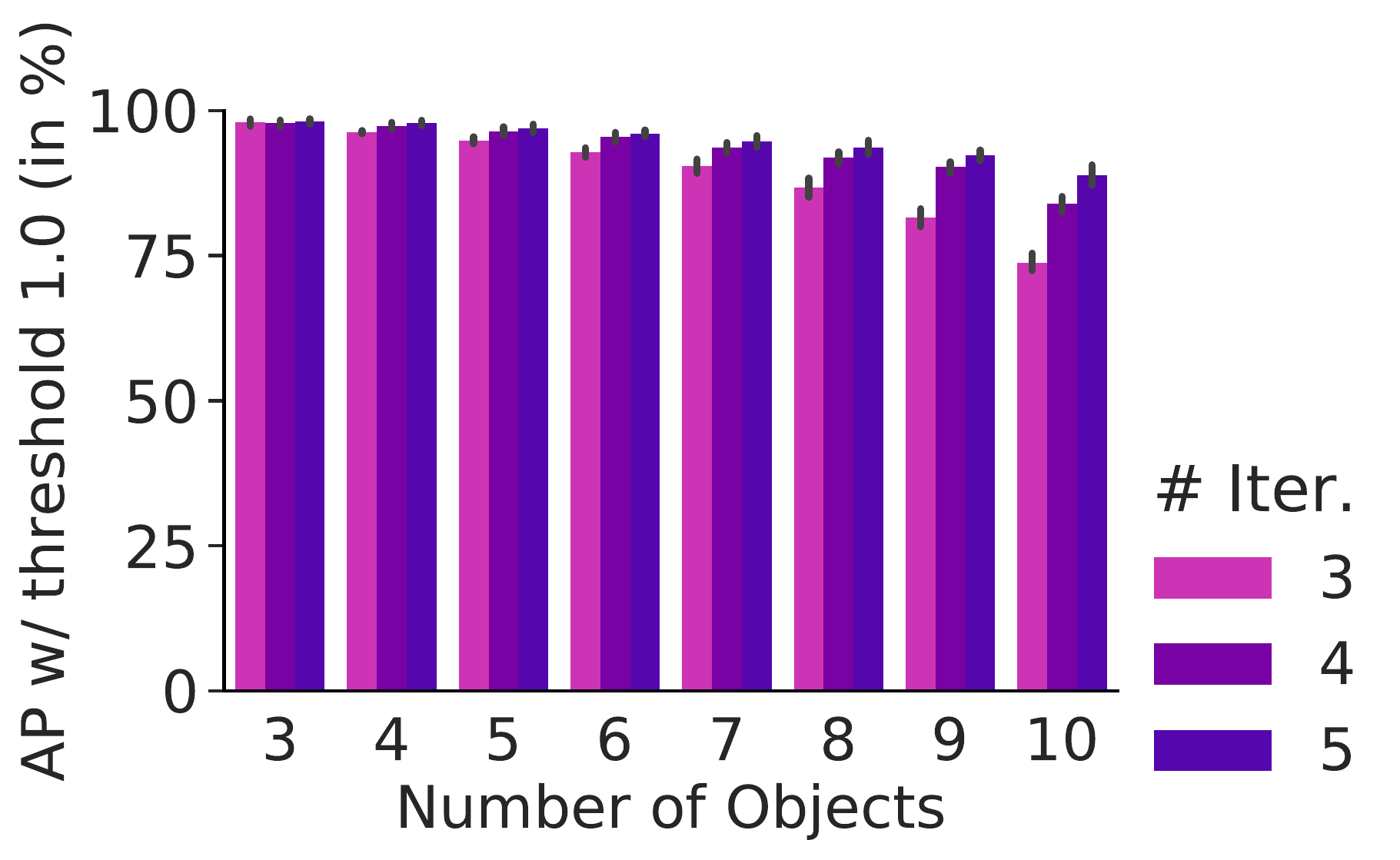}
    \end{subfigure}
    ~
    \begin{subfigure}[t]{0.32\textwidth}
    \includegraphics[scale=0.245,trim={0 -1.19cm 0 0},clip]{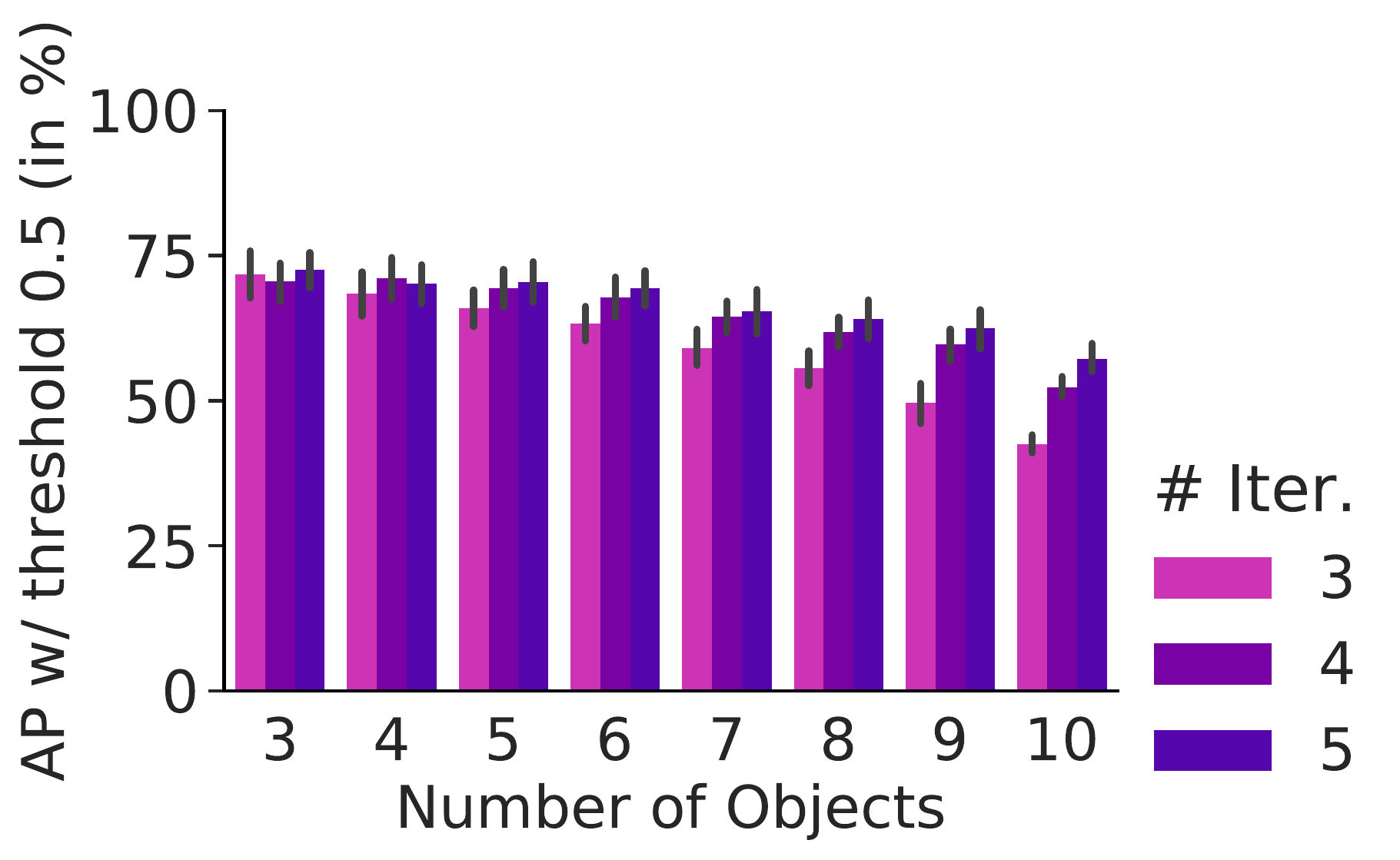}
    \end{subfigure}
    \caption{AP scores binned by number of objects in the scene. Difficult scenes that contain many objects require more Slot Attention iterations.}
    \label{fig:stratified_iter}
\end{figure}

\begin{figure}[htp!]
    \centering
    \includegraphics[width=0.85\textwidth,trim={0 0.3cm 0 0.25cm},clip]{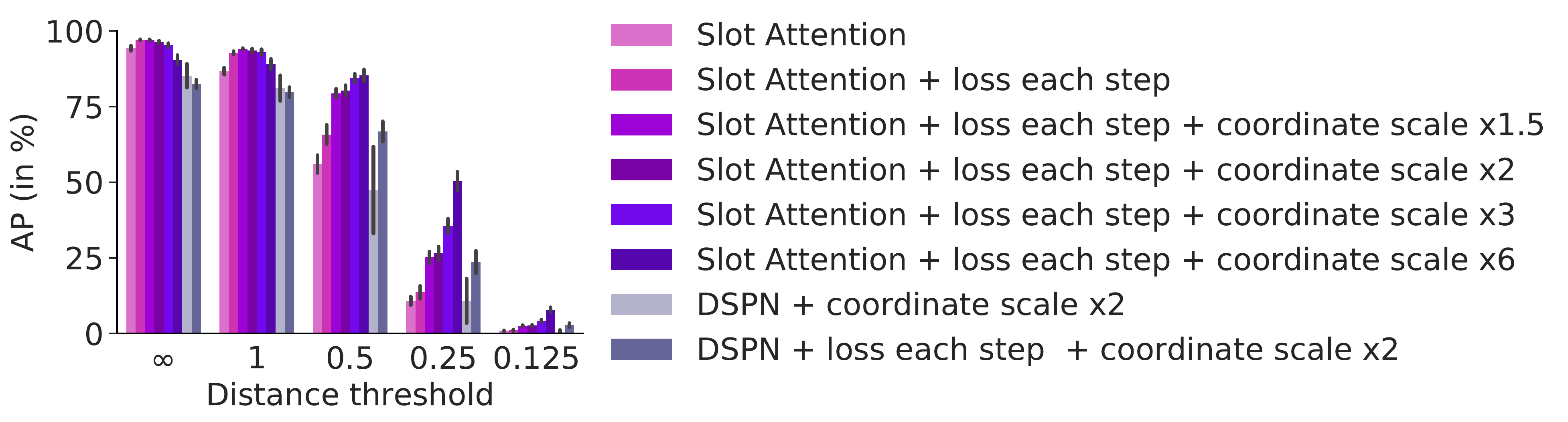}
    \caption{Computing the loss at each iteration generally improves results for both Slot Attention and the DSPN (while increasing the computational cost as well). As expected, re-scaling the coordinate to have a higher weight in the loss, positively impacts the AP at small distance thresholds where the position of objects needs to be predicted more accurately.}
    \label{fig:prop_pred_hist_loss}
\end{figure}

\paragraph{Step-wise loss \& coordinate scaling} We investigate a variant of our model where we apply the set prediction component and loss at every iteration of the attention mechanism, as opposed to only after the final step. A similar experiment was reported in~\citep{zhang2019deep} for the DSPN model. As DSPN uses a different scale for the position coordinates of objects by default, we further compare against a version of our model where we similarly use a different scale. Using a different scale for the object location increases its weight in the loss. We observe the effect of the coordinate scale and of computing the loss at each step in Figure~\ref{fig:prop_pred_hist_loss}.

A scale of 1 corresponds to our default coordinate normalization of $[0, 1]$, whereas larger scales correspond to a $[0, \texttt{scale}]$ normalization of the coordinates (or shifted by an arbitrary constant). Overall, we observe that computing the loss at each step in Slot Attention improves the AP score at all distance thresholds as opposed to DSPN, where it is only beneficial at small distance thresholds. We conjecture that this is an optimization issue in DSPN. As expected, increasing the importance of accurately modeling position in the loss impacts the AP positively at smaller distance thresholds, but can otherwise have a negative effect on predicting other object attributes correctly.

\section{Permutation Invariance and Equivariance}\label{app:proof}
\subsection{Definitions}
Before giving the proof for Proposition~1, we formally define permutation invariance and equivariance.
\begin{definition}[Permutation Invariance]
A function $f: \mathbb{R}^{M\times D_1} \rightarrow \mathbb{R}^{M\times D_2}$ is \emph{permutation invariant} if for any arbitrary permutation matrix $\pi\in\mathbb{R}^{M\times M}$ it holds that:
\begin{equation*}
    f(\pi x) = f(x) \, .
\end{equation*}
\end{definition}
\begin{definition}[Permutation Equivariance]
A function $f: \mathbb{R}^{M\times D_1} \rightarrow \mathbb{R}^{M\times D_2}$ is \emph{permutation equivariant} if for any arbitrary permutation matrix $\pi\in\mathbb{R}^{M\times M}$ it holds that:
\begin{equation*}
    f(\pi x) = \pi f(x) \, .
\end{equation*}
\end{definition}
\subsection{Proof}

The proof is straightforward and is reported for completeness. We rely on the fact that the sum operation is permutation invariant.
\paragraph{Linear projections} As the linear projections are applied independently per slot/input element with shared parameters, they are permutation equivariant.

\paragraph{Equation~1}
The dot product of the attention mechanism (i.e. computing the matrix $M\in\mathbb{R}^{N\times K}$) involves a sum over the feature axis (of dimension $D$) and is therefore permutation equivariant w.r.t.~both input and slots. The output of the softmax is also equivariant, as:
\begin{align*}
    \softmax{\pi_s\cdot \pi_i \cdot M}_{(k,l)} &= \frac{e^{(\pi_s\cdot \pi_i \cdot M)_{k,l}}}{\sum_s e^{(\pi_s\cdot \pi_i \cdot M)_{k,s}}} \\&=  \frac{e^{M_{\pi_i(k),\pi_s(l)}}}{\sum_{\pi_s(l)} e^{M_{\pi_i(k),\pi_s(l)}}} \\&= \softmax{M}_{(\pi_i(k),\pi_s(l))}\,,
\end{align*}
where we indicate with e.g. $\pi_i(k)$ the transformation of the coordinate $k$ with the permutation matrix $\pi_i$. The second equality follows from the fact that the sum is permutation invariant.

\paragraph{Equation~2}
The matrix product in the computation of the $\updates$ involves a sum over the input elements which makes the operation invariant w.r.t.~permutations of the input order (i.e. $\pi_i$) and equivariant w.r.t.~the slot order (i.e. $\pi_s$).

\paragraph{Slot update:} The slot update applies the same network to each slot with shared parameters. Therefore, it is a permutation equivariant operation w.r.t.~the slot order.
\paragraph{Combining all steps:}
As all steps in the algorithms are permutation equivariant wrt $\pi_s$, the overall module is permutation equivariant. On the other hand, Equation~2 is permutation invariant w.r.t.~to $\pi_i$. Therefore, after the first iteration the algorithm becomes permutation invariant w.r.t.~the input order.

\section{Implementation and Experimental Details}\label{app:implementation_details}
For the \sam module we use a slot feature dimension of $D=D_{\slots}=64$. The GRU has $64$-dimensional hidden state and the feedforward block is a MLP with single hidden layer of size 128 and ReLU activation followed by a linear layer.

\subsection{CNN Encoder}\label{sec:app_cnn_enc}
The CNN Encoder used in our experiments is depicted in Table~\ref{table:architecture_cnn_enc} for CLEVR and Table~\ref{table:architecture_cnn_enc_tetr} for Tetrominoes and Multi-dSprites.
For the property prediction task on CLEVR, we reduce the size of the representation by using strides in the CNN backbone. All convolutional layers use padding \texttt{SAME} and have a bias weight. After this backbone, we add position embeddings (Section~\ref{app:pos_embedd}) and then flatten the spatial dimensions. After applying a layer normalization, we finally add $1\times 1$ convolutions which we implement as a shared MLP applied at each spatial location with one hidden layer of $64$ units (32 for Tetrominoes and Multi-dSprites) with ReLU non-linearity followed by a linear layer with output dimension of $64$ (32 for Tetrominoes and Multi-dSprites).

\begin{table*}[htp!]
\centering
\caption{CNN encoder for the experiments on CLEVR. For the property prediction experiments on CLEVR10 we use stride of 2 on the layers marked with * which decreases the memory footprint.}
\begin{tabular}{ccccc}
\toprule
Type & Size/Channels & Activation & Comment \\
\midrule
Conv $5\times 5$ & $64$ & ReLU & stride: $1$\\
Conv $5\times 5$ & $64$ & ReLU & stride: $1*$\\
Conv $5\times 5$ & $64$ & ReLU & stride: $1*$\\
Conv $5\times 5$ & $64$ & ReLU & stride: $1$\\
Position Embedding & - & - & See Section~\ref{app:pos_embedd}\\
Flatten & axis: $[0,1\times 2, 3]$ & - & flatten x, y pos. \\
Layer Norm & - & - & -  \\
MLP (per location) & 64 & ReLU & - \\
MLP (per location) & 64 & - & - \\
\bottomrule
\end{tabular}
\label{table:architecture_cnn_enc}
\end{table*}

\begin{table*}[htp!]
\centering
\caption{CNN encoder for the experiments on Tetrominoes and Multi-dSprites.}
\begin{tabular}{ccccc}
\toprule
Type & Size/Channels & Activation & Comment \\
\midrule
Conv $5\times 5$ & $32$ & ReLU & stride: $1$\\
Conv $5\times 5$ & $32$ & ReLU & stride: $1$\\
Conv $5\times 5$ & $32$ & ReLU & stride: $1$\\
Conv $5\times 5$ & $32$ & ReLU & stride: $1$\\
Position Embedding & - & - &  See Section~\ref{app:pos_embedd} \\
Flatten & axis: $[0,1\times 2, 3]$ & - & flatten x, y pos. \\
Layer Norm & - & - & - \\
MLP (per location) & 32 & ReLU & - \\
MLP (per location) & 32 & - & - \\
\bottomrule
\end{tabular}
\label{table:architecture_cnn_enc_tetr}
\end{table*}

\subsection{Positional Embedding}\label{app:pos_embedd}
As \sam is invariant with respect to the order of the input elements (i.e., it treats the input as a set of vectors, even if it is an image), position information is not directly accessible. In order to give \sam access to position information, we augment input features (CNN feature maps) with positional embeddings as follows: (i) We construct a $W\times H\times 4$ tensor, where $W$ and $H$ are width and height of the CNN feature maps, with a linear gradient $[0,1]$ in each of the four cardinal directions. In other words, each point on the grid is associated with a 4-dimensional feature vector that encodes its distance (normalized to $[0,1]$) to the borders of the feature map along each of the four cardinal directions. (ii) We project each feature vector to the same dimensionality as the image feature vectors (i.e., number of feature maps) using a learnable linear map and add the result to the CNN feature maps.

\subsection{Deconvolutional Slot Decoder}
\label{sec:app_slot_dec}
For the object discovery task, our architecture is based on an auto-encoder, where we decode the respresentations produced by Slot Attention with the help of a slot-wise spatial broadcast decoder~\citep{watters2019spatial} with shared parameters between slots. Each spatial broadcast decoder produces an output of size width$\times$height$\times4$, where the first 3 output channels denote RGB channels of the reconstructed image and the last output channel denotes a predicted alpha mask, that is later used to recombine individual slot reconstructions into a single image. The overall architecture for used for CLEVR is described in Table~\ref{table:architecture_cnn_dec_clevr} and for Tetrominoes and Multi-dSprites in Table~\ref{table:architecture_cnn_dec_tetr}.

\paragraph{Spatial broadcast decoder}
The spatial broadcast decoder~\citep{watters2019spatial} is applied independently on each slot representation with shared parameters between slots. We first copy the slot representation vector of dimension $D_{\slots}$ onto a grid of shape width$\times$height$\times D_{\slots}$, after which we add a positional embedding (see Section~\ref{app:pos_embedd}). Finally, this representation is passed through several de-convolutional layers.

\paragraph{Slot recombination}
The final output of the spatial broadcast decoder for each slot is of shape width$\times$height$\times4$ (ignoring the slot and batch dimension). We first split the final channels into three RGB channels and an alpha mask channel. We apply a softmax activation function \textit{across} slots on the alpha masks and lastly recombine all individual slot-based reconstructions into a single reconstructed image by multiplying each alpha mask with each respective reconstructed image (per slot) and lastly by performing a sum reduction on this respective output over the slot dimension to arrive at the final reconstructed image. For visualization of the reconstruction masks in a single image, we replace each individual reconstructed image (per slot) with a unique slot-specific color (see, e.g., third column in Figure~\ref{fig:app_clevr_11_slots}).

\begin{table*}[htp!]
\centering
\caption{Deconv-based slot decoder for the experiments on CLEVR.}
\begin{tabular}{ccccc}
\toprule
Type & Size/Channels & Activation & Comment \\
\midrule
Spatial Broadcast & 8$\times$8 & - &  - \\
Position Embedding & - & - &  See Section~\ref{app:pos_embedd} \\
Conv $5\times 5$ & $64$ & ReLU & stride: $2$\\
Conv $5\times 5$ & $64$ & ReLU & stride: $2$\\
Conv $5\times 5$ & $64$ & ReLU & stride: $2$\\
Conv $5\times 5$ & $64$ & ReLU & stride: $2$\\
Conv $5\times 5$ & $64$ & ReLU & stride: $1$\\
Conv $3\times 3$ & 4 & - & stride: $1$\\
Split Channels & RGB (3), alpha mask (1) & Softmax (on alpha masks) & - \\
Recombine Slots & - & - & - \\
\bottomrule
\end{tabular}
\label{table:architecture_cnn_dec_clevr}
\end{table*}

\begin{table*}[htp!]
\centering
\caption{Deconv-based slot decoder for the experiments on Tetrominoes and Multi-dSprites.}
\begin{tabular}{ccccc}
\toprule
Type & Size/Channels & Activation & Comment \\
\midrule
Spatial Broadcast & width$\times$height & - &  - \\
Position Embedding & - & - &  See Section~\ref{app:pos_embedd} \\
Conv $5\times 5$ & $32$ & ReLU & stride: $1$\\
Conv $5\times 5$ & $32$ & ReLU & stride: $1$\\
Conv $5\times 5$ & $32$ & ReLU & stride: $1$\\
Conv $3\times 3$ & 4 & - & stride: $1$\\
Split Channels & RGB (3), alpha mask (1) & Softmax (on alpha masks) & - \\
Recombine Slots & - & - & - \\
\bottomrule
\end{tabular}
\label{table:architecture_cnn_dec_tetr}
\end{table*}

\subsection{Set Prediction Architecture}
\label{sec:app_set_pred_arch}
For the property prediction task, we apply a MLP on each slot (with shared parameters between slots) and train the overall network with the Huber loss following~\cite{zhang2019deep}. The Huber loss takes the form of a squared error $0.5x^2$ for values $|x|<1$ and a linearly increasing error with slope 1 for $|x|\geq1$. The MLP has one hidden layer with 64 units and ReLU.

The output of this MLP uses a sigmoid activation as we one-hot encode the discrete features and normalize continuous features between $[0, 1]$. The overall network is presented in Table~\ref{table:architecture_mlp_pp}
\begin{table*}[htp!]
\centering
\caption{MLP for the property prediction experiments.}
\begin{tabular}{ccccc}
\toprule
Type & Size/Channels & Activation  \\
\midrule
MLP (per slot) & 64 & ReLU  \\
MLP (per slot) & output size & Sigmoid   \\
\bottomrule
\end{tabular}
\label{table:architecture_mlp_pp}
\end{table*}

\subsection{Slot MLP Baseline}
\label{sec:app_slot_mlp}
For the Slot MLP baseline we predict the slot representation with a MLP as shown in Tables~\ref{table:architecture_slot_mlp_set_pred} and \ref{table:architecture_slot_mlp_obj_discovery}. This module replaces our Slot Attention module and is followed by the same decoder/classifier. Note that we resize images to $16\times16$ before flattening them into a single feature vector to reduce the number of parameters in the MLP.

\begin{table*}[htp!]
\centering
\caption{Slot MLP architecture for set prediction. This block replaces the Slot Attention module.}
\begin{tabular}{ccc}
\toprule
Type & Size/Channels & Activation  \\
\midrule
Resize & $16\times 16$ & -  \\
Flatten & - & -  \\
MLP & 512 & ReLU   \\
MLP & 512 & ReLU \\
MLP & slot size $\times$ num slots & - \\
Reshape & [slot size, num slots]   & -\\
\bottomrule
\end{tabular}
\label{table:architecture_slot_mlp_set_pred}
\end{table*}

\begin{table*}[htp!]
\centering
\caption{Slot MLP architecture for object discovery. This block replaces the Slot Attention module. We use a deeper MLP with more hidden units and a separate slot-wise MLP with shared parameters in this setting, as we found that it significantly improves performance compared to a simpler MLP baseline on the object discovery task.}
\begin{tabular}{ccc}
\toprule
Type & Size/Channels & Activation  \\
\midrule
Resize & $16\times 16$ & -  \\
Flatten & - & -  \\
MLP & 512 & ReLU   \\
MLP & 1024 & ReLU \\
MLP & 1024 & ReLU \\
MLP & slot size $\times$ num slots & - \\
Reshape & [slot size, num slots]   & -\\
MLP (per slot) & 64  & ReLU \\
MLP (per slot) & 64  & -\\
\bottomrule
\end{tabular}
\label{table:architecture_slot_mlp_obj_discovery}
\end{table*}

\subsection{Other Hyperparameters}
\label{sec:app_hyp}
All shared hyperparameters common to each experiments can be found in Table~\ref{table:other_hyp}. The hyperparameters specific to the object discovery and property prediction experiments can be found in Tables~\ref{table:od_hyp} and~\ref{table:pp_hyp} respectively.

In both experiments, we use a learning rate warm-up and exponential decay schedules. For the learning rate warm-up, we linearly increase the learning rate from zero to the final learning rate during the first steps of training. For the decay, we decrease the learning rate by multiplying it by an exponentially decreasing decay rate:
\begin{align*}
    \textnormal{learning\_rate} * \textnormal{decay\_rate}^{(\textnormal{step} / \textnormal{decay\_steps})}
\end{align*}
where the decay rate governs how much we decrease the learning rate. See Table~\ref{table:all_hyperparams} for the parameters of the two schedules.

\subsection{Hyperparameter Optimization}
We started with an architecture and a hyperparameter setting close to that of~\citep{greff2019multi}. We tuned hyperparameters on the object discovery task based on the achieved ARI score on a small subset of training images (320) from CLEVR. We only considered $5$ values for the learning rate $[1e-4, 4e-4, 2e-4, 4e-5, 1e-5]$ and batch sizes of $[32, 64, 128]$.
For property prediction, we took the same learning rate as in object discovery and we computed the AP on a small subset of training images (500). We considered batch sizes of $[64, 128, 512]$ (as we were able to fit larger batches onto a single GPU due to the lower memory footprint of this model).

\begin{table*}
\centering
\caption{Other hyperparameters for all experiments.\label{table:all_hyperparams} }
 \begin{subtable}[t]{0.32\linewidth}
 \centering
 \small
 \caption{Shared hyperparameters. }
\label{table:other_hyp}
 \begin{tabular}{cc}
\toprule
Name & Value  \\
\midrule
\texttt{attn}: $\epsilon$ & 1e-08\\
Adam: $\beta_1$ & 0.9\\
Adam: $\beta_2$ & 0.999\\
Adam: $\epsilon$ & 1e-08\\
Adam: learning rate & 0.0004 \\
Exponential decay & rate 0.5\\
Slot dim. & 64\\
\bottomrule
\end{tabular}
\end{subtable}%
\hspace{5mm}
\begin{subtable}[t]{0.29\linewidth}
\centering
\small
\caption{Hyperparameters for object discovery.}\label{table:od_hyp}
\vspace{2mm}
\begin{tabular}{cc}
\toprule
Name & Value  \\
\midrule
Warmup iters. & 10K\\
Decay steps & 100K\\
Batch size & 64\\
Train steps & 500K\\
\bottomrule
\end{tabular}
\end{subtable}%
\hspace{5mm}
\begin{subtable}[t]{0.29\linewidth}
\centering
\small
\caption{Hyperparameters for property prediction.}\label{table:pp_hyp}
\vspace{2mm}
\begin{tabular}{cc}
\toprule
Name & Value  \\
\midrule
Warmup iters. & 1K\\
Decay steps & 50K\\
Batch size & 512\\
Train steps & 150K\\
\bottomrule
\end{tabular}
\end{subtable}
\end{table*}

\subsection{Datasets}
\label{sec:app_datasets}

\paragraph{Set Prediction}
\looseness=-1 We use the CLEVR~\citep{johnson2017clevr} dataset, which consists of rendered scenes containing multiple objects. Each object has annotations for position ($x, \ y, \ z)$ coordinates in $[-3, 3]$), color (\num{8} possible values), shape (\num{3} possible values), material, and size (\num{2} possible values). The number of objects varies between \num{3} and \num{10} and, similarly to~\citep{zhang2019deep}, we zero-pad the targets so that their number is constant in the batch and add an extra dimension indicating whether labels correspond to true objects or padding. For this task, we use the original version of CLEVR~\citep{johnson2017clevr} to be consistent with~\citep{zhang2019deep} and compare with their reported numbers as well as our best-effort re-implementation. We pre-process the object location to be in $[0, 1]$ and reshape the images to a resolution of $128\times128$. Image features (RGB values) are normalized to $[-1, 1]$.

\paragraph{Object Discovery}
For object discovery, we use three of the datasets provided by the Multi-Object Datasets library~\citep{multiobjectdatasets19}, available at \url{https://github.com/deepmind/multi_object_datasets}. See the aforementioned repository for a detailed description of the datasets. We use CLEVR (with masks), Multi-dSprites, and Tetrominoes. We split the TFRecords file of the CLEVR (with masks) dataset into multiple shards to allow for faster loading of the dataset from disk. We normalize all image features (RGB values) to $[-1, 1]$. Images in Tetrominoes and Multi-dSprites are of resolution $35\times35$ and $64\times64$, respectively. For CLEVR (with masks), we perform a center-crop with boundaries $[29, 221]$ (width) and $[64, 256]$ (height), as done in~\citep{greff2019multi}, and afterwards resize the cropped images to a resolution of $128\times128$. As done in~\citep{greff2019multi}, we filter the CLEVR (with masks) dataset to only retain scenes with a maximum number of 6 objects, and we refer to this dataset as CLEVR6, whereas the original dataset is referred to as CLEVR10.

\subsection{Metrics}
\paragraph{ARI}
Following earlier work~\citep{greff2019multi}, we use the Adjusted Rand Index (ARI)~\citep{rand1971objective,hubert1985comparing} score to compare the predicted alpha masks produced by our decoder against ground truth instance segmentation masks. ARI is a score that measures clustering similarity, where an ARI score of 1 corresponds to a perfect match and a score of 0 corresponds to chance level. We exclude the background label when computing the ARI score as done in~\citep{greff2019multi}. We use the implementation provided by the Multi-Object Datasets library~\citep{multiobjectdatasets19}, available at \url{https://github.com/deepmind/multi_object_datasets}. For a detailed description of the ARI score, see the aforementioned repository.

\paragraph{Average Precision} We consider the same setup of~\citet{zhang2019deep}, where the average precision is computed across all images in the validation set. As the network predicts a confidence for each detection (real objects have target 1, padding objects 0), we first sort the predictions based on their prediction confidence. For each prediction, we then check if in the corresponding ground truth image there was an object with the matching properties. A detection is considered a true positive if the discrete predicted properties (obtained with an argmax) exactly match the ground truth and the position of the predicted object is within a distance threshold of the ground truth. Otherwise, a detection is considered a false positive. We then compute the area under the smoothed precision recall curve at unique recall values as also done in~\citep{everingham2015pascal}. Ours is a best-effort re-implementation of the AP score as described in~\citep{zhang2019deep}. The implementation provided by~\citep{zhang2019deep} can be found at \url{https://github.com/Cyanogenoid/dspn}.

\end{document}